\documentclass{article}

    \usepackage[preprint]{neurips_2026}

\usepackage[utf8]{inputenc} 
\usepackage[T1]{fontenc}    
\usepackage{hyperref}       
\usepackage{url}            
\usepackage{booktabs}       
\usepackage{amsfonts}       
\usepackage{nicefrac}       
\usepackage{microtype}      
\usepackage{xcolor}         
\usepackage{graphicx}
\usepackage{epstopdf}
\usepackage{subcaption}
\usepackage{amsmath}
\usepackage{amsthm}
\usepackage{amssymb}
\usepackage{mathtools}
\usepackage{tikz}
\usepackage{multirow}
\usepackage{listings}
\lstdefinestyle{promptlst}{%
  basicstyle=\small\ttfamily,
  backgroundcolor=\color{gray!3},
  frame=single,
  framerule=0.5pt,
  rulecolor=\color{black!50},
  breaklines=true,
  breakatwhitespace=false,
  columns=fullflexible,
  keepspaces=true,
  upquote=true,
  aboveskip=4pt,
  belowskip=4pt,
  xleftmargin=4pt,
  xrightmargin=4pt,
}
\usepackage{tcolorbox}
\usepackage{colortbl}
\usepackage{makecell}
\usepackage{siunitx}
\usepackage{array}
\usepackage{fontawesome}
\usepackage{threeparttable}
\hypersetup{
    colorlinks=true,
    linkcolor=blue,
    filecolor=magenta,
    urlcolor=cyan,
    pdftitle={Overleaf Example},
    pdfpagemode=FullScreen,
    }

\usepackage{amsmath,amsthm,amsfonts,bm}

\DeclareMathAlphabet{\mathsfit}{\encodingdefault}{\sfdefault}{m}{sl}
\SetMathAlphabet{\mathsfit}{bold}{\encodingdefault}{\sfdefault}{bx}{n}

\renewcommand{\hat}{\widehat}

\newcommand{\vertiii}[1]{{\left\vert\kern-0.25ex\left\vert\kern-0.25ex\left\vert #1 
    \right\vert\kern-0.25ex\right\vert\kern-0.25ex\right\vert}}

\usetikzlibrary{arrows.meta, positioning}

\definecolor{highlight}{RGB}{220, 50, 47}
\definecolor{lightblue}{rgb}{0.85, 0.93, 1}
\definecolor{lightyellow}{rgb}{1.0, 0.96, 0.82}

\newcolumntype{M}{>{$}c<{$}}

\usepackage{color}

\usepackage{silence}
\usepackage{titletoc}

\title{UltraVR: A Diagnostic Ultra-Resolution Image–VQA Benchmark for Evidence-Grounded Reasoning}

\author{%
  \textbf{Gexin Huang}$^{1,2}$, 
  \textbf{Yanting Yang}$^{1,2}$, 
  \textbf{Myeongkyun Kang}$^{1,2}$, 
  \textbf{Beidi Zhao}$^{1,2}$, 
  \textbf{Jun Zhou}$^{4}$, \\
  \textbf{Chen Zhou}$^{1,3}$, 
  \textbf{Gang Wang}$^{1,3}$, 
  \textbf{Zu-hua Gao}$^{1,3}$, 
  \textbf{Xiaoxiao Li}$^{1,2}$ \\
  \\
  $^1$University of British Columbia, Vancouver, BC, Canada \\
  $^2$Vector Institute, Toronto, ON, Canada \\
  $^3$BC Cancer Agency \\
  $^4$The Hong Kong Polytechnic University, Hong Kong SAR, China \\
  \\
  \texttt{gexinml@gmail.com, yanting.yang@ece.ubc.ca, myeongkyun.kang@ubc.ca,} \\
  \texttt{beidiz@student.ubc.ca, zachary-jun.zhou@connect.polyu.hk,} \\
  \texttt{czhou@bccancer.bc.ca, gang.wang1@bccancer.bc.ca,} \\
  \texttt{zuhua.gao@ubc.ca, xiaoxiao.li@ece.ubc.ca}
}

\begin{document}

\maketitle

\begin{abstract}
Vision-language models (VLMs) have advanced rapidly on visual question answering and multimodal reasoning benchmarks. Yet it remains unclear whether they can reason over ultra-resolution images, where answer-critical evidence may be tiny, subtle, spatially distant, or distributed across regions. Existing evaluations largely report final-answer accuracy, offering limited insight into whether models acquire and integrate the necessary visual evidence. We introduce \textbf{UltraVR}, a diagnostic benchmark for evidence-grounded visual reasoning over ultra-resolution images. UltraVR spans four high-value scenarios: Closed-Circuit Television (CCTV) surveillance, remote sensing (RS), whole-slide image (WSI) pathology, and industrial anomaly detection (AD). These domains pose complementary challenges, including fine-grained object grounding in crowded CCTV scenes, long-range spatial comparison in RS, multi-scale evidence navigation in WSI, and subtle irregularity detection in repetitive industrial layouts. Beyond image-question-answer triples, each UltraVR instance includes a structured ground-truth chain of thought with step-level questions, intermediate answers, and reasoning operation labels. These labels decompose reasoning into evidence grounding, local perception, quantification, evidence integration, and decision inference, enabling process-level diagnosis rather than black-box answer scoring. 
Using UltraVR, we evaluate frontier proprietary and open-weight VLMs and show that current models remain far from reliable on ultra-resolution reasoning. More importantly, the structured annotations allow us to localize failures across the visual-to-decision pipeline: errors concentrate in evidence grounding and local perception, while downstream inference often recovers when intermediate visual facts are supplied. These findings demonstrate UltraVR as a diagnostic testbed for measuring not only whether VLMs answer correctly, but where their ultra-resolution reasoning process breaks.


\end{abstract}

\section{Introduction}

Vision-language models (VLMs) have rapidly progressed from image-level perception toward complex visual reasoning.
However, many high-value applications require reasoning over ultra-resolution imagery, including public safety and urban monitoring~\cite{feng2025urbanllava}, geospatial and environmental intelligence~\cite{liu2024remoteclip}, biomedical decision support~\cite{lu2024conch}, and industrial inspection~\cite{jeong2023winclip}.
These inputs are not merely larger versions of ordinary VQA images: their answer-critical evidence may be tiny, subtle, spatially distant, or distributed across multiple regions, making coarse image-level understanding insufficient.
Solving such tasks requires models to acquire task-relevant evidence from a large image, preserve fine local details, integrate visual facts across regions or scales, and infer the answer from the accumulated evidence.
This poses fundamental challenges for current VLM pipelines, which compress ultra-resolution inputs through resizing, tiling, cropping, multi-scale encoding, or token compression~\citep{qwen2026qwen35,wang2025internvl35,google2026gemma4modelcard,glmvteam2025glm45v,wu2024deepseekvl2,yu2025minicpmv45,aneja2026phi4reasoningvision,kimiteam2025kimivl}.
As a result, model errors can arise from multiple stages of the visual-to-decision pipeline, including missed evidence, lost local detail, weak cross-region integration, or incorrect final inference.
Final-answer accuracy alone cannot distinguish these sources of failure.

Recent benchmarks have made important progress in evaluating either multimodal reasoning or large-image understanding, but they still leave an evaluation gap for traceable reasoning over ultra-resolution images.
Visual reasoning benchmarks test spatial, physical, knowledge-intensive, and puzzle-like reasoning~\citep{feng2025reasonmap,jiang2026pix2fact,li2025quantiphy,zhou2025mira}, but they are often built on standard images, diagrams, videos, or controlled task settings rather than tasks that require recovering sparse evidence from very large images.
High-resolution benchmarks stress image scale and detail preservation~\citep{wu2024vstar,wang2024hrbench,ma2024gigagrounding,zhang2025hrscene}, but many focus on perception, search, grounding, or regional utilization rather than evidence-based inference.
Recent large-image reasoning benchmarks move closer to this setting~\citep{wang2025xlrsbench,dang2025rshrbench,li2026urbench}, but their supervision is mainly task-level QA: they show whether a model reaches the correct answer, while providing limited support for tracing which intermediate visual operation fails.
A benchmark for ultra-resolution visual reasoning should therefore measure not only whether a model answers correctly, but also whether it can acquire, perceive, integrate, and use the necessary visual evidence.
\begin{figure*}[t]
    \centering
    \includegraphics[width=\textwidth]{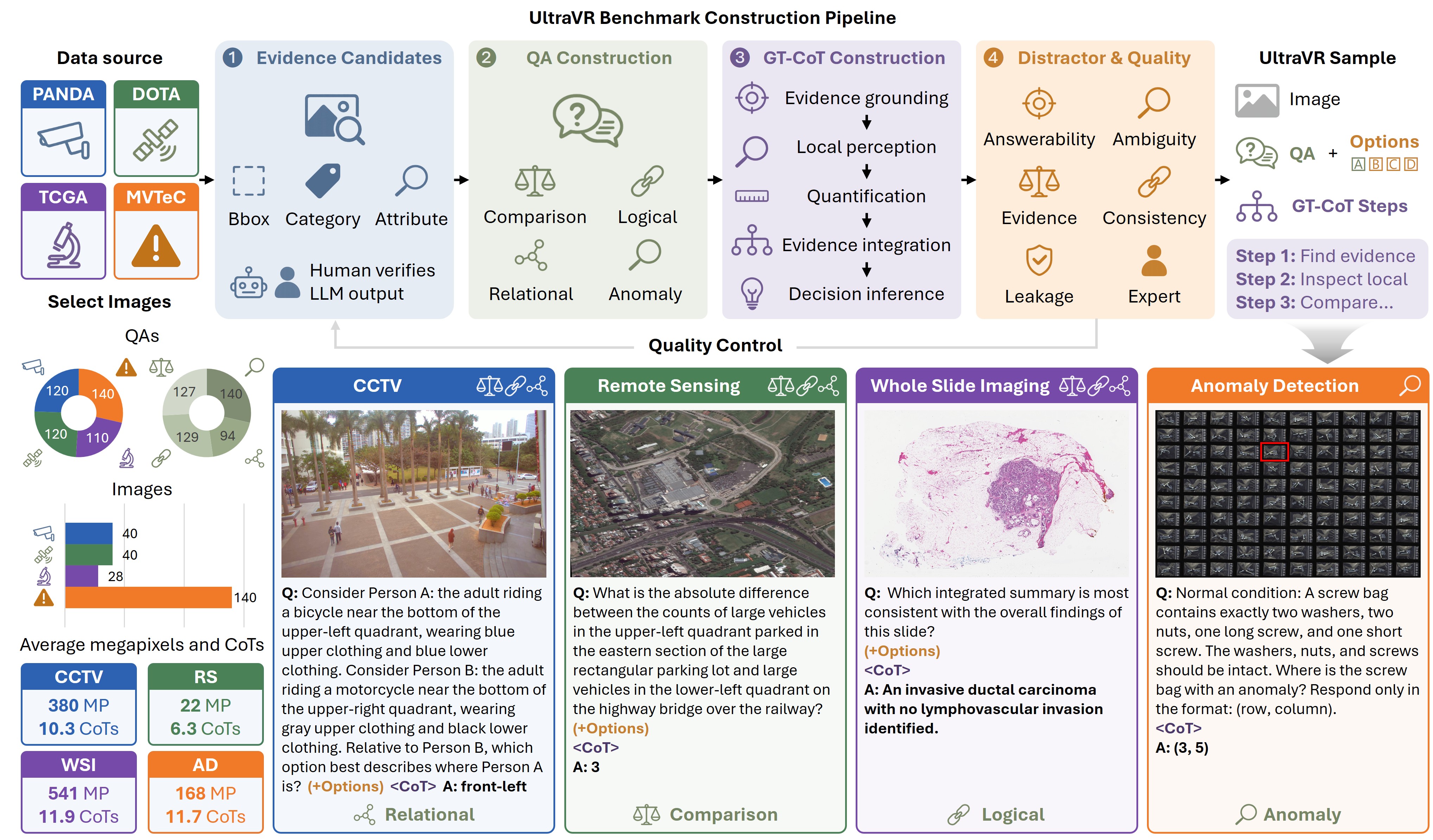}
    \caption{
    \textbf{Overview and construction pipeline of UltraVR.}
    UltraVR converts ultra-resolution images from four domains into evidence-grounded QA instances with structured GT-CoT annotations.
    Each instance contains a final question, answer options, step-level intermediate questions, verified step answers, and operation labels.
    The construction pipeline extracts domain-specific evidence candidates, instantiates reasoning templates, generates plausible distractors, and applies quality-control checks for visual answerability, option ambiguity, evidence visibility, GT-CoT consistency, text-only leakage, and expert review.
    }
    \label{fig:ultravr_overview}
    \vspace{-0.5cm}
\end{figure*}

To address this gap, we introduce \textbf{UltraVR} (see Fig.~\ref{fig:ultravr_overview}), a multi-domain diagnostic benchmark for evidence-grounded visual reasoning over ultra-resolution images.
UltraVR spans four domains: CCTV surveillance, remote sensing (RS), whole-slide imaging (WSI), and anomaly detection (AD), which instantiate complementary forms of large-image evidence acquisition and reasoning.
CCTV, RS, and WSI are organized around comparison, logical verification, and relational inference, while AD focuses on logical and structural anomaly reasoning.
Rather than testing direct recognition from salient objects or coarse scene context, UltraVR questions require models to search for task-relevant evidence, perceive local details, integrate visual facts, and draw a conclusion from the accumulated evidence.

A central feature of UltraVR is its structured ground-truth chain of thought (GT-CoT).
Each question is paired with a sequence of verifiable intermediate steps, where each step contains a step question, a ground-truth step answer, and an operation label.
These operation labels abstract the visual-to-decision process into five shared operations: evidence grounding (GND), local perception (PER), quantification (QUA), evidence integration (INT), and decision inference (INF).
This annotation design turns each QA item into a traceable diagnostic sample, enabling process-level evaluation beyond black-box final-answer scoring.

Using UltraVR, we benchmark frontier proprietary and open-weight VLMs under three complementary evaluation axes: end-to-end QA, visual-evidence access evaluation, and GT-CoT process-level evaluation.
The end-to-end evaluation shows that current VLMs remain far from saturation on UltraVR, with the strongest model reaching only 44.9\% macro accuracy.
The visual-evidence access evaluation shows that localized evidence improves performance only partially and model-dependently, suggesting that default full-image inference is limited by evidence access, preservation, and retrieval.
The GT-CoT process-level evaluation further shows that correcting intermediate visual facts through GT-prefix evaluation substantially improves performance, while operation-level analysis localizes many errors to early evidence grounding and local perception.
Together, these results demonstrate the diagnostic utility of UltraVR: the benchmark not only measures final-answer accuracy, but also identifies where the ultra-resolution visual reasoning process breaks.

Our contributions are:
(1) We introduce UltraVR, a multi-domain diagnostic benchmark for evidence-grounded ultra-resolution visual reasoning across CCTV, RS, WSI, and AD.
(2) We design a unified reasoning taxonomy covering comparison, logical verification, relational inference, and anomaly reasoning, with domain-specific templates that require visually grounded evidence acquisition rather than text-only priors.
(3) We provide structured GT-CoT annotations with step-level questions, verified intermediate answers, and operation labels, enabling process-level evaluation over GND, PER, QUA, INT, and INF.
(4) We benchmark frontier VLMs under controlled end-to-end, visual-evidence, and GT-CoT process-level evaluations, revealing systematic bottlenecks in early visual evidence grounding and local perception.

\section{Related Work}
\label{sec:related_work}
\subsection{Visual Reasoning Benchmarks}
General multimodal benchmarks have substantially advanced the evaluation of VLM reasoning.
MME evaluates perception and cognition abilities with manually designed instruction-answer pairs~\citep{fu2023mme}.
MMBench provides a multiple-choice suite for broad multimodal capability assessment~\citep{liu2023mmbench}.
MMMU evaluates college-level multimodal problems that require domain knowledge and deliberate reasoning~\citep{yue2023mmmu}.
Recent benchmarks further target more specific forms of visual reasoning.
REASONMAP evaluates fine-grained spatial reasoning over transit maps~\citep{feng2025reasonmap}.
Pix2Fact studies fine-grained visual grounding, multi-hop reasoning, and external knowledge in high-resolution daily-life images~\citep{jiang2026pix2fact}.
QuantiPhy evaluates whether VLMs can infer quantitative physical properties from videos~\citep{li2025quantiphy}.
MIRA studies visual reasoning problems where intermediate visual representations, such as sketches or diagrams, are important for solving the task~\citep{zhou2025mira}.
These works reveal important reasoning limitations of VLMs, but they do not primarily target evidence acquisition, fine-detail preservation, and cross-region integration under ultra-resolution visual inputs.
UltraVR complements them by constructing evidence-grounded reasoning tasks over heterogeneous ultra-resolution domains.

\subsection{Ultra-Resolution VLM Evaluation}
A growing line of work evaluates VLMs under high-resolution inputs.
V* introduces guided visual search for locating important details in high-resolution and crowded images~\citep{wu2024vstar}.
HR-Bench evaluates MLLM perception on 4K and 8K images and shows that downsampling can remove answer-critical information~\citep{wang2024hrbench}.
GigaGrounding extends visual grounding to gigapixel-level large-scale scenes with large spatial context, scale variation, and multi-hop referring expressions~\citep{ma2024gigagrounding}.
HRScene covers diverse high-resolution scenes and studies regional utilization under large-image inputs~\citep{zhang2025hrscene}.
These benchmarks establish the importance of large-image evaluation, but their main focus is perception, search, grounding, or regional utilization rather than process-level reasoning diagnosis.
Recent work moves closer to ultra-resolution reasoning, but differs from UltraVR in how the benchmark instances are constructed.
XLRS-Bench constructs a large ultra-high-resolution remote-sensing benchmark by organizing RS images into VQA, captioning, and grounding annotations across perception and reasoning dimensions~\citep{wang2025xlrsbench}.
RSHR-Bench further revisits high-resolution RS reasoning by designing visually dependent QA tasks and applying text-only adversarial filtering with human verification to reduce language-prior shortcuts~\citep{dang2025rshrbench}.
UR-Bench constructs ultra-high-resolution QA over humanistic and natural scenes by composing Micro, Regional, and Global questions with increasing spatial integration requirements, and studies an agent framework for tool-based image exploration~\citep{li2026urbench}.
These benchmarks are closest to our setting in image scale and reasoning motivation.
UltraVR is complementary but differs in construction target: each instance is built around domain-specific evidence candidates and a functional visual-reasoning chain, then annotated with structured GT-CoT steps and operation labels.
This design turns each QA item into a traceable diagnostic sample, enabling operation-level failure localization in addition to task-level accuracy.

\section{UltraVR: A Diagnostic Benchmark}
\label{sec:benchmark}

\subsection{Design Principles and Task Taxonomy}
\label{sec:benchmark_overview}

UltraVR is designed as a cross-domain diagnostic benchmark for evidence-grounded visual reasoning over ultra-resolution images, with dataset statistics summarized in Fig.~\ref{fig:ultravr_overview}.
The benchmark follows three design principles.
First, each question should require visually grounded evidence rather than being answerable from language priors or coarse scene context alone.
Second, answer-critical evidence should reflect the core challenges of ultra-resolution imagery: it can be tiny, subtle, spatially distant, or distributed across regions and scales.
Third, each instance should support process-level diagnosis through verifiable intermediate visual facts, allowing evaluation beyond final-answer accuracy.

Following these principles, UltraVR constructs QA tasks according to a functional definition of visual reasoning: solving a question requires a model to search for task-relevant evidence in a large image, perceive fine local details, integrate visual facts across regions or scales, and infer the answer from the accumulated evidence.
UltraVR therefore does not target direct recognition from salient objects, but evidence-grounded multi-step reasoning over large visual inputs.

UltraVR spans four high-value ultra-resolution domains: CCTV, RS, WSI, and AD.
These domains instantiate complementary evidence challenges.
CCTV requires grounding fine-grained people or vehicles in crowded wide-field scenes.
RS requires comparing small objects and spatial structures across long-range aerial layouts.
WSI requires navigating multi-scale pathology evidence and integrating diagnostic findings.
AD requires detecting subtle violations in repetitive industrial grids.
For CCTV, RS, and WSI, questions are organized into three reasoning categories: comparison, logical verification, and relational inference.
Comparison requires comparing quantities, attributes, or interpretations across objects or regions; logical verification checks whether a visually grounded condition or rule is satisfied; and relational inference reasons about spatial, structural, or semantic relationships between visual entities.
For AD, we use anomaly reasoning, where the model must identify a local logical or structural pattern that violates the global regularity of the grid.

\subsection{Structured GT-CoT and Operation Labels}
\label{sec:gtcot_annotation}

A central annotation component of UltraVR is the structured ground-truth chain of thought (GT-CoT), which is not a model-generated rationale.
Instead, it is a benchmark annotation schema consisting of verifiable intermediate visual facts.
Each GT-CoT trace is represented as a sequence of reasoning steps, and each step contains a \emph{step question}, a \emph{ground-truth step answer}, and an \emph{operation label}.
The step question specifies the evidence-seeking or reasoning action, the step answer records the verified intermediate visual fact, and the operation label identifies the role of the step in the visual-to-decision chain.
Unlike free-form rationales, this structured format enables process-level evaluation: models can be assessed not only by whether they reach the correct final answer, but also by which intermediate operation fails.

We define five operation types shared across domains.
\textbf{Evidence grounding (GND)} locates task-relevant image evidence, such as objects, regions, grid cells, ROIs, or landmarks.
\textbf{Local perception (PER)} recognizes local properties of the acquired evidence, such as attributes, orientations, morphology, defect patterns, or visual states.
\textbf{Quantification (QUA)} converts visual observations into numerical or ordinal information, such as counts, size comparisons, differences, or ratios.
\textbf{Evidence integration (INT)} combines evidence across objects, regions, scales, or previous steps.
\textbf{Decision inference (INF)} derives the task-level conclusion by applying task-specific criteria, domain knowledge, logical conditions, pathology references, anomaly definitions, or answer-option rules to the accumulated evidence.
This shared taxonomy abstracts heterogeneous CCTV, RS, WSI, and AD tasks into a common visual-to-decision schema, enabling operation-level evaluation across domains.

\subsection{Benchmark Construction and Quality Control}
\label{sec:domain_construction}

UltraVR follows a shared construction protocol across domains, as shown in Fig.~\ref{fig:ultravr_overview}.
For each domain, we first select ultra-resolution images whose relevant evidence is spatially distributed or locally subtle.
We then extract domain-specific evidence candidates, instantiate reasoning-oriented QA templates, annotate GT-CoT traces with operation labels, and construct plausible distractors or constrained answer spaces. The detailed construction pipeline is provided in Appendix~\ref{app:benchmark_construction}.
Detailed construction templates, filtering rules, and domain-specific annotation procedures are provided in Appendix~\ref{app:experimental_details}.

\paragraph{CCTV surveillance.}
The CCTV split is built from PANDA~\cite{wang2020panda}, where wide-field surveillance frames contain dense pedestrians, small distant people, occlusions, and long-range spatial layouts.
This domain evaluates whether VLMs can ground people or vehicles in crowded scenes, perceive local appearance or orientation cues, and reason about counts or relations across distant regions.
We use person and vehicle annotations to form evidence candidates such as semantic object bundles, quadrant-level groups, and pairs of people with visually grounded descriptions.
Questions cover comparison, logical verification, and relational inference: models compare counts across bundles, verify count-and-region claims, or infer the relative position of one person with respect to another.
The corresponding GT-CoT traces ground the relevant boxes, perceive local cues when needed, quantify regional evidence, and integrate the evidence into the final decision.

\paragraph{Remote sensing.}
The RS split is constructed from DOTA 1.5~\cite{xia2018dota}, which contains aerial imagery with small objects distributed over large geographic layouts.
This domain evaluates long-range spatial comparison and fine-grained object grounding: relevant evidence may consist of tiny vehicles, ships, bridges, storage tanks, or sports fields located in different regions of a large image.
We parse object annotations into region-level evidence candidates and instantiate questions that compare object counts, verify orientation- or presence-based conditions, or infer spatial relations between selected objects across regions.
Unlike ordinary object recognition, these tasks require models to jointly preserve small targets and global layout, making them sensitive to downsampling and patch fragmentation.

\paragraph{Whole-slide imaging.}
The WSI split uses breast tumor whole-slide images from the TCGA-BRCA cohort~\cite{weinstein2013cancer}, where diagnostic evidence is distributed over gigapixel tissue slides and must be interpreted through pathology morphology.
Unlike CCTV and RS, WSI does not provide dense object boxes; instead, evidence is organized through slide-level diagnostic attributes and expert-reviewed regions of interest.
Questions ask models to compare clinically confusable interpretations, verify integrated pathology summaries, or reason about relationships between pathological components such as invasive and in-situ regions.
The GT-CoT follows a structured diagnostic workflow: low-power screening, ROI selection, local morphology inspection, criterion selection, and evidence-to-conclusion integration.
This design evaluates whether VLMs can navigate multi-scale pathology evidence rather than only match a diagnostic label.

\paragraph{Anomaly detection.}
The AD split simulates ultra-resolution industrial inspection by composing large repetitive grids from normal and anomalous samples extracted from MVTec LOCO AD~\citep{bergmann2022mvtecloco}.
Each image contains many visually similar normal cells and a rare cell that violates a logical or structural normality rule.
This domain evaluates exhaustive search, subtle irregularity perception, and rule-based anomaly localization under repetitive layouts.
We decompose each category-level normality description into visual conditions and annotate which condition is violated by the anomalous cell.
The GT-CoT verifies local or structural conditions and identifies the cell that violates the global regularity.

\paragraph{Quality control.}
We apply quality-control checks at both the QA and GT-CoT levels.
At the QA level, we verify visual answerability, single-answer validity, evidence visibility, distractor plausibility, and text-only leakage.
At the GT-CoT level, we check that each intermediate answer is well defined, supported by the specified evidence, and consistent with the final answer.
Domain-specific validation is also applied: WSI samples are reviewed by pathology experts, AD samples are checked for unique anomaly placement and normality-rule consistency, and CCTV/RS samples are verified against their evidence annotations.
Samples are revised or discarded when the answer is ambiguous, the evidence is not visually supported, or the question can be solved without image inspection.

\section{Evaluation Protocols}
\label{sec:experiments}

UltraVR defines three complementary evaluation axes. 
First, end-to-end QA evaluates whether a model can solve the task from the original ultra-resolution image under its default inference pipeline. 
Second, visual-evidence access conditions test whether errors arise from inaccessible or poorly preserved evidence by varying only the visual input. 
Third, GT-CoT process evaluation uses structured intermediate steps to localize errors across the visual-to-decision chain.

\subsection{Models and Metrics}
\label{sec:experimental_settings}

We evaluate both proprietary frontier VLMs and open-weight VLMs.
The proprietary models include GPT-5.5, GPT-5.4, and Gemini-3.1-Pro~\cite{openai2026gpt55,openai2026gpt54,googledeepmind2026gemini31pro}.
The open-weight models include Qwen3.5, InternVL3.5, Gemma-4, GLM-4.5V, DeepSeek-VL2, MiniCPM-V-4.5, Phi-4-Reasoning-Vision, and Kimi-VL-A3B-Thinking~\cite{qwen2026qwen35,wang2025internvl35,google2026gemma4modelcard,glmvteam2025glm45v,wu2024deepseekvl2,yu2025minicpmv45,aneja2026phi4reasoningvision,kimiteam2025kimivl}.
This model pool covers different parameter scales, visual-token budgets, and high-resolution image processing strategies.
Implementation details, including preprocessing, decoding, and API settings, are provided in Appendix~\ref{app:implementation_details}.

All tasks are evaluated in a four-choice format.
We report category-level and domain-level accuracy, together with a domain-balanced macro average over CCTV, RS, WSI, and AD.
For the end-to-end evaluation, we additionally report text-only accuracy, abstention rate, and the visual gap between the full-image Direct QA setting and the text-only control.
For process-level evaluation, we report final-answer accuracy under different GT-CoT conditions, operation-wise accuracy/error, and first-error distributions.
Answer extraction is described in Appendix~\ref{app:answer_extraction}, and all metric definitions are given in Appendix~\ref{app:metric_definitions}.

\subsection{Visual-Evidence Access Conditions}
\label{sec:evaluation_protocols}

We define five visual-input conditions to evaluate how access to visual evidence affects UltraVR performance.
\textbf{V0: Text-only.} The model receives only the question and answer options. This condition measures language-prior leakage and answer-option bias.
\textbf{V1: Thumbnail.} The model receives an externally downsampled global view of the full image. This condition tests whether coarse global context is sufficient for the task.
\textbf{V2: Full image.} The model receives the original full image under its default preprocessing pipeline. This is the standard visual-input condition for end-to-end evaluation.
\textbf{V3: Local evidence.} The model receives domain-specific local visual evidence without the original full image. WSI provides cropped ROI regions; AD provides a $5{\times}5$ cell sheet cropped from the original $10{\times}10$ grid, with the anomalous cell randomly placed within the crop; CCTV and RS provide quadrant-level crops rather than fine-grained answer-localizing crops. All crops are preserved at their original resolution before being fed into the model.
\textbf{V4: Full + local evidence.} The model receives both the original full image and the local evidence package. This condition tests whether models can use localized evidence together with global context.

\subsection{GT-CoT Process Evaluation}
\label{sec:gt_cot_protocols}

We define six GT-CoT conditions to evaluate whether different reasoning formats help models instantiate and use intermediate visual facts.
\textbf{S0: Direct QA.} The model receives the full image, final question, and answer options, and directly predicts the answer.
\textbf{S1: Generic CoT.} The model receives the same input as S0 plus a generic step-by-step reasoning instruction.
\textbf{S2: Schema-CoT.} The model receives the full image, final question, answer options, and the task-aware GT-CoT step schema, and is asked to produce all intermediate step answers and the final answer in a single response.
\textbf{S3: Few-shot GT-CoT.} The model receives in-context examples with complete ground-truth reasoning trajectories before answering the current test sample; no ground-truth intermediate answer for the test sample is revealed.
\textbf{S4: Pred-Step.} The model answers structured sub-questions sequentially, conditioning each step on its own previous predictions.
\textbf{S5: GT-Prefix.} At step $t$, all previous intermediate answers are replaced by ground-truth intermediate answers, while the model still has to answer the current step and final question.

\section{Benchmark Results and Diagnostic Analysis}
\label{sec:results}

\begin{table*}[t]
\centering
\scriptsize
\setlength{\tabcolsep}{1.55pt}
\renewcommand{\arraystretch}{1.08}
\caption{
\textbf{End-to-end UltraVR evaluation.}
We report accuracy (\%) across four ultra-resolution domains and their reasoning categories under the standard full-image setting.
Max Visual Budget denotes the maximum visual input tokens, and Max Res. denotes the effective equivalent square resolution.
Comp.: comparison; Logic: logical verification; Rel.: relational inference; L-Anom.: logical anomaly; S-Anom.: structural anomaly.
Overall is the domain-level accuracy, and Macro Acc. is the domain-balanced average across CCTV, RS, WSI, and AD.
Text-only Acc. and Abs. rate are measured without visual input, and Visual Gap is computed as Macro Acc. $-$ Text-only Acc.
}
\label{tab:main_results}
\resizebox{\textwidth}{!}{%
\begin{tabular}{lcc cccc cccc cccc ccc ccc}
\toprule
\multirow{2}{*}{\textbf{Model}}
& \multirow{2}{*}{\makecell{\textbf{Max Visual}\\\textbf{Budget}}}
& \multirow{2}{*}{\makecell{\textbf{Max Res.}\\\textbf{(px)}}}
& \multicolumn{4}{c}{\textbf{CCTV}}
& \multicolumn{4}{c}{\textbf{RS}}
& \multicolumn{4}{c}{\textbf{WSI}}
& \multicolumn{3}{c}{\textbf{AD}}
& \multicolumn{3}{c}{\textbf{Summary}} \\
\cmidrule(lr){4-7}
\cmidrule(lr){8-11}
\cmidrule(lr){12-15}
\cmidrule(lr){16-18}
\cmidrule(lr){19-21}
& &
& Comp. & Logic & Rel. & Overall
& Comp. & Logic & Rel. & Overall
& Comp. & Logic & Rel. & Overall
& L-Anom. & S-Anom. & Overall
& \makecell{Macro\\Acc.}
& \makecell{Text-only\\Acc./Abs. rate}
& \makecell{Visual\\Gap} \\
\midrule

\rowcolor{gray!20}
\multicolumn{21}{l}{\textit{Closed-source frontier VLMs}} \\
GPT-5.5~\cite{openai2026gpt55}
& -- & $3200^2$
& 38.9 & \textbf{34.2} & \textbf{80.0} & \textbf{51.8}
& 27.5 & 40.0 & \textbf{65.0} & \textbf{44.2}
& 36.2 & 71.4 & 42.9 & 52.7
& \textbf{34.8} & 25.9 & \textbf{30.8}
& \textbf{44.9} & 3.9/95.7 & \cellcolor{lightyellow}+41.0 \\
GPT-5.4~\cite{openai2026gpt54}
& -- & $3200^2$
& 41.7 & 28.9 & 42.5 & 37.7
& \textbf{47.5} & 12.5 & 47.5 & 35.8
& 14.9 & 55.1 & 50.0 & 37.3
& 28.8 & 11.1 & 20.8
& 32.9 & 3.5/94.2 & \cellcolor{lightyellow}+29.4 \\
Gemini-3.1-Pro~\cite{googledeepmind2026gemini31pro}
& 2240 & $2260^2$
& \textbf{50.0} & 28.9 & 60.0 & 46.5
& 22.5 & 30.0 & 51.3 & 34.5
& \textbf{68.1} & 53.1 & 14.3 & \textbf{54.5}
& 27.5 & \textbf{32.4} & 30.0
& 41.4 & 3.1/94.0 & \cellcolor{lightyellow}+38.3 \\
\midrule

\rowcolor{gray!20}
\multicolumn{21}{l}{\textit{Open-weight large VLMs}} \\
Qwen3.5-27B~\cite{qwen2026qwen35}
& 16384 & $4096^2$
& 33.3 & 26.3 & 35.0 & 31.6
& 2.5 & \textbf{45.0} & 50.0 & 32.5
& 38.3 & \textbf{73.5} & 21.4 & 51.8
& 22.7 & 18.5 & 20.8
& 34.2 & 5.7/82.2 & \cellcolor{lightyellow}+28.5 \\
InternVL3.5-38B~\cite{wang2025internvl35}
& 32768 & $5068^2$
& 25.0 & 21.1 & 55.0 & 34.2
& 15.0 & 27.5 & 40.0 & 27.5
& 46.8 & 63.3 & 21.4 & 50.9
& 4.5 & 1.9 & 3.3
& 29.0 & 13.9/42.4 & \cellcolor{lightyellow}+15.1 \\
Gemma-4-26B~\cite{google2026gemma4modelcard}
& 1120 & $1606^2$
& 41.7 & 28.9 & 72.5 & 48.2
& 2.5 & 20.0 & 42.5 & 21.7
& 34.0 & 57.1 & 28.6 & 43.6
& 12.1 & 5.6 & 9.2
& 30.7 & 2.9/95.7 & \cellcolor{lightyellow}+27.8 \\
GLM-4.5V~\cite{glmvteam2025glm45v}
& 12288 & $3103^2$
& 45.0 & 25.0 & 37.5 & 35.8
& 25.0 & 32.5 & 32.5 & 30.0
& 38.2 & 50.3 & 35.4 & 41.3
& 5.8 & 5.6 & 5.7
& 28.2 & 3.5/76.9 & \cellcolor{lightyellow}+24.7 \\
\midrule

\rowcolor{gray!20}
\multicolumn{21}{l}{\textit{Open-weight small VLMs}} \\
DeepSeek-VL2~\cite{wu2024deepseekvl2}
& 2101 & $1152^2$
& 13.9 & 26.3 & 35.0 & 25.4
& 35.0 & 42.5 & 55.0 & \textbf{44.2}
& 44.7 & 46.9 & 14.3 & 41.8
& 0.0 & 0.0 & 0.0
& 27.8 & 16.3/36.7 & \cellcolor{lightyellow}+11.5 \\
MiniCPM-V 4.5~\cite{yu2025minicpmv45}
& 640 & $1344^2$
& 19.4 & 31.6 & 57.5 & 36.8
& 12.5 & 32.5 & 45.0 & 30.0
& 38.3 & 55.1 & 57.1 & 48.2
& 1.4 & 0.0 & 0.7
& 28.9 & 18.6/14.5 & \cellcolor{lightyellow}+10.3 \\
Phi-4-Reasoning-Vision-15B~\cite{aneja2026phi4reasoningvision}
& 3600 & $960^2$
& 38.9 & 26.3 & 52.5 & 39.5
& 30.0 & 32.5 & 62.5 & 41.7
& 19.1 & 69.4 & 21.4 & 41.8
& 2.9 & 0.0 & 1.4
& 31.1 & 3.7/92.2 & \cellcolor{lightyellow}+27.4 \\
Kimi-VL-A3B-Thinking~\cite{kimiteam2025kimivl}
& 4096 & $1792^2$
& 22.2 & 26.3 & 72.5 & 41.2
& 0.0 & 20.0 & 45.0 & 21.7
& 34.0 & 69.4 & \textbf{64.3} & 53.6
& 0.0 & 1.4 & 0.7
& 29.3 & 10.0/62.9 & \cellcolor{lightyellow}+19.3 \\
\midrule

\rowcolor{gray!10}
\multicolumn{3}{l}{\textit{Mean}}
& 35.1 & 28.2 & 54.3 & 39.0
& 19.7 & 30.7 & 49.6 & 33.1
& 37.5 & 60.1 & 35.6 & 47.0
& 12.8 & 9.4 & 11.2
& 32.6 & 7.7/71.6 & \cellcolor{lightyellow}+24.8 \\
\bottomrule
\end{tabular}%
}

\end{table*}

\subsection{End-to-End UltraVR Evaluation}
\label{sec:main_results}

The end-to-end UltraVR evaluation measures whether current VLMs can solve evidence-grounded ultra-resolution visual reasoning under their default inference setting. 
Specifically, we evaluate each model under S0/V2, where the model receives the original ultra-resolution image, the question, and the answer options, and directly predicts the final answer without local evidence packages, intermediate supervision, or reasoning scaffolds.
This setting evaluates the full visual-to-decision pipeline: acquiring task-relevant evidence from a large image, preserving fine details under the model's native preprocessing, integrating evidence across regions, and producing the final decision.

Table~\ref{tab:main_results} reports domain-level accuracy and domain-balanced macro accuracy. 
We also include the text-only control V0, where the image is removed and only the question and answer options are provided.
The abstention rate measures the fraction of V0 responses in which the model refuses to choose an option or states that the answer cannot be determined without the image, while the visual gap measures the performance gain from adding the image back.

\paragraph{Observations.}
UltraVR remains far from saturated by current VLMs. 
Under the standard S0/V2 setting, the strongest model, GPT-5.5, reaches only 44.9\% macro accuracy, and the average across all models is 32.6\%, indicating substantial headroom for ultra-resolution visual reasoning. 
The text-only control further supports the visual dependence of UltraVR: removing the image reduces the mean accuracy to 7.7\%, while stronger instruction-following models often abstain when visual evidence is unavailable. 
Performance also varies substantially across domains and reasoning categories, and larger visual budgets or higher maximum resolutions do not consistently predict better accuracy. 
These results establish UltraVR as a challenging and visually grounded benchmark, and motivate the following diagnostics on visual-evidence access and process-level reasoning behavior.

\begin{table*}[t]
\centering
\small
\setlength{\tabcolsep}{4.2pt}
\caption{
\textbf{Visual-evidence access evaluation.}
We compare V1--V4 while keeping the question, answer options, and decoding setup fixed.
V1 provides a coarse global view, V2 uses the original full image, V3 provides domain-specific local evidence without directly revealing the answer, and V4 combines the full image with local evidence.
This evaluation measures whether UltraVR errors are associated with limited evidence access, poor evidence preservation, or unreliable evidence use.
}
\label{tab:visual_evidence_intervention}
\resizebox{\textwidth}{!}{%
\begin{tabular}{llccccccc}
\toprule
& &
\multicolumn{4}{c}{Accuracy} &
\multicolumn{3}{c}{Diagnostic gap} \\
\cmidrule(lr){3-6} \cmidrule(lr){7-9}
Type & Model
& \begin{tabular}{@{}c@{}}Thumbnail \\ (V1)\end{tabular}
& \begin{tabular}{@{}c@{}}Full image \\ (V2)\end{tabular}
& \begin{tabular}{@{}c@{}}Local evidence \\ (V3)\end{tabular}
& \begin{tabular}{@{}c@{}}Full + local evidence \\ (V4)\end{tabular}
& $\Delta_{\mathrm{res}}$
& $\Delta_{\mathrm{access}}$
& $\Delta_{\mathrm{aug}}$ \\
\midrule
\multirow{3}{*}{Closed}
& GPT-5.5~\cite{openai2026gpt55}
& 36.0 & 44.6 & \textbf{50.1} & 43.0 & \cellcolor{lightyellow}+8.6 & \cellcolor{lightyellow}+5.5 & \cellcolor{lightblue}-1.6 \\
& GPT-5.4~\cite{openai2026gpt54}
& 29.2 & 32.8 & \textbf{38.4} & 33.7
& \cellcolor{lightyellow}+3.6
& \cellcolor{lightyellow}+5.6
& \cellcolor{lightyellow}+0.9 \\
& Gemini-3.1-Pro~\cite{googledeepmind2026gemini31pro}
& 36.7 & 40.6 & \textbf{48.8} & 39.4
& \cellcolor{lightyellow}+3.9
& \cellcolor{lightyellow}+8.2
& \cellcolor{lightblue}-1.2 \\
\midrule
\multirow{4}{*}{Open-Large}
& Qwen3.5-27B~\cite{qwen2026qwen35}
& 24.5 & 33.8 & \textbf{37.8} & 24.7
& \cellcolor{lightyellow}+9.3
& \cellcolor{lightyellow}+4.0
& \cellcolor{lightblue}-9.1 \\
& InternVL3.5-38B~\cite{wang2025internvl35}
& 28.2 & 28.4 & 27.6 & \textbf{29.0}
& \cellcolor{lightyellow}+0.2
& \cellcolor{lightblue}-0.8
& \cellcolor{lightyellow}+0.6 \\
& Gemma-4-26B~\cite{google2026gemma4modelcard}
& 29.4 & 30.2 & \textbf{38.0} & 29.8
& \cellcolor{lightyellow}+0.8
& \cellcolor{lightyellow}+7.8
& \cellcolor{lightblue}-0.4 \\
& GLM-4.5V~\cite{glmvteam2025glm45v}
& 22.7 & 27.3 & \textbf{30.8} & 24.5
& \cellcolor{lightyellow}+4.6
& \cellcolor{lightyellow}+3.5
& \cellcolor{lightblue}-2.8 \\
\midrule
\multirow{4}{*}{Open-Small}
& DeepSeek-VL2~\cite{wu2024deepseekvl2}
& 27.8 & 26.4 & \textbf{29.4} & 26.1
& \cellcolor{lightblue}-1.4
& \cellcolor{lightyellow}+3.0
& \cellcolor{lightblue}-0.3 \\
& MiniCPM-V 4.5~\cite{yu2025minicpmv45}
& 21.8 & 27.3 & \textbf{31.2} & 30.4
& \cellcolor{lightyellow}+5.5
& \cellcolor{lightyellow}+3.9
& \cellcolor{lightyellow} +3.1 \\
& Phi-4-RV-15B~\cite{aneja2026phi4reasoningvision}
& 27.1 & 29.5 & \textbf{30.2} & 30.0
& \cellcolor{lightyellow}+2.4
& \cellcolor{lightyellow}+0.7
& \cellcolor{lightyellow}+0.5 \\
& Kimi-VL-A3B-Thinking~\cite{kimiteam2025kimivl}
& 26.3 & 27.5 & \textbf{30.4} & 25.5
& \cellcolor{lightyellow}+1.2
& \cellcolor{lightyellow}+2.9
& \cellcolor{lightblue}-2.0 \\
\midrule
\multicolumn{2}{l}{Mean}
& 27.4 & 30.4 & 33.2 & 26.0
& \cellcolor{lightyellow}+3.0
& \cellcolor{lightyellow}+2.8
& \cellcolor{lightblue}-1.1 \\
\bottomrule
\end{tabular}
}
\end{table*}

\subsection{Visual-Evidence Access Evaluation}
\label{sec:visual_evidence_intervention}

The visual-evidence access evaluation measures whether model errors arise from limited access to answer-critical evidence or from unreliable use of evidence once it is surfaced. We keep the language prompt, answer options, and decoding setup fixed, and vary only the visual input. We compare four conditions: V1 provides a coarse global view, V2 uses the original full image under each model's default preprocessing, V3 provides domain-specific local evidence without directly revealing the answer, and V4 combines the full image with local evidence.
Table~\ref{tab:visual_evidence_intervention} reports the results.

\textbf{Observations.}
First, V2 improves over V1, but only modestly, indicating that the full image contains useful information beyond a coarse global view while current VLMs do not fully exploit the additional detail. Second, V3 is the strongest visual-input condition for most models, suggesting that default full-image inference is limited by evidence access, compression, or retrieval. Third, V4 does not reliably outperform V2 or V3, indicating that simply adding local evidence to the full image is not a robust solution. Overall, the visual-evidence access evaluation shows that localized evidence reduces, but does not eliminate, the ultra-resolution reasoning bottleneck.

\subsection{Process-Level Evaluation with GT-CoT}
\begin{figure*}[t]
    \centering
    \includegraphics[width=\textwidth]{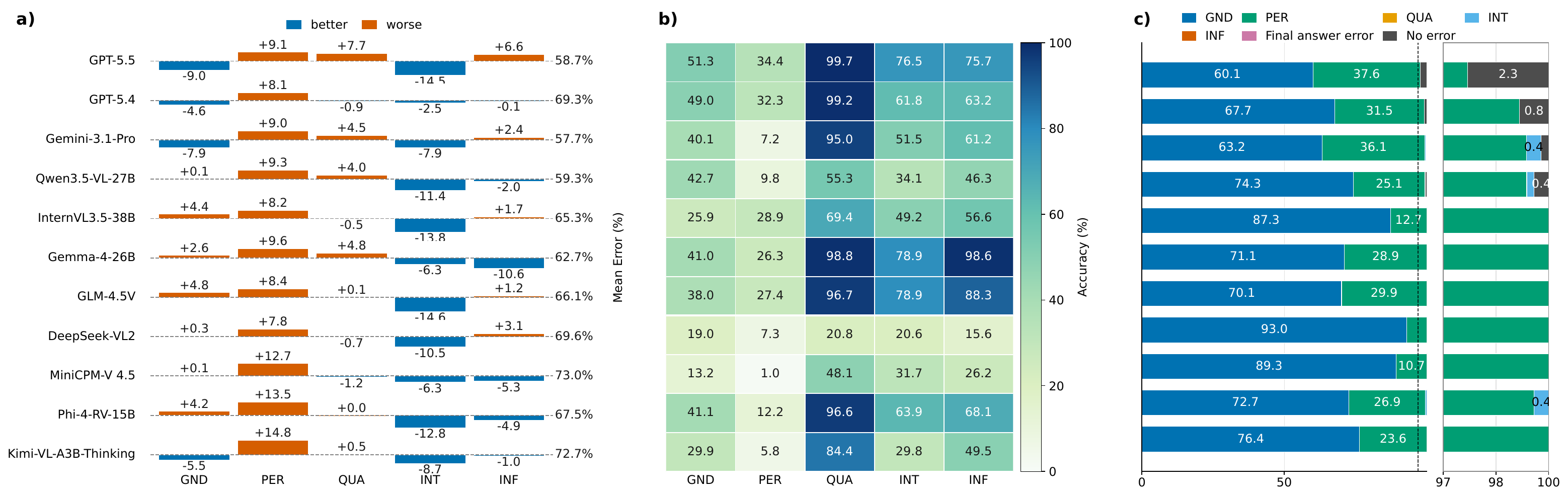}
    \caption{
    \textbf{Operation-level evaluation with GT-CoT.}
    (a) \textbf{Relative operation difficulty under S2.}
    S2 completes the GT-CoT schema in a single response; each bar reports an operation's error deviation from that model's mean step error.
    (b) \textbf{Operation recoverability under S5.}
    Previous intermediate answers are replaced with ground truth, isolating the current operation from earlier accumulated errors.
    (c) \textbf{First-error locations under S4.}
    S4 rolls out the trajectory sequentially with model-predicted prefixes, allowing us to identify where errors first enter the reasoning chain.
    The inset zooms into rare final-answer-only and no-error cases.
    }
    \label{fig:operation_analysis}
\end{figure*}
The GT-CoT process-level evaluation extends UltraVR beyond final-answer scoring by evaluating the intermediate visual-to-decision process.
Each sample is paired with structured intermediate questions, verified step answers, and operation labels, allowing us to evaluate whether errors arise from missing reasoning structure, incorrect intermediate visual states, or error propagation across steps.

\begin{table*}[t]
\centering
\small
\setlength{\tabcolsep}{3.5pt}
\renewcommand{\arraystretch}{1.1}
\caption{
\textbf{GT-CoT process-level evaluation.}
Domain-balanced macro accuracy (\%) is reported under six controlled reasoning formats:
direct prediction (S0), generic reasoning prompt (S1), one-shot GT-CoT schema completion (S2), few-shot GT trajectories (S3), sequential prediction with model-generated prefixes (S4), and GT-prefix evaluation with previous intermediate answers replaced by ground truth (S5).
The gaps measure the effect of generic prompting
($\Delta_{\mathrm{step}}$), explicit decomposition
($\Delta_{\mathrm{decomp}}$), sequential execution
($\Delta_{\mathrm{iter}}$), and oracle correction of intermediate visual states
($\Delta_{\mathrm{state}}$).
}
\label{tab:reasoning_intervention}
\resizebox{\textwidth}{!}{%
\begin{tabular}{llcccccccccc}
\toprule
Type & Model
& \begin{tabular}{@{}c@{}}Direct QA \\ (S0)\end{tabular}
& \begin{tabular}{@{}c@{}}Generic CoT \\ (S1)\end{tabular}
& \begin{tabular}{@{}c@{}}Schema-CoT \\ (S2)\end{tabular}
& \begin{tabular}{@{}c@{}}Few-shot GT-CoT \\ (S3)\end{tabular}
& \begin{tabular}{@{}c@{}}Pred-Step \\ (S4)\end{tabular}
& \begin{tabular}{@{}c@{}}GT-Prefix \\ (S5)\end{tabular}
& $\Delta_{\mathrm{text}}$
& $\Delta_{\mathrm{schema}}$
& $\Delta_{\mathrm{iter}}$
& $\Delta_{\mathrm{GT}}$ \\
\midrule
\multirow{3}{*}{Closed}
& GPT-5.5~\cite{openai2026gpt55}
& 44.6 & 45.3 & 42.7 & 47.2 & 38.8 & 89.2 & \cellcolor{lightyellow}+0.6 & \cellcolor{lightblue}-1.9 & \cellcolor{lightblue}-3.9 & \cellcolor{lightyellow}+50.4 \\
& GPT-5.4~\cite{openai2026gpt54}
& 32.8 & 34.3 & 26.5 & 30.0 & 28.4 & 86.0
& \cellcolor{lightyellow}+1.5
& \cellcolor{lightblue}-6.3
& \cellcolor{lightyellow}+1.9
& \cellcolor{lightyellow}+57.6 \\
& Gemini-3.1-Pro~\cite{googledeepmind2026gemini31pro}
& 40.6 & 39.1 & 45.5 & 34.1 & 41.5 & 89.6
& \cellcolor{lightblue}-1.5
& \cellcolor{lightyellow}+4.9
& \cellcolor{lightblue}-4.0
& \cellcolor{lightyellow}+48.1 \\
\midrule
\multirow{4}{*}{Open-Large}
& Qwen3.5-27B~\cite{qwen2026qwen35}
& 33.8 & 39.2 & 40.5 & 44.2 & 41.4 & 53.6
& \cellcolor{lightyellow}+5.4
& \cellcolor{lightyellow}+6.7
& \cellcolor{lightyellow}+0.9
& \cellcolor{lightyellow}+12.2 \\
& InternVL3.5-38B~\cite{wang2025internvl35}
& 28.4 & 31.7 & 31.7 & 34.5 & 27.2 & 80.8
& \cellcolor{lightyellow}+3.3
& \cellcolor{lightyellow}+3.3
& \cellcolor{lightblue}-4.5
& \cellcolor{lightyellow}+53.6 \\
& Gemma-4-26B~\cite{google2026gemma4modelcard}
& 30.2 & 25.6 & 31.2 & 32.8 & 27.2 & 88.1
& \cellcolor{lightblue}-4.6
& \cellcolor{lightyellow}+1.0
& \cellcolor{lightblue}-4.0
& \cellcolor{lightyellow}+60.9 \\
& GLM-4.5V~\cite{glmvteam2025glm45v}
& 27.3 & 27.5 & 28.8 & 30.1 & 24.3 & 85.4
& \cellcolor{lightyellow}+0.2
& \cellcolor{lightyellow}+1.5
& \cellcolor{lightblue}-4.5
& \cellcolor{lightyellow}+61.1 \\
\midrule
\multirow{4}{*}{Open-Small}
& DeepSeek-VL2~\cite{wu2024deepseekvl2}
& 26.4 & 26.2 & 23.0 & 18.2 & 25.0 & 72.3
& \cellcolor{lightblue}-0.2
& \cellcolor{lightblue}-3.4
& \cellcolor{lightyellow}+2.0
& \cellcolor{lightyellow}+47.3 \\
& MiniCPM-V 4.5~\cite{yu2025minicpmv45}
& 27.3 & 23.8 & 20.9 & 25.9 & 27.5 & 78.7
& \cellcolor{lightblue}-3.5
& \cellcolor{lightblue}-6.4
& \cellcolor{lightyellow}+6.6
& \cellcolor{lightyellow}+51.2 \\
& Phi-4-RV-15B~\cite{aneja2026phi4reasoningvision}
& 29.5 & 26.0 & 27.3 & 27.5 & 29.3 & 87.4
& \cellcolor{lightblue}-3.5
& \cellcolor{lightblue}-2.2
& \cellcolor{lightyellow}+2.0
& \cellcolor{lightyellow}+58.1 \\
& Kimi-VL-A3B-Thinking~\cite{kimiteam2025kimivl}
& 27.5 & 25.8 & 25.1 & 23.6 & 36.3 & 80.2
& \cellcolor{lightblue}-1.7
& \cellcolor{lightblue}-2.4
& \cellcolor{lightyellow}+11.2
& \cellcolor{lightyellow}+43.9 \\
\midrule
\multicolumn{2}{l}{Mean}
& 30.4 & 29.9 & 30.1 & 30.1 & 30.8 & 80.2
& \cellcolor{lightblue}-0.5
& \cellcolor{lightblue}-0.3
& \cellcolor{lightyellow}+0.8
& \cellcolor{lightyellow}+49.4 \\
\bottomrule
\end{tabular}
}
\vspace{-0.3cm}
\end{table*}

\paragraph{Reasoning-format evaluation.}
Table~\ref{tab:reasoning_intervention} compares S0--S5 as controlled contrasts rather than independent prompting variants.
S1 is slightly below S0 on average, suggesting that a generic ``think step by step'' instruction does not reliably improve ultra-resolution reasoning.
S2 is nearly identical to S1, and S3 has the same mean performance as S2, indicating that explicit GT-CoT schemas and few-shot GT trajectories change the reasoning format but do not substantially improve the model's ability to instantiate correct visual facts.
We therefore use S2 as the canonical schema-based setting for operation-level comparison.
S4 is only slightly above S2, showing that sequential execution provides limited benefit when later steps depend on previous model-predicted states.
In contrast, S5 reaches 80.2\% average accuracy and improves over S4 by +49.4 points.
This controlled contrast separates visual-state acquisition from downstream inference: many final-answer errors become recoverable once the required intermediate visual facts are supplied.
Overall, the reasoning-format evaluation shows that current VLMs are less limited by the absence of an explicit reasoning format than by their inability to obtain reliable intermediate visual states from ultra-resolution images.

\paragraph{Operation-level evaluation.}
The GT-CoT annotations further allow UltraVR to localize which operations produce these incorrect intermediate states, as shown in Figure~\ref{fig:operation_analysis}.
We first analyze S2, where the model completes the full GT-CoT schema in a single response and therefore avoids iterative error propagation from previous predicted steps.
Figure~\ref{fig:operation_analysis}(a) reports relative operation difficulty by normalizing each model's operation error against its own mean step error.
PER is consistently harder than the model average, while later operations such as INT are often easier, indicating that explicit decomposition alone does not resolve local visual-state formation.
We then analyze S5, where previous intermediate answers are replaced with ground truth to isolate the current operation from earlier accumulated errors.
Figure~\ref{fig:operation_analysis}(b) shows that GND and especially PER remain weaker even under corrected prefixes, whereas downstream operations such as QUA, INT, and INF become highly accurate once the required prior visual facts are supplied.
Finally, Figure~\ref{fig:operation_analysis}(c) analyzes S4, where the model rolls out the trajectory sequentially with its own predicted prefixes.
Most trajectories first fail at GND or PER; once these early visual-state operations are correct, the remaining trajectory is usually completed successfully.
Together, Table~\ref{tab:reasoning_intervention} and Figure~\ref{fig:operation_analysis} demonstrate the process-level diagnostic utility of UltraVR: the benchmark not only measures final-answer errors, but localizes them to specific operations in the visual-to-decision chain, with current VLM failures concentrated in evidence grounding and local perception.

\section{Discussion}
\label{sec:discussion}
UltraVR is designed to diagnose where ultra-resolution visual reasoning breaks beyond final-answer accuracy.
Our results indicate that the dominant bottleneck is not simply language-side reasoning, but unreliable visual-fact acquisition from large images: models often recover when intermediate visual facts are corrected, yet fail to ground sparse evidence or perceive fine local details in the first place.
This suggests that future ultra-resolution VLMs may require stronger visual search, adaptive evidence selection, multi-scale memory, and uncertainty-aware grounding, rather than relying only on larger visual budgets or generic CoT prompts.
UltraVR is a diagnostic benchmark rather than an exhaustive representation of all ultra-resolution applications.
Its domains and templates isolate representative bottlenecks across CCTV, RS, WSI, and AD under controlled, evidence-grounded reasoning tasks.
Future work can extend this framework to additional domains, open-ended outputs, interactive tool use, and human-model collaborative workflows.


\newpage
\bibliographystyle{unsrt}
\bibliography{bibliography}

\newpage
\appendix
\numberwithin{table}{section}
\counterwithin{figure}{section}
\startcontents[appendix]
\printcontents[appendix]{l}{1}{\section*{Appendix Contents}}
\newpage

\section{Detailed Benchmark Construction}
\label{app:benchmark_construction}

This appendix provides the domain-specific construction details for UltraVR. The main text describes the benchmark at a high level; here we expand the data sources, evidence-candidate extraction, question construction, GT-CoT annotation, distractor design, and quality-control procedures for each domain.

\subsection{CCTV Surveillance}
\label{app:cctv_construction}

\paragraph{Data source and image selection.}
We build the CCTV split from the gigapixel human-centric PANDA dataset~\citep{wang2020panda}, which provides wide-field surveillance frames captured by static camera arrays. We select scenes from campus canteens, plazas, intersections, and crowded streets. These scenes contain dense pedestrian crowds, small distant people, occlusions, person-vehicle interactions, and long-range spatial layouts. Such properties make them suitable for ultra-resolution reasoning: aggressive downsampling removes small people and local attributes, while naive tiling can break global spatial relations.

\paragraph{Evidence candidates.}
We use released person and vehicle annotations as the primary evidence source. Person annotations include full-body, visible-body, and head boxes, together with attributes such as age group, pose, and rider status. We normalize boxes into image coordinates and assign each box to a quadrant based on its center. We then form semantic object bundles, such as adults riding bicycles, children being carried, standing adults, or people near vehicles. For relational tasks, we additionally describe selected people using visual cues such as clothing color, headwear, carried objects, approximate quadrant location, and facing direction. These cues are used to construct unambiguous references to target people and to support relative-position reasoning.

\paragraph{Question construction.}
CCTV questions are organized into comparison, logical verification, and relational inference. Comparison questions pair two semantic bundles and ask which option correctly describes the relationship between their counts, such as the absolute difference, total count, reduced ratio, or ordering relation. Logical verification questions combine a count claim and a region-extremum claim for the same bundle, requiring the model to verify both global and quadrant-level evidence. Relational inference questions select two visually describable people and ask for the relative position of one person with respect to the other using a four-way label, such as front-left, front-right, back-left, or back-right. These questions require grounding small people, perceiving local visual cues, counting or comparing groups, and integrating spatial relations.

\paragraph{GT-CoT construction.}
Each CCTV question is paired with a fixed-schema GT-CoT trace. For comparison, the trace grounds each object bundle, counts its instances, computes the total, difference, ratio, and ordering, and maps the result to the supported option. For logical verification, the trace lists per-quadrant evidence, counts the relevant instances in each region, aggregates the global count, identifies the quadrant with the maximum count, checks each claim, and selects the final truth pattern. For relational inference, the trace grounds Person A and Person B, perceives their facing directions when needed, derives front/behind and left/right relations from box centers and orientations, and integrates them into the final four-way relation. Intermediate answers are typed as box lists, integers, categorical attributes, spatial relations, or option labels.

\paragraph{Distractors and quality control.}
Distractors are generated from false but structurally plausible alternatives. For comparison, we use nearby count differences, alternative totals, reversed orderings, and incorrect ratios. For logical verification, distractors correspond to the remaining count-claim and region-claim truth patterns. For relational inference, distractors are the other relative-position labels. We automatically verify that every count matches the corresponding box list, that quadrant assignments are stable, and that the final option is consistent with the intermediate answers. We reject samples with ambiguous bundles, ties that would make a region-extremum claim unstable, low-confidence facing directions, or text-only shortcuts in the options. Final samples are further checked for visual answerability and reference clarity.

\subsection{Remote Sensing}
\label{app:rs_construction}

\paragraph{Data source and image selection.}
We construct the RS split from DOTA~\citep{xia2018dota}, an aerial image dataset with oriented object annotations over categories such as planes, ships, vehicles, bridges, harbors, storage tanks, roundabouts, swimming pools, and sports fields. We retain images with spatially distributed targets in urban blocks, harbors, airports, and sports complexes. The selected images require models to reason over small objects while preserving global layout, making them sensitive to downsampling and patch fragmentation.

\paragraph{Evidence candidates.}
We parse the annotation files into normalized bounding boxes and category labels. Each object is assigned to a quadrant by its box center. For each quadrant, we maintain category-specific box lists, object counts, and orientation labels derived from object geometry. We then construct candidate pools for the three reasoning categories. Comparison candidates pair regions and object types whose conditioned counts can produce meaningful numeric differences. Logical verification candidates identify region pairs and orientation-restricted objects that yield clean presence or absence patterns. Relational candidates select region pairs with unique extreme objects, such as the leftmost, rightmost, topmost, or bottommost instance of a category.

\paragraph{Question construction.}
Comparison questions ask for the absolute difference between object counts in two regions, often with landmark or local-context restrictions that force the model to ground both regions before counting. Logical verification questions ask whether orientation-restricted instances of an object category appear in specified regions, producing a four-way presence or absence decision. Relational inference questions select an extreme object in each of two regions and ask for their cross-axis spatial relation, such as north versus south or west versus east. These templates require models to preserve tiny objects, distinguish region-level evidence, and compare distant spatial locations.

\paragraph{GT-CoT construction.}
The RS GT-CoT traces follow the structure of the task. For comparison, the trace grounds the landmark or region in the first quadrant, enumerates and counts the conditioned objects, repeats the procedure for the second quadrant, and computes the final difference. For logical verification, the trace lists and counts the orientation-filtered objects in both regions and maps the observed pattern to the correct option. For relational inference, the trace verifies object existence in both regions, locates the selected extreme object in each region, compares the box centers along the required axis, and derives the final spatial relation. All intermediate box lists are subsets of the precomputed annotation-derived candidates, allowing automatic consistency checks.

\paragraph{Distractors and quality control.}
RS distractors are derived from task structure. Comparison distractors are numeric perturbations around the true difference. Logical verification distractors correspond to incorrect region-presence patterns. Relational distractors include the reversed spatial relation and options indicating missing objects in one of the regions. We automatically verify that object lists are region-specific, counts match the lists, selected extreme objects are unique, and the final answer follows from the intermediate steps. We remove candidates with ambiguous orientations, broad or non-localizable landmarks, near-boundary extreme objects, or options that can be guessed from language priors without inspecting the image.

\subsection{Whole-Slide Imaging}
\label{app:wsi_construction}

\paragraph{Data source and slide selection.}
We construct the WSI split from hematoxylin and eosin stained tumor whole-slide images in the TCGA-BRCA cohort~\citep{weinstein2013cancer}. WSI differs from object-centric domains because diagnostic evidence is distributed over gigapixel tissue slides and is interpreted through pathology morphology rather than dense object boxes. We select slides whose pathology reports contain morphology-related findings that can in principle be grounded in visible tissue, such as high-grade morphology, invasive carcinoma, in-situ or DCIS components, histologic subtype, focality, lymphovascular invasion, and margin-related findings. Slides are excluded when the report evidence is purely administrative, molecular, or not visually assessable.

\paragraph{Evidence candidates.}
Report-derived findings are normalized into slide-level diagnostic attributes and used to form candidate interpretation targets. Region-level evidence is organized through expert-reviewed ROIs on a $6\times6$ whole-slide grid, enabling the trace to refer to meaningful slide regions without requiring dense pixel-level annotation. The evidence pool includes tumor-rich regions, invasive components, in-situ components, tumor-stroma interfaces, margin or edge regions, high-atypia hotspots, high-cellularity regions, and benign reference tissue. Within selected ROIs, annotators record morphologic cues such as nuclear atypia, pleomorphism, hyperchromasia, mitotic activity, atypical mitoses, gland or lumen formation, necrosis, infiltrative boundaries, stromal or desmoplastic reaction, architectural irregularity, loss of polarity, cell cohesion or discohesion, mucin production, and suspicion of lymphovascular invasion.

\paragraph{Question construction.}
WSI questions are organized into comparison, logical verification, and relational inference. Comparison questions contrast clinically plausible interpretations along the same diagnostic axis, such as invasive ductal versus invasive lobular carcinoma, purely invasive versus mixed invasive-and-in-situ disease, convincing in-situ component versus no convincing in-situ component, or lymphovascular invasion present versus absent. Logical verification questions ask whether an integrated slide-level summary is supported by the visible tissue, where distractors differ by one key attribute such as grade, in-situ status, invasive status, focality, lymphovascular invasion, or margin status. Relational inference questions ask about relationships between pathological components, such as whether the invasive component is dominant, whether in-situ disease is secondary or substantial, whether both components coexist, or whether the relationship cannot be determined confidently.

\paragraph{GT-CoT construction.}
WSI GT-CoT traces follow a structured diagnostic workflow rather than a free-form explanation. The trace begins with low-power screening to identify broad regions and components, including tumor-rich areas, invasive and in-situ components, tumor-stroma interfaces, margins, and benign reference tissue. It then determines whether closer ROI review is required and selects diagnostic ROIs from the $6\times6$ grid. The trace specifies the magnification level needed for closer inspection and lists fine-grained morphology features to check. When appropriate, it asks whether standardized pathology criteria should be consulted, such as Nottingham grading, DCIS grading, margin assessment, or lymphovascular invasion criteria. The final steps follow an evidence-reason-conclusion structure: observed morphologic clues are mapped to intermediate interpretations, such as cytologic malignancy, architectural abnormality, infiltrative growth, glandular differentiation, high-grade morphology, or malignant morphologic pattern, and then integrated into the slide-level conclusion. Intermediate answers include ROI lists, magnification choices, morphology feature lists, criterion selections, categorical interpretations, and diagnostic conclusions.

\paragraph{Expert review, distractors, and quality control.}
WSI distractors are designed as clinically plausible near-miss descriptions rather than artificial negatives. For comparison, distractors reverse the preferred interpretation or alter the confidence level while keeping both alternatives medically meaningful. For logical verification, distractors flip one core attribute in an otherwise plausible summary. For relational inference, distractors encode alternative component relationships, such as dominant-secondary, co-substantial, absent additional component, or indeterminate relation. Three pathology experts review the question design, GT-CoT decomposition, answer labels, and distractor plausibility. Samples are revised or removed when the evidence is ambiguous, when the reasoning path is inconsistent with pathology workflow, or when an option can be rejected without visual assessment.

\subsection{Anomaly Detection}
\label{app:ad_construction}

\paragraph{Data source and image composition.}
We construct the AD split from MVTec LOCO AD~\citep{bergmann2022mvtecloco}, which contains normal samples, structural anomalies, and logical anomalies. We select anomaly samples from four object categories: breakfast box, juice bottle, pushpins, and screw bag. To simulate ultra-resolution industrial inspection, each benchmark image is composed as a grid with 99 normal cells and one anomalous cell. The anomalous cell is randomly placed, and the model must identify its row and column. The resulting composite images can reach resolutions such as $16{,}000\times12{,}800$, requiring exhaustive search over many visually similar cells.

\paragraph{Evidence candidates.}
Normality descriptions are adapted from prior anomaly detection studies~\citep{bergmann2022mvtecloco}. Each normality description is decomposed into checkable visual conditions, such as the presence of required components, the correct number of objects, expected spatial arrangement, or the absence of structural defects. We manually label which conditions are satisfied or violated for each abnormal sample. Every anomaly included in UltraVR violates at least one normality condition, ensuring that the answer is grounded in a visible local or structural irregularity.

\paragraph{Question construction.}
Each AD instance is generated from a category-specific template consisting of the inspection question, the normality description, and a constrained response instruction. The question asks the model to identify the anomalous cell in the grid. The normality description specifies the expected normal pattern for the product category, such as the required contents of a breakfast box or the expected arrangement of objects in a bag. The final answer is the grid coordinate of the cell that violates the normality description.

\paragraph{GT-CoT construction.}
For GT-CoT construction, the decomposed normality conditions are converted into step questions. Each step verifies a local or structural condition, such as whether a required component is present, whether a count is correct, whether the arrangement follows the expected rule, or whether a visible defect violates the global pattern. The final step identifies the cell that violates at least one normality condition. This trace reflects the intended inspection process: scan the repetitive grid, compare cells against the normality rule, and localize the violation.

\paragraph{Quality control.}
Because AD uses a constrained coordinate answer rather than multiple-choice distractors, quality control focuses on uniqueness and trace consistency. We verify that each composed grid contains exactly one anomalous cell, that the anomaly location matches the final coordinate answer, and that the GT-CoT steps correctly identify the violated condition. We also check that the anomaly is visually observable at the composed image resolution and that no normal cell accidentally violates the same normality rule.

\section{Details of Methodology}

\subsection{Comparison with Existing Benchmarks}
\label{app:benchmark_comparison}

Table~\ref{tab:benchmark_comparison} summarizes the positioning of UltraVR relative to representative VLM, high-resolution, and large-image reasoning benchmarks.
We compare benchmarks along six dimensions: ultra-resolution input, domain coverage, evidence-grounded reasoning, step-level ground truth, operation labels, and process-level diagnosis.

\begin{table*}[t]
\centering
\scriptsize
\setlength{\tabcolsep}{3.0pt}
\renewcommand{\arraystretch}{1.02}
\caption{
\textbf{Comparison with existing VLM benchmarks.}
UltraVR combines ultra-resolution inputs, multi-domain evidence-grounded reasoning, structured GT-CoT annotations, operation labels, and process-level diagnostic evaluation.
}
\label{tab:benchmark_comparison}
\begin{tabular}{lcccccc}
\toprule
\textbf{Benchmark} 
& \textbf{UHR} 
& \textbf{Multi-dom.} 
& \textbf{Evid.} 
& \textbf{Step GT} 
& \textbf{Op. label} 
& \textbf{Proc. diag.} \\
\midrule
MME~\citep{fu2023mme}               
& -- & \checkmark & -- & -- & -- & -- \\
MMBench~\citep{liu2023mmbench}      
& -- & \checkmark & -- & -- & -- & -- \\
MMMU~\citep{yue2023mmmu}            
& -- & \checkmark & \checkmark & -- & -- & -- \\
REASONMAP~\citep{feng2025reasonmap} 
& -- & -- & \checkmark & -- & -- & -- \\
Pix2Fact~\citep{jiang2026pix2fact}  
& -- & \checkmark & \checkmark & -- & -- & -- \\
MIRA~\citep{zhou2025mira}           
& -- & \checkmark & \checkmark & \checkmark & -- & \checkmark \\
\midrule
V*~\citep{wu2024vstar}              
& \checkmark & -- & -- & -- & -- & -- \\
HR-Bench~\citep{wang2024hrbench}    
& \checkmark & \checkmark & -- & -- & -- & -- \\
GigaGrounding~\citep{ma2024gigagrounding} 
& \checkmark & -- & \checkmark & -- & -- & -- \\
HRScene~\citep{zhang2025hrscene}    
& \checkmark & \checkmark & -- & -- & -- & -- \\
\midrule
XLRS-Bench~\citep{wang2025xlrsbench} 
& \checkmark & -- & \checkmark & -- & -- & -- \\
RSHR-Bench~\citep{dang2025rshrbench} 
& \checkmark & -- & \checkmark & -- & -- & -- \\
UR-Bench~\citep{li2026urbench}       
& \checkmark & \checkmark & \checkmark & -- & -- & -- \\
\midrule
\textbf{UltraVR} 
& \checkmark & \checkmark & \checkmark & \checkmark & \checkmark & \checkmark \\
\bottomrule
\end{tabular}

\vspace{2pt}
\begin{flushleft}
\scriptsize
\textit{UHR}: ultra-resolution input; \textit{Multi-dom.}: multi-domain coverage; 
\textit{Evid.}: evidence-grounded reasoning; \textit{Step GT}: step-level ground-truth annotation; 
\textit{Op. label}: operation labels; \textit{Proc. diag.}: process-level diagnostic evaluation.
\end{flushleft}
\vspace{-6pt}
\end{table*}

\subsection{Quality Control}
We apply quality-control checks at both the QA level and the GT-CoT level.
\textbf{Visual answerability.}
We verify that each question can be answered from the provided image and the specified evidence metadata.
Questions requiring unavailable information or ambiguous visual evidence are removed or revised.
\textbf{Option ambiguity.}
We ensure that each multiple-choice item has a single correct answer.
Distractors are designed to be plausible but visually or logically incorrect, rather than arbitrary answer choices.
\textbf{Evidence visibility.}
We check whether the answer-relevant evidence is present at its original resolution and whether the task requires non-trivial ultra-resolution processing.
For oracle-evidence settings, cropped evidence is preserved at its original resolution before being provided to the model.
\textbf{GT-CoT consistency.}
We verify that each intermediate answer is consistent with the final answer and that the step sequence forms a valid reasoning path.
For step-level evaluation, each GT-CoT step must have a well-defined expected answer.
\textbf{Text-only leakage.}
We check whether the question and answer options contain linguistic shortcuts that could reveal the answer without image inspection.
When a sample is answerable from text alone, we revise or discard it.
This check is especially important for multiple-choice QA, where answer-option artifacts may otherwise inflate performance.
\textbf{Domain-specific review.}
For WSI, 3 pathologist participated in the validation to ensure that the visual evidence and diagnostic interpretation are medically meaningful. For AD, we verify that the inserted anomaly is unique within the grid and that the normality description correctly characterizes the non-anomalous cells. For CCTV and RS, we verify the final answer and object localization based on the model.

\section{Reasoning steps and operations}

Table~\ref{tab:op_definitions} defines the five reasoning operations used in UltraVR GT-CoT annotations.
Tables~\ref{tab:cctv_steps}--\ref{tab:ad_steps} list the step-to-operation assignments for each dataset and task type.
``---'' indicates no steps are assigned to that operation for that task type.

\begin{table}[ht]
\centering
\caption{Definitions of the five reasoning operations.}
\label{tab:op_definitions}
\small
\begin{tabular}{@{} l p{3cm} p{9cm} @{}}
\toprule
\textbf{Code} & \textbf{Name} & \textbf{Definition} \\
\midrule
GND & Evidence Grounding    & Grounds relevant visual evidence in image space, including objects, regions, quadrants, grid cells, ROIs, landmarks, or WSI areas that define the search space for subsequent reasoning. \\[4pt]
PER & Local Perception      & Recognizes local visual properties of grounded evidence, including object attributes, orientations, morphology, defect patterns, visual states, or fine-grained target appearances. \\[4pt]
QUA & Quantification        & Converts visual evidence into numerical or ordinal information, such as object counts, quantity estimates, size comparisons, differences, or ratios. \\[4pt]
INT & Evidence Integration  & Combines multiple pieces of evidence across objects, regions, or reasoning steps, including distant-region comparison, spatial-relation inference, regional aggregation, or multi-evidence synthesis. \\[4pt]
INF & Decision Inference    & Draws a task-level conclusion from integrated evidence by applying task-specific criteria, domain knowledge, logical conditions, pathology references, anomaly definitions, or answer-option rules. \\
\bottomrule
\end{tabular}
\end{table}

\begin{table}[ht]
\centering
\caption{CCTV dataset: step-to-operation mapping across task types.}
\label{tab:cctv_steps}
\small
\begin{threeparttable}
\begin{tabular}{@{} l p{2.8cm} p{4.0cm} p{3.0cm} @{}}
\toprule
\textbf{Op.} & \textbf{Comparison} & \textbf{Logical Verification} & \textbf{Relational Inference} \\
\midrule
GND & S1, S4          & S1, S3, S5, S7      & S1, S2 \\
PER & S2, S5          & ---                 & S3, S4 \\
QUA & S3, S6--S9      & S2, S4, S6, S8, S9  & ---    \\
INT & S10             & S10                 & S5, S6 \\
INF & S11             & S11--S13            & S7     \\
\bottomrule
\end{tabular}
\begin{tablenotes}
\footnotesize
\item \textit{Comparison:} S1/S4 acquire whether each semantic group exists; S2/S5 inspect and localize detailed target instances; S11 maps computed evidence to the supported claim family.
\item \textit{Logical verification:} S1/S3/S5/S7 acquire region-wise target coordinates; S10 identifies the region with the most instances; S11--S13 verify claim truth and select the supported evidence pattern.
\item \textit{Relational inference:} S1/S2 ground Person~A and Person~B; S3/S4 perceive facing direction; S7 applies the label-composition rule over front/behind and left/right spatial relations.
\end{tablenotes}
\end{threeparttable}
\end{table}

\begin{table}[ht]
\centering
\caption{RS dataset: step-to-operation mapping across task types.}
\label{tab:rs_steps}
\small
\begin{threeparttable}
\begin{tabular}{@{} l p{2.8cm} p{3.5cm} p{3.5cm} @{}}
\toprule
\textbf{Op.} & \textbf{Comparison} & \textbf{Logical Verification} & \textbf{Relational Inference} \\
\midrule
GND & S1, S5 & S1, S3 & S1, S3 \\
PER & S3, S7 & ---    & S2, S4 \\
QUA & S4, S8 & S2, S4 & ---    \\
INT & S2, S6 & ---    & S5     \\
INF & S9     & S5     & ---    \\
\bottomrule
\end{tabular}
\begin{tablenotes}
\footnotesize
\item \textit{Comparison:} S1/S5 acquire landmarks that define the relation predicates; S2/S6 evaluate object-landmark relation existence; S3/S7 localize fine-grained targets under landmark constraints; S9 applies the final absolute-difference decision.
\item \textit{Logical verification:} S5 is assigned exclusively to INF to apply the final logical condition over regional count/presence evidence, ensuring mutually exclusive sample-level trajectory decomposition.
\item \textit{Relational inference:} S1/S3 acquire whether each quadrant contains the target object; S2/S4 perceive/localize the selected instance (e.g., the leftmost ship); S5 compares bounding-box centers to infer the spatial relation.
\end{tablenotes}
\end{threeparttable}
\end{table}

\begin{table}[ht]
\centering
\caption{WSI dataset: step-to-operation mapping (shared template for all three task types).}
\label{tab:wsi_steps}
\small
\begin{tabular}{@{} l p{3cm} p{9cm} @{}}
\toprule
\textbf{Op.} & \textbf{Steps} & \textbf{Description} \\
\midrule
GND & Q1, Q2, Q2.1       & Ground global screening targets, closer-review needs, and selected ROIs in the WSI grid. \\[3pt]
PER & Q3, Q4             & Specify local inspection strategy and fine-grained morphologic features. \\[3pt]
QUA & ---                & \\[3pt]
INT & Q5, Q5.1, Q5.2     & Connect visual context with standardized pathology criteria and reproducibility needs. \\[3pt]
INF & Q6.1, Q6.2, Q6.3  & Apply pathology-oriented decision criteria to select evidence, infer interpretation, and reach the final conclusion. \\
\bottomrule
\end{tabular}
\end{table}

\begin{table}[ht]
\centering
\caption{AD dataset: step-to-operation mapping (shared template across all scenes).}
\label{tab:ad_steps}
\small
\begin{tabular}{@{} l l @{}}
\toprule
\textbf{Op.} & \textbf{Steps} \\
\midrule
GND & S1, S2, S3 \\
PER & S4--end    \\
QUA & ---        \\
INT & ---        \\
INF & ---        \\
\bottomrule
\end{tabular}
\begin{tablenotes}
\footnotesize
\item S1-S3 acquire the global grid/object-set structure (GND). S4 onward inspects local visual states or defect patterns of individual product instances (PER); the final step varies by scene: breakfast box (S15), juice bottle (S14), pushpins (S7), screw bag (S11).
\end{tablenotes}
\end{table}


\section{Additional Experimental Details}
\label{app:experimental_details}

\subsection{Implementation Details}
\label{app:implementation_details}

Open-weight model evaluation is implemented with the vLLM backend.
For each model, we use the official or commonly adopted preprocessing pipeline whenever available, including model-specific resizing, tiling, patch selection, and visual-token allocation.
Because different VLM families use different visual encoders and token compression mechanisms, the maximum visual budget and effective maximum resolution in Table~\ref{tab:main_results} should be interpreted as nominal implementation limits rather than directly comparable measures of visual capacity.
Unless otherwise specified, we use a shared decoding configuration with temperature $=0.7$, top-$p=0.95$, top-$k=20$, no repetition penalty, and a maximum output length of 32,768 tokens.
For closed-source models, we use the closest available configuration supported by each API.
All models are evaluated with the same question format, answer options, and evaluation scripts.

For V0, the image is removed and the prompt explicitly allows abstention when the answer cannot be determined from the provided text.
For V1, we externally downsample the full image to a global view before sending it to the model.
For V2, we provide the original image and allow each model to apply its native preprocessing.
For V3 and V4, all local evidence crops are preserved at their original resolution before being passed to the model; we do not encode the answer option in the crop name or prompt.
For S4 and S5, each intermediate step is queried as a separate model call, and the final answer is queried after the structured intermediate trajectory is completed.

\subsection{Answer Extraction}
\label{app:answer_extraction}

All benchmark tasks use a four-choice answer format.
For final-answer evaluation, we instruct the model to output the final option letter.
When the response contains additional reasoning, we apply a deterministic rule-based parser.
The parser first searches for explicit final-answer markers, such as ``final answer'', ``answer'', or their common variants, and then extracts the nearest valid option letter.
If no marker is found, the parser searches the response from the end to the beginning and uses the last unambiguous option letter.
If multiple incompatible option letters are presented without a final-answer marker, or if no valid option letter can be extracted, the response is marked as incorrect.
For V0, explicit abstentions such as ``not sure'' are counted as abstentions and are not treated as correct unless the benchmark item itself has an abstention option, which UltraVR does not use.

For step-level evaluation, each predicted intermediate answer is parsed according to the expected answer type of the corresponding GT-CoT step.
Choice-like intermediate answers are parsed with the same option-letter parser.
Short textual answers are normalized by lowercasing, trimming whitespace, removing trailing punctuation, and canonicalizing simple variants defined in the annotation metadata.
Numeric answers are parsed after removing commas and units when the unit is fixed by the step question.
If a response cannot be parsed into the expected format, the corresponding step is marked as incorrect.

\subsection{Metric Definitions}
\label{app:metric_definitions}

Let $\mathcal{D}$ denote the set of domains, where $\mathcal{D}=\{\mathrm{CCTV},\mathrm{RS},\mathrm{WSI},\mathrm{AD}\}$.
For a domain $d$, let $\mathcal{I}_d$ be its evaluation samples.
For sample $i$, let $y_i$ be the ground-truth final answer and $\hat{y}_{i}^{p}$ be the final answer predicted under protocol $p$.
The domain-level final-answer accuracy under protocol $p$ is
\begin{equation}
A_{d}^{p}=\frac{1}{|\mathcal{I}_d|}\sum_{i\in\mathcal{I}_d}\mathbf{1}\left[\hat{y}_{i}^{p}=y_i\right].
\end{equation}
The domain-balanced macro accuracy is
\begin{equation}
A_{\mathrm{macro}}^{p}=\frac{1}{|\mathcal{D}|}\sum_{d\in\mathcal{D}} A_{d}^{p}.
\end{equation}
For a category $c$ within domain $d$, category-level accuracy is computed analogously over $\mathcal{I}_{d,c}$.
The visual gap in Table~\ref{tab:main_results} is
\begin{equation}
\Delta_{\mathrm{visual}} = A_{\mathrm{macro}}^{\mathrm{V2/S0}} - A_{\mathrm{macro}}^{\mathrm{V0}},
\end{equation}
where V2/S0 denotes the full-image Direct QA setting.

For operation-level metrics, let $\mathcal{O}=\{\mathrm{GND},\mathrm{PER},\mathrm{QUA},\mathrm{INT},\mathrm{INF}\}$ be the set of operation labels.
Each sample $i$ has $T_i$ annotated intermediate steps.
For step $t$, let $z_{i,t}$ be the ground-truth step answer, $\hat{z}_{i,t}^{p}$ be the predicted step answer under protocol $p$, and $o_{i,t}\in\mathcal{O}$ be the operation label.
The operation-wise step accuracy for operation $o$ under protocol $p$ is first computed within each domain:
\begin{equation}
A_{d,o}^{p}=\frac{\sum_{i\in\mathcal{I}_d}\sum_{t=1}^{T_i}\mathbf{1}[o_{i,t}=o]\mathbf{1}[\hat{z}_{i,t}^{p}=z_{i,t}]}{\sum_{i\in\mathcal{I}_d}\sum_{t=1}^{T_i}\mathbf{1}[o_{i,t}=o]}.
\end{equation}
The reported operation-wise accuracy is then macro-averaged across domains that contain operation $o$:
\begin{equation}
A_{o}^{p}=\frac{1}{|\mathcal{D}_o|}\sum_{d\in\mathcal{D}_o} A_{d,o}^{p},\quad
\mathcal{D}_o=\left\{d:\sum_{i\in\mathcal{I}_d}\sum_{t=1}^{T_i}\mathbf{1}[o_{i,t}=o]>0\right\}.
\end{equation}
Figure~\ref{fig:operation_analysis}(b) reports $A_o^{\mathrm{S5}}$.
Figure~\ref{fig:operation_analysis}(a) reports the corresponding S2 step-error profile.
For each model and operation, we define
\begin{equation}
E_{o}^{\mathrm{S2}}=1-A_{o}^{\mathrm{S2}},\quad
\bar{E}^{\mathrm{S2}}=\frac{1}{|\mathcal{O}|}\sum_{o\in\mathcal{O}}E_{o}^{\mathrm{S2}},\quad
\delta_{o}^{\mathrm{S2}}=E_{o}^{\mathrm{S2}}-\bar{E}^{\mathrm{S2}}.
\end{equation}
The right-side number in Figure~\ref{fig:operation_analysis}(a) is $\bar{E}^{\mathrm{S2}}$, and each bar is $\delta_{o}^{\mathrm{S2}}$.
Thus, a negative value indicates an operation with lower-than-average step error for that model, while a positive value indicates higher-than-average step error.

For the first-error distribution in Figure~\ref{fig:operation_analysis}(c), each S4 trajectory is assigned to exactly one outcome.
Let
\begin{equation}
t_i^{\star}=\min\left\{t:\hat{z}_{i,t}^{\mathrm{S4}}\neq z_{i,t}\right\}.
\end{equation}
If such a step exists, the trajectory is assigned to operation $o_{i,t_i^{\star}}$.
If all intermediate steps are correct but the final answer is wrong, the trajectory is assigned to $\mathrm{Final}$.
If all intermediate steps and the final answer are correct, the trajectory is assigned to $\mathrm{NoErr}$.
For outcome $r\in\mathcal{O}\cup\{\mathrm{Final},\mathrm{NoErr}\}$, the first-error proportion is
\begin{equation}
P_{d,r}=\frac{1}{|\mathcal{I}_d|}\sum_{i\in\mathcal{I}_d}\mathbf{1}\left[\mathrm{outcome}(i)=r\right],\quad
P_{r}=\frac{1}{|\mathcal{D}|}\sum_{d\in\mathcal{D}}P_{d,r}.
\end{equation}
This macro-averaging reduces biases from dataset size, CoT length, and operation frequency.

\clearpage
\newpage

\section{Examples}

\vfill

\begin{figure}[!htb]
\centering
\includegraphics[width=\textwidth]{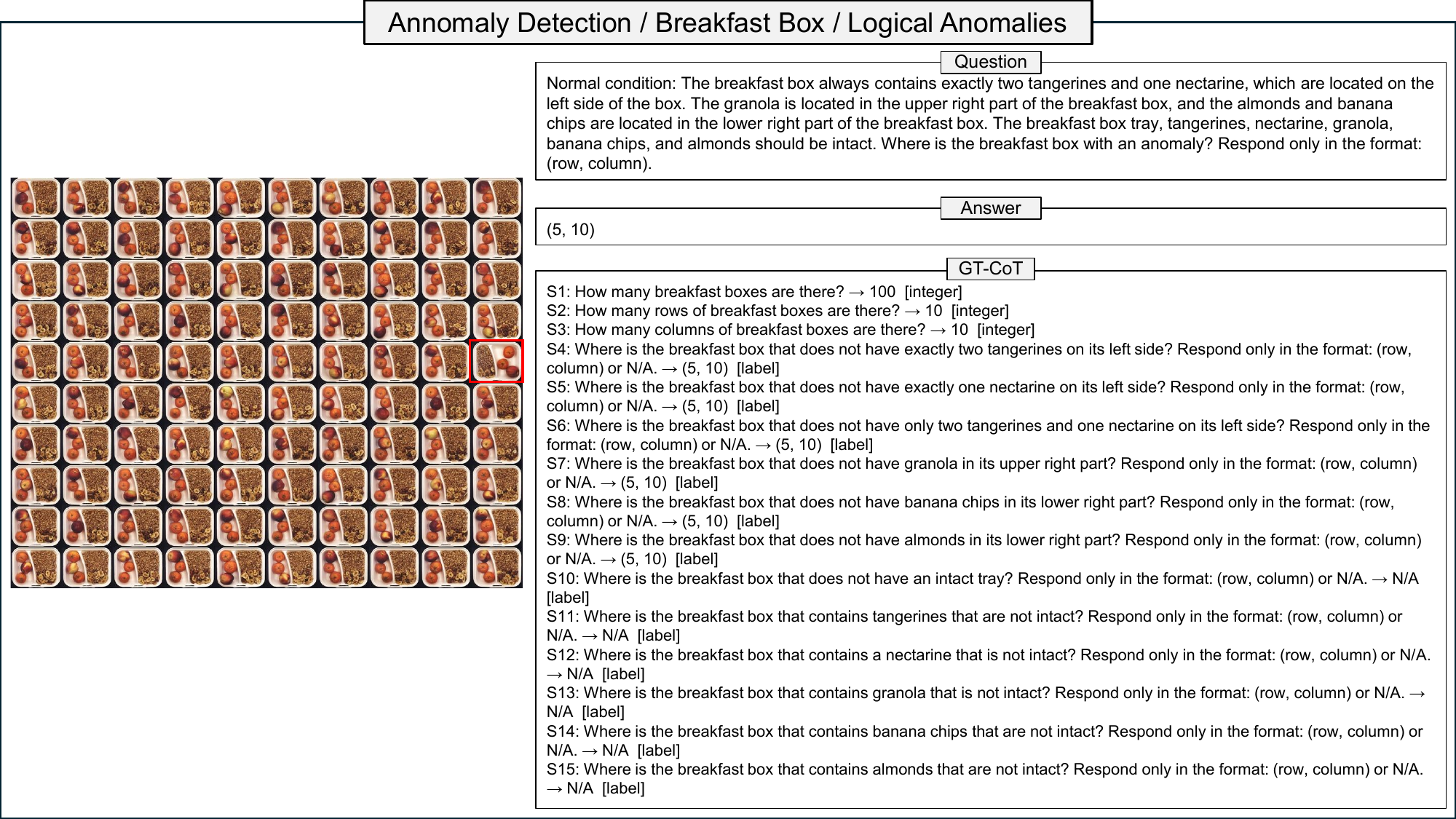}
\caption{Anomaly detection example: Breakfast Box -- Logical Anomalies.}
\label{fig:ex_ad_breakfast_box_logical}
\end{figure}

\vfill

\begin{figure}[!htb]
\centering
\includegraphics[width=\textwidth]{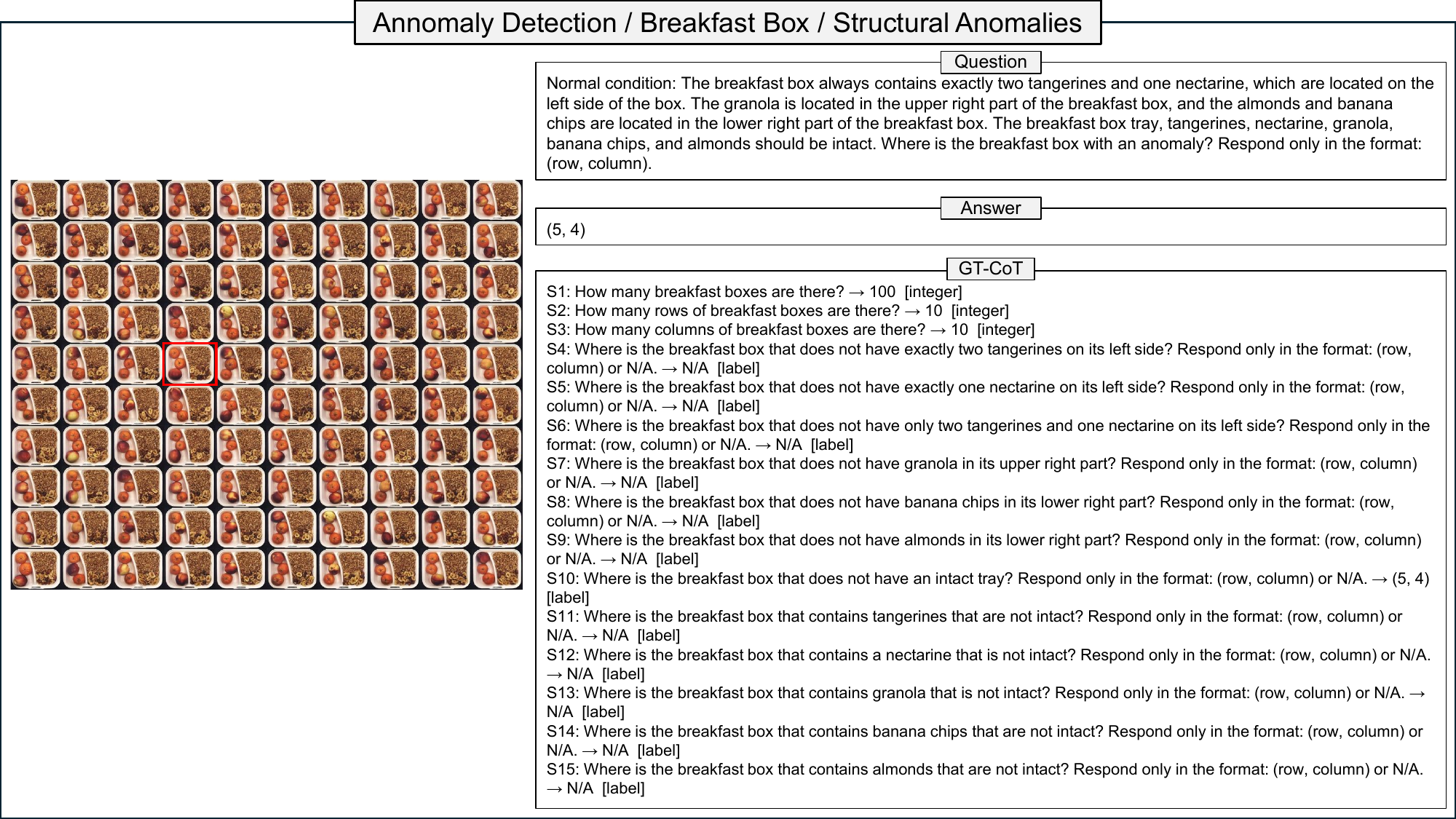}
\caption{Anomaly detection example: Breakfast Box -- Structural Anomalies.}
\label{fig:ex_ad_breakfast_box_structural}
\end{figure}

\vfill

\clearpage
\newpage

\vfill

\begin{figure}[!htb]
\centering
\includegraphics[width=\textwidth]{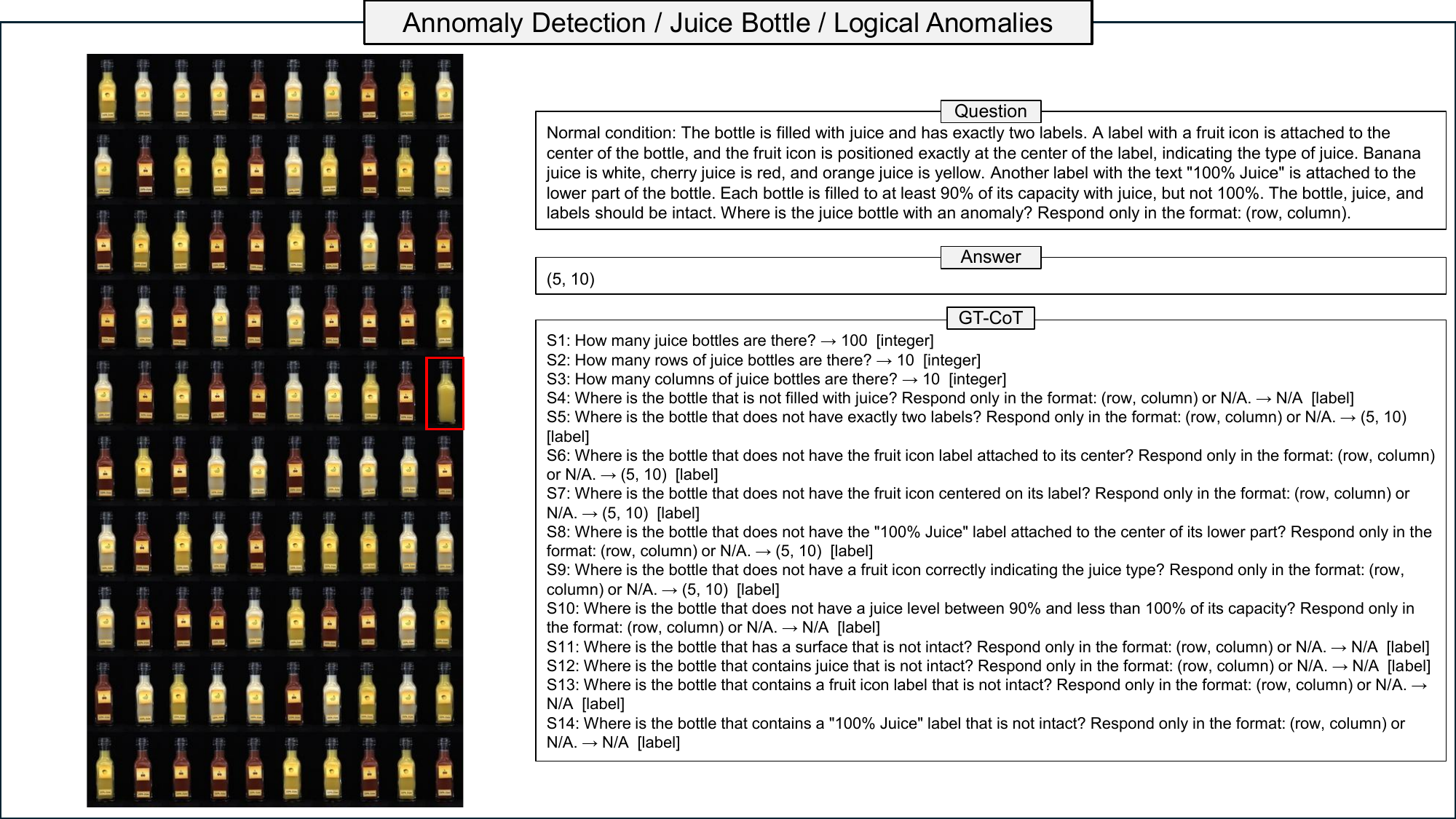}
\caption{Anomaly detection example: Juice Bottle -- Logical Anomalies.}
\label{fig:ex_ad_juice_bottle_logical}
\end{figure}

\vfill

\begin{figure}[!htb]
\centering
\includegraphics[width=\textwidth]{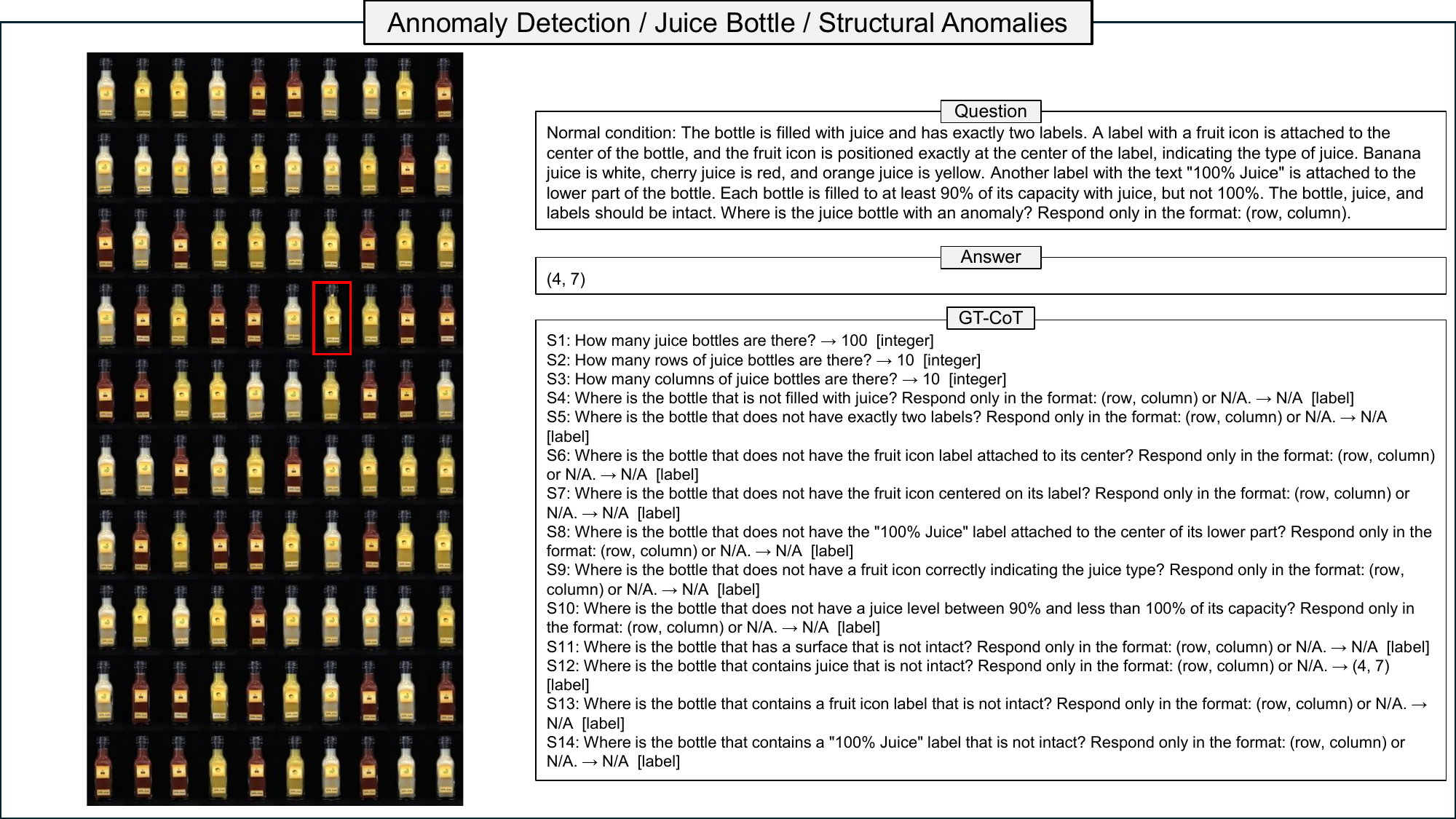}
\caption{Anomaly detection example: Juice Bottle -- Structural Anomalies.}
\label{fig:ex_ad_juice_bottle_structural}
\end{figure}

\vfill

\clearpage
\newpage

\vfill

\begin{figure}[!htb]
\centering
\includegraphics[width=\textwidth]{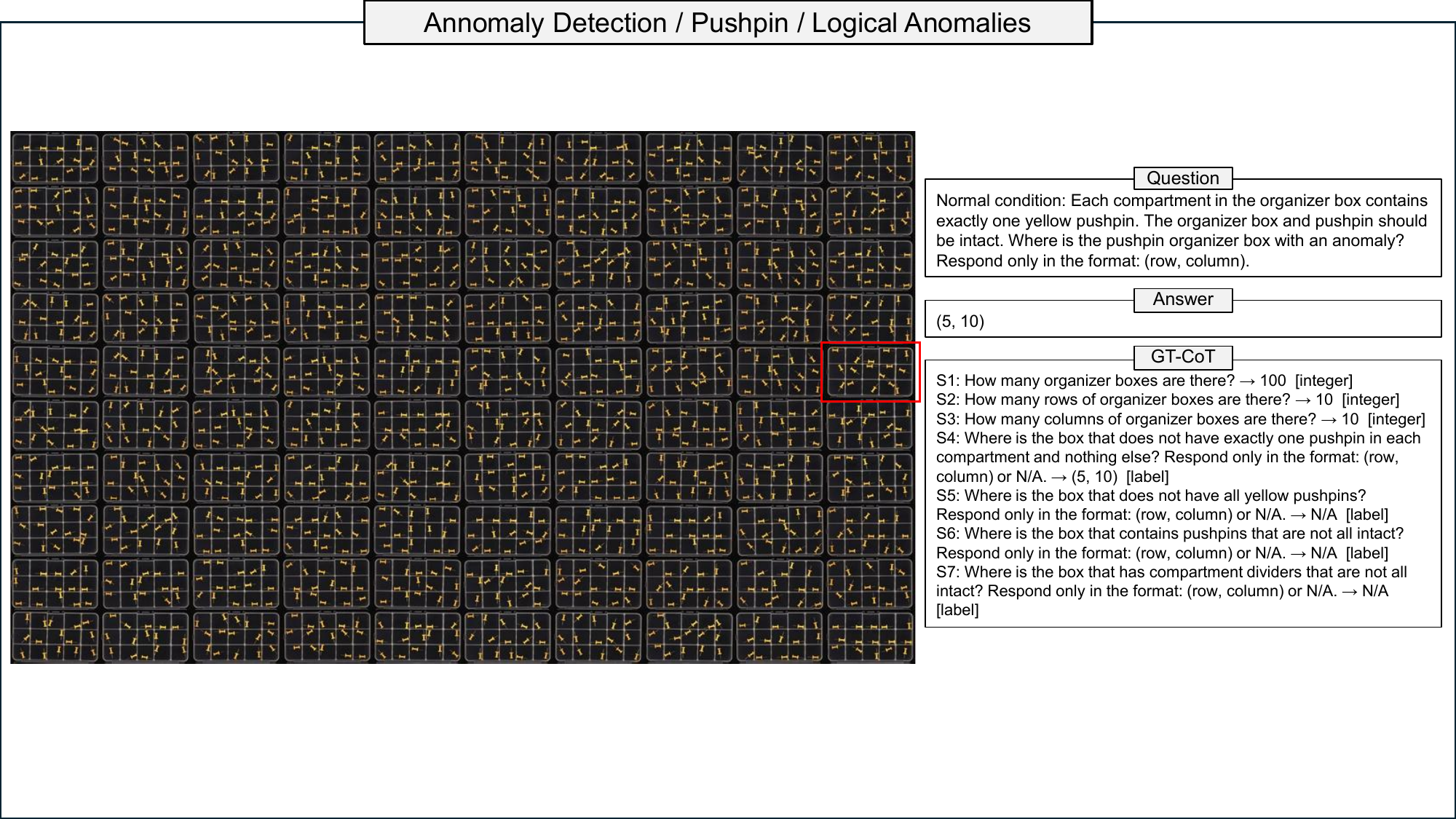}
\caption{Anomaly detection example: Pushpins -- Logical Anomalies.}
\label{fig:ex_ad_pushpins_logical}
\end{figure}

\vfill

\begin{figure}[!htb]
\centering
\includegraphics[width=\textwidth]{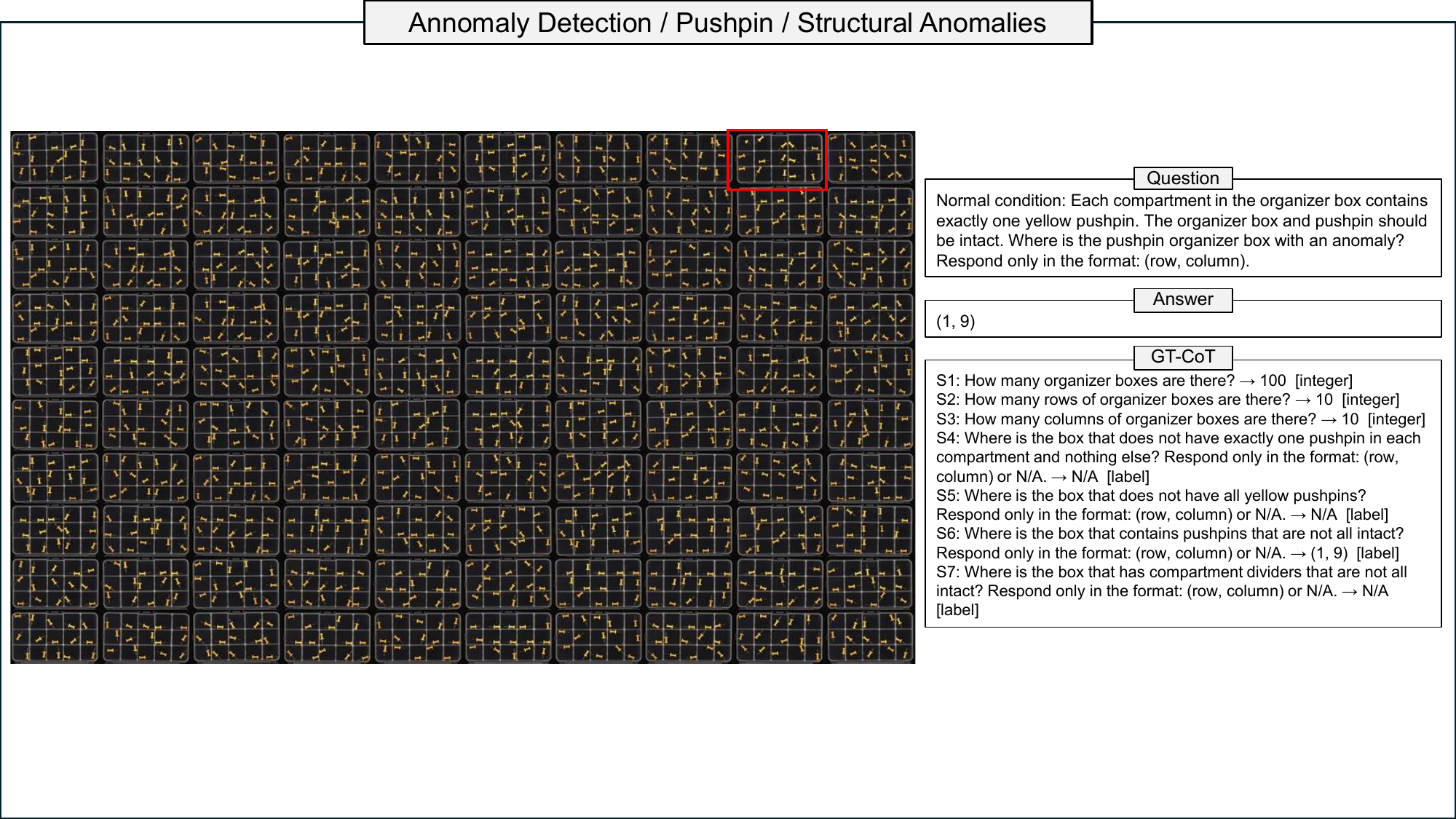}
\caption{Anomaly detection example: Pushpins -- Structural Anomalies.}
\label{fig:ex_ad_pushpins_structural}
\end{figure}

\vfill

\clearpage
\newpage

\vfill

\begin{figure}[!htb]
\centering
\includegraphics[width=\textwidth]{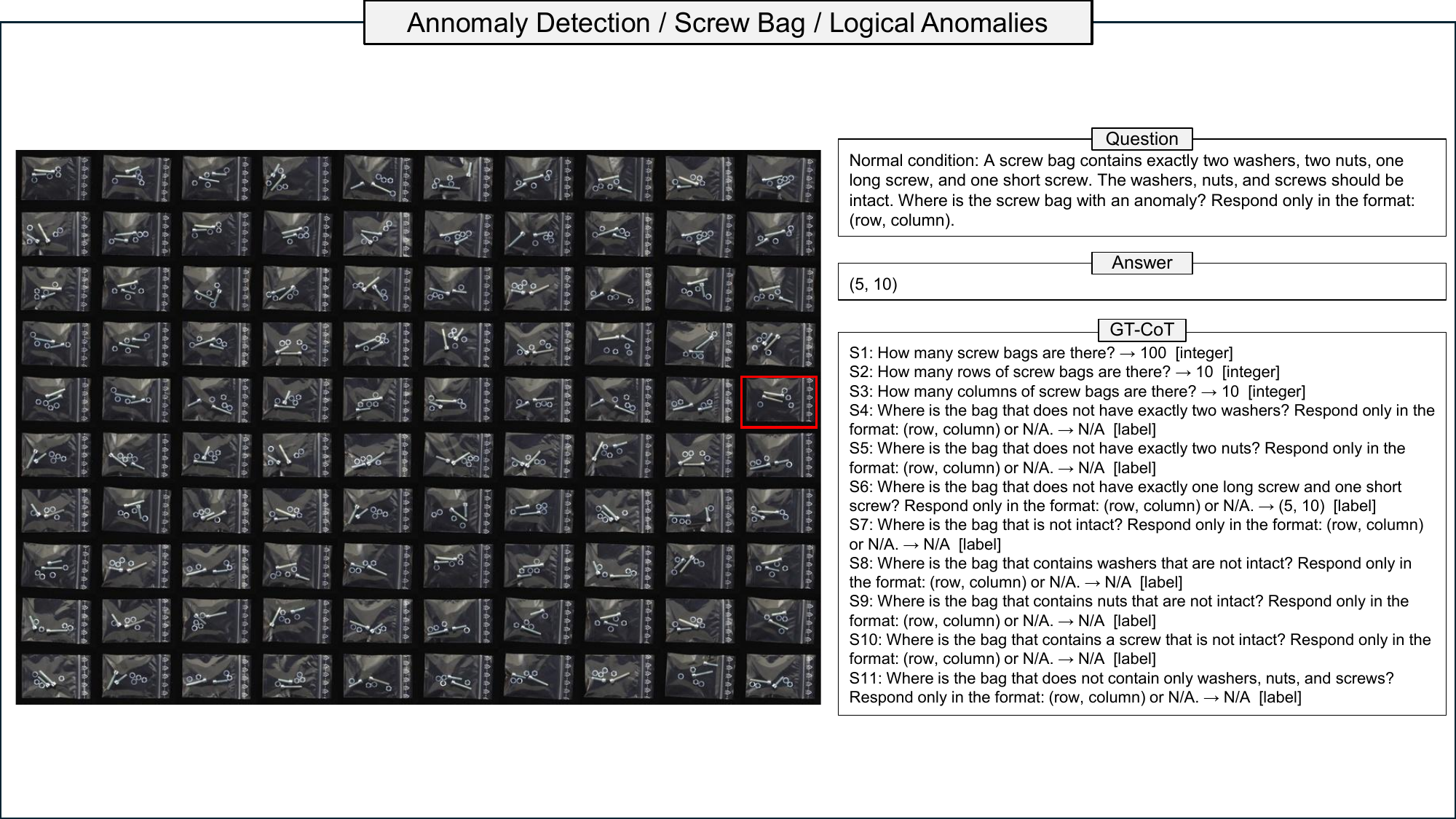}
\caption{Anomaly detection example: Screw Bag -- Logical Anomalies.}
\label{fig:ex_ad_screw_bag_logical}
\end{figure}

\vfill

\begin{figure}[!htb]
\centering
\includegraphics[width=\textwidth]{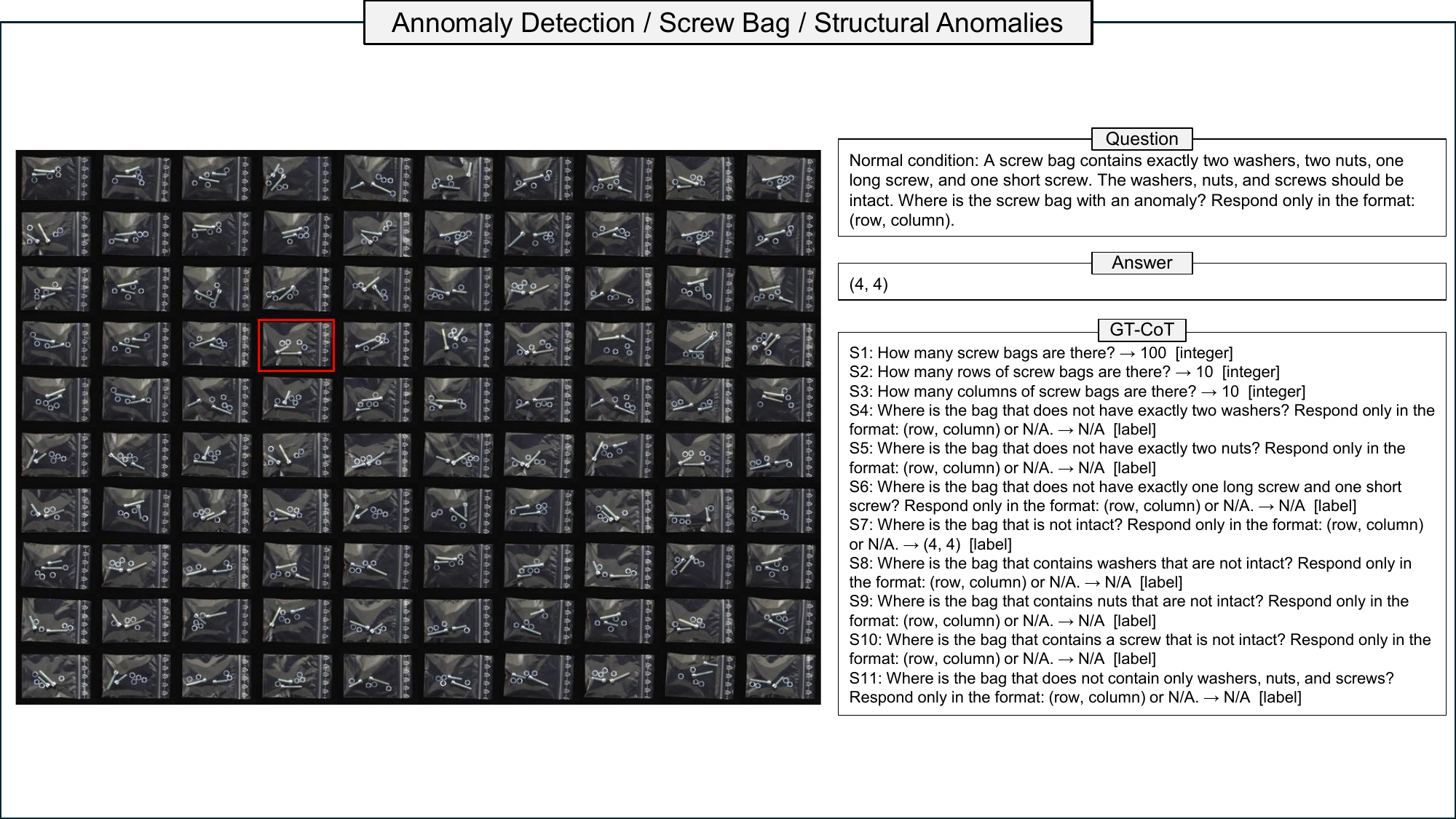}
\caption{Anomaly detection example: Screw Bag -- Structural Anomalies.}
\label{fig:ex_ad_screw_bag_structural}
\end{figure}

\vfill

\clearpage
\newpage

\vfill

\begin{figure}[!htb]
\centering
\includegraphics[width=\textwidth]{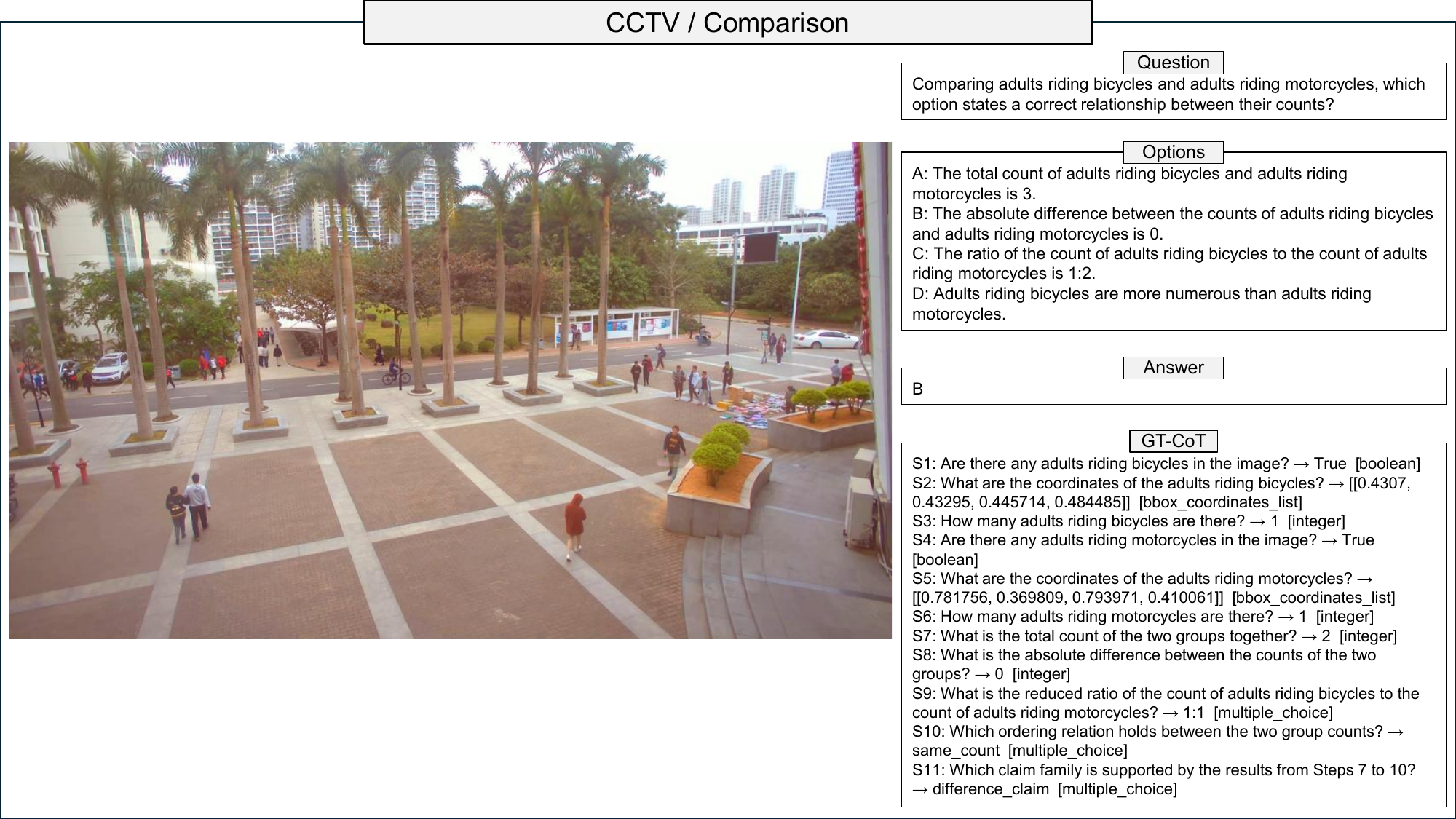}
\caption{CCTV example: Comparison.}
\label{fig:ex_cctv_comparison}
\end{figure}

\vfill

\begin{figure}[!htb]
\centering
\includegraphics[width=\textwidth]{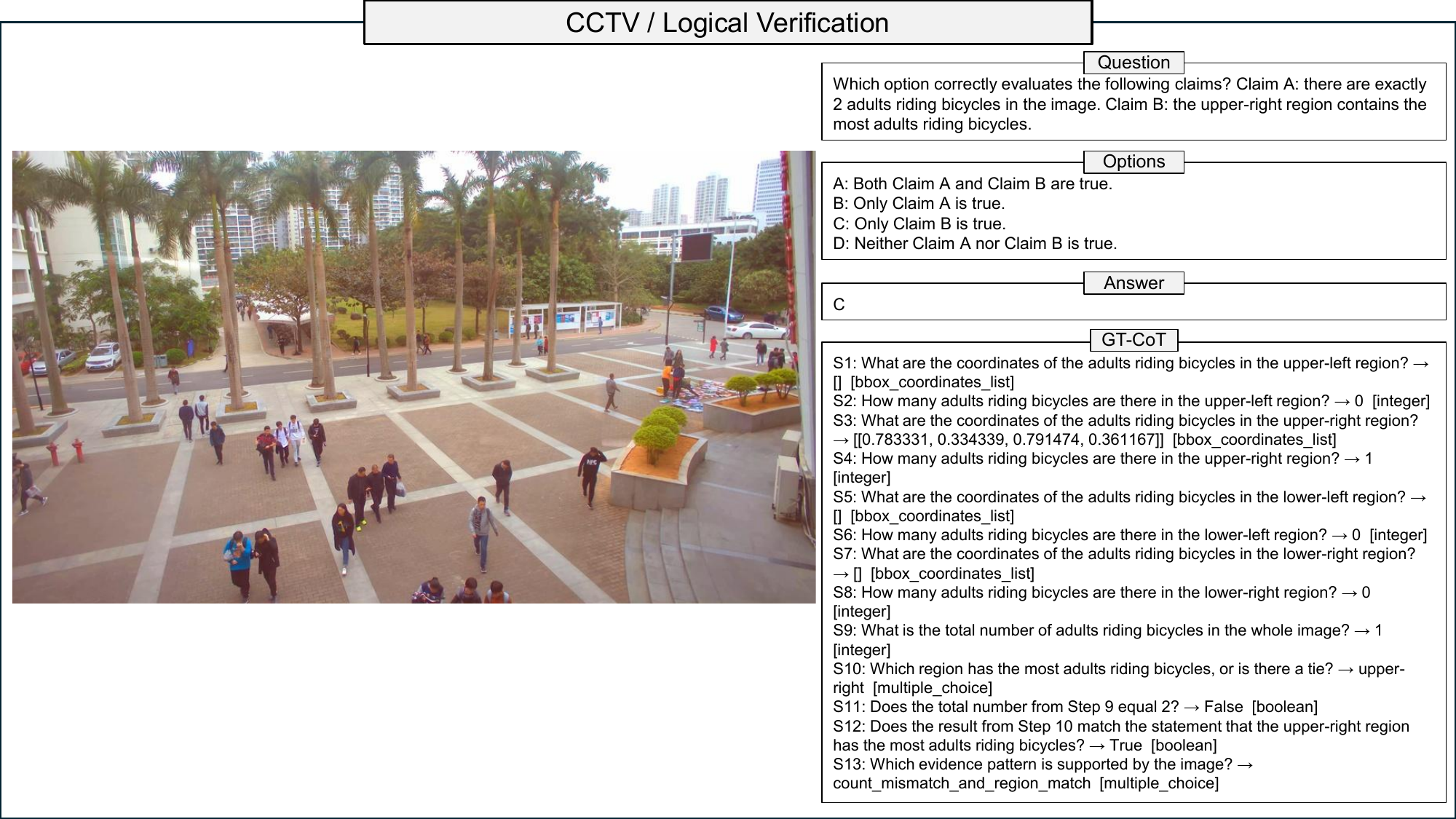}
\caption{CCTV example: Logical Verification.}
\label{fig:ex_cctv_logical_verification}
\end{figure}

\vfill

\clearpage
\newpage

\vfill

\begin{figure}[!htb]
\centering
\includegraphics[width=\textwidth]{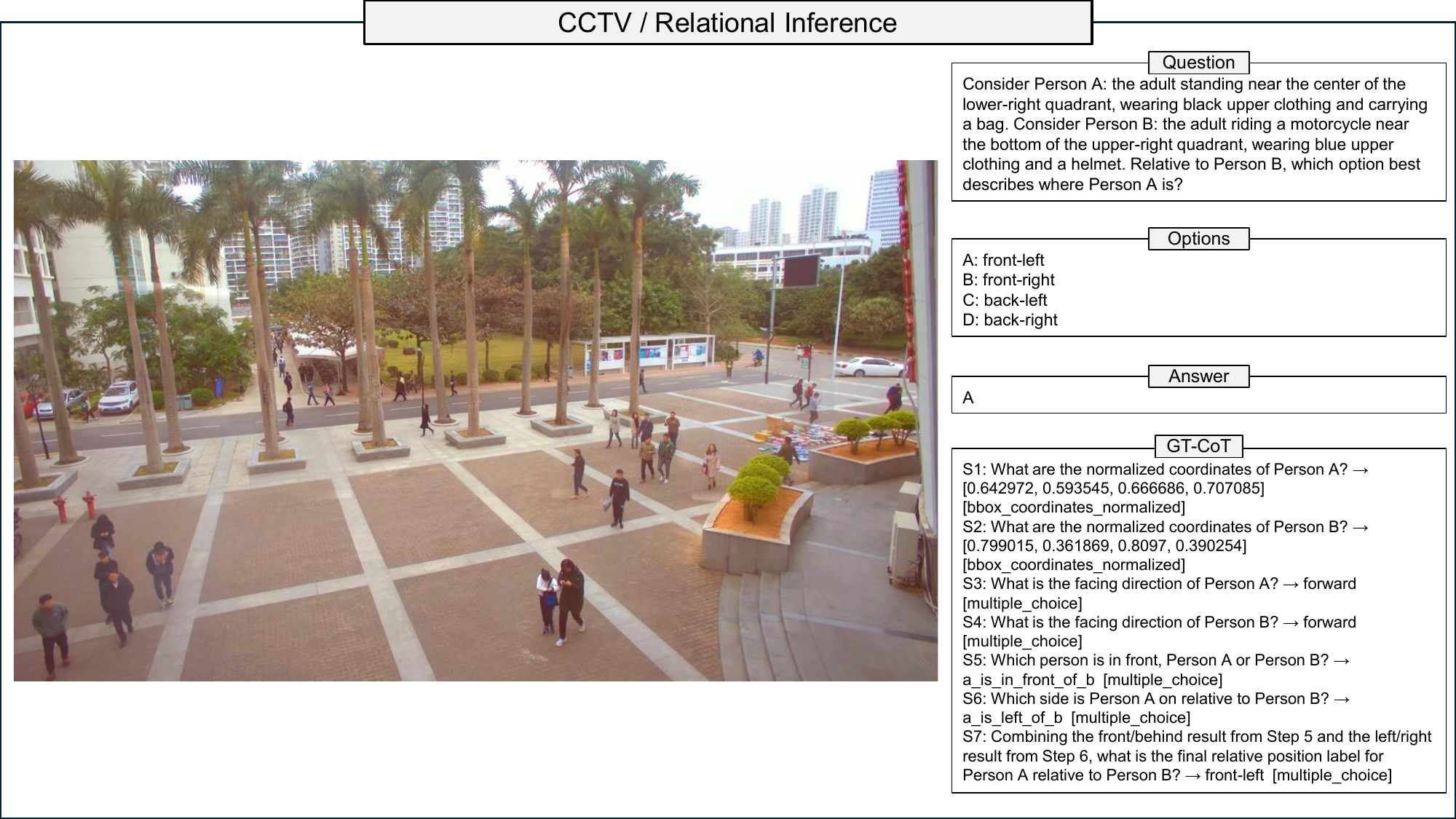}
\caption{CCTV example: Relational Inference.}
\label{fig:ex_cctv_relational_inference}
\end{figure}

\vfill

\begin{figure}[!htb]
\centering
\includegraphics[width=\textwidth]{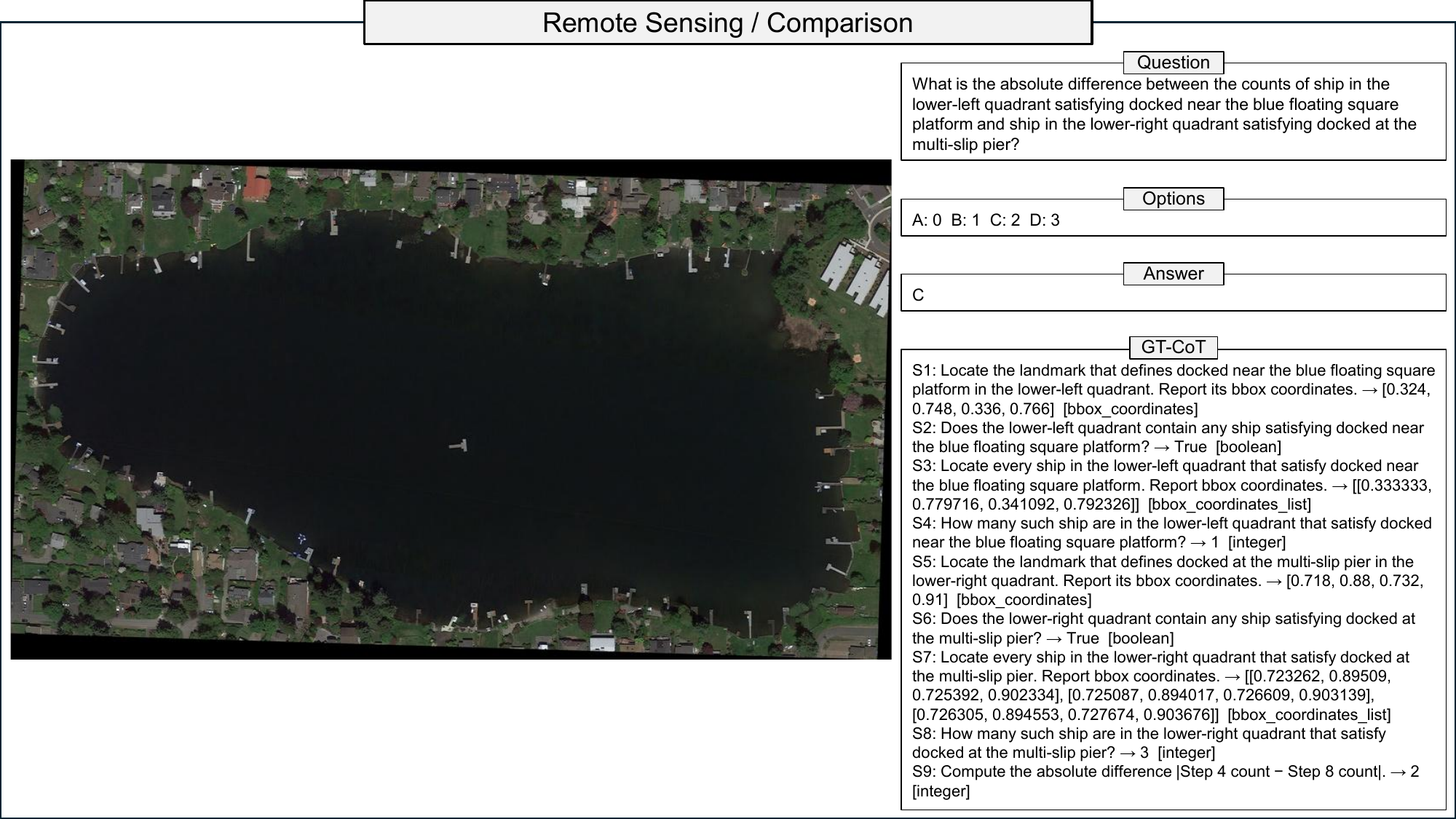}
\caption{Remote sensing example: Comparison.}
\label{fig:ex_rs_comparison}
\end{figure}

\vfill

\clearpage
\newpage

\vfill

\begin{figure}[!htb]
\centering
\includegraphics[width=\textwidth]{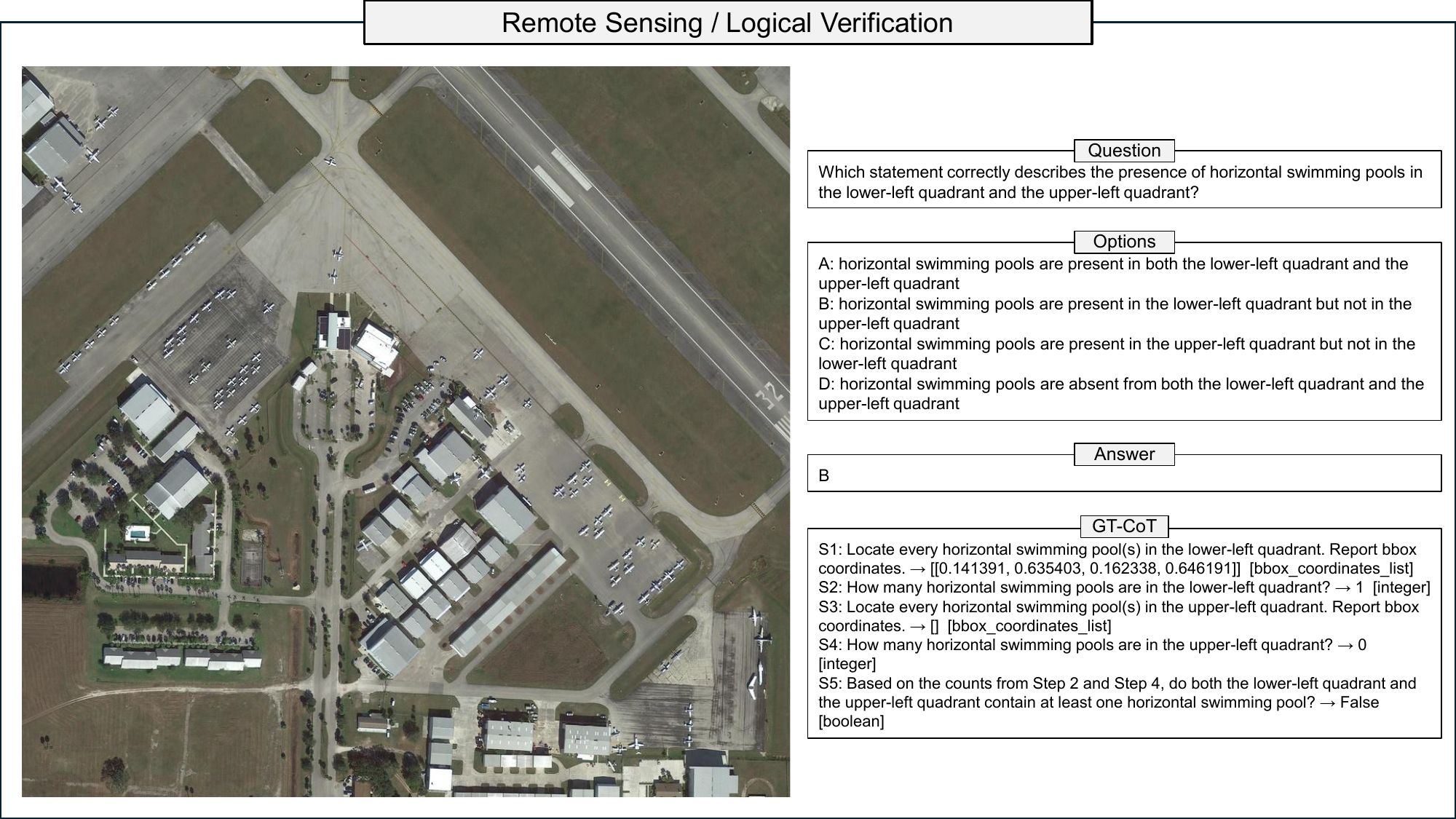}
\caption{Remote sensing example: Logical Verification.}
\label{fig:ex_rs_logical_verification}
\end{figure}

\vfill

\begin{figure}[!htb]
\centering
\includegraphics[width=\textwidth]{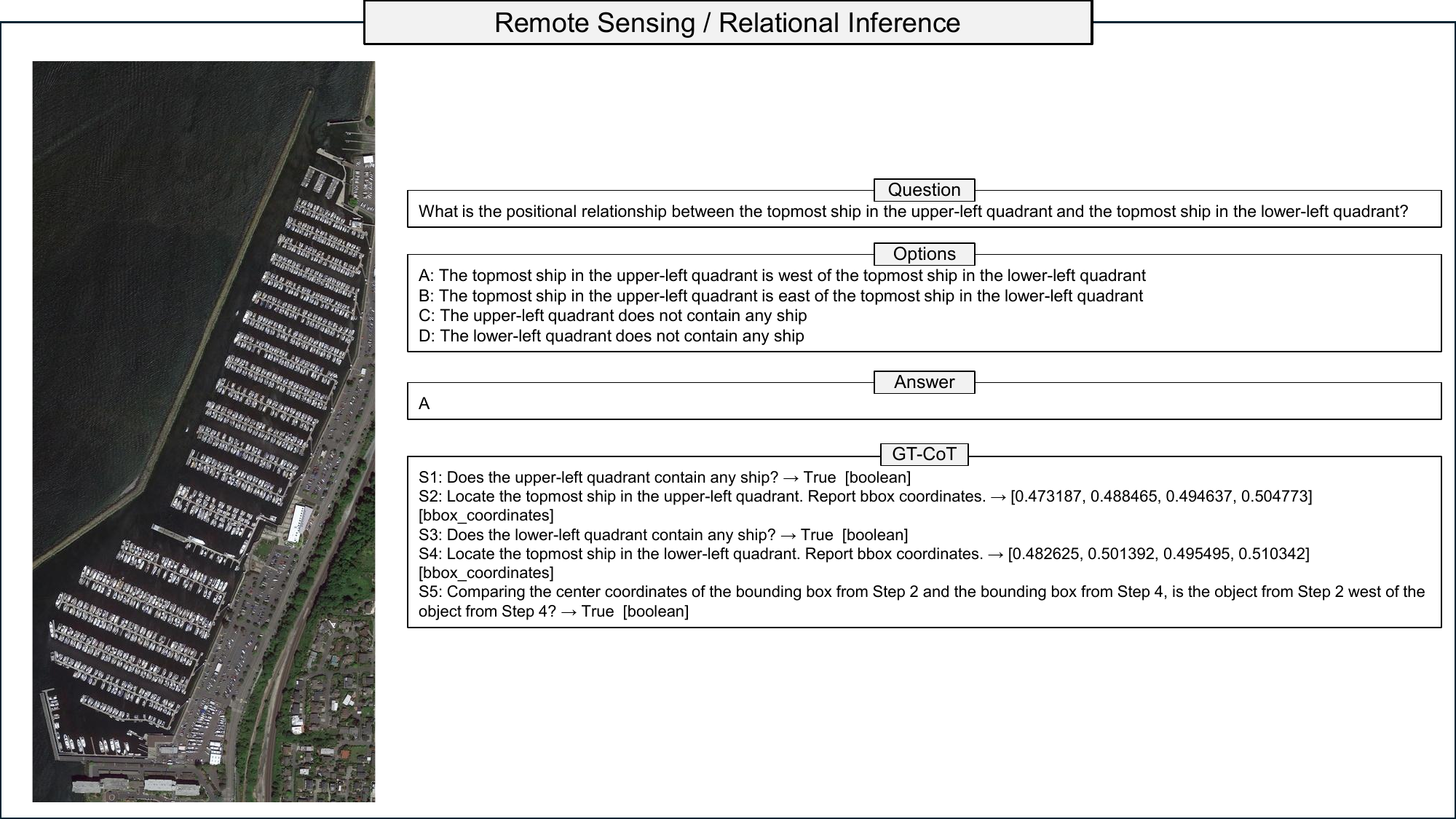}
\caption{Remote sensing example: Relational Inference.}
\label{fig:ex_rs_relational_inference}
\end{figure}

\vfill

\clearpage
\newpage

\vfill

\begin{figure}[!htb]
\centering
\includegraphics[width=\textwidth]{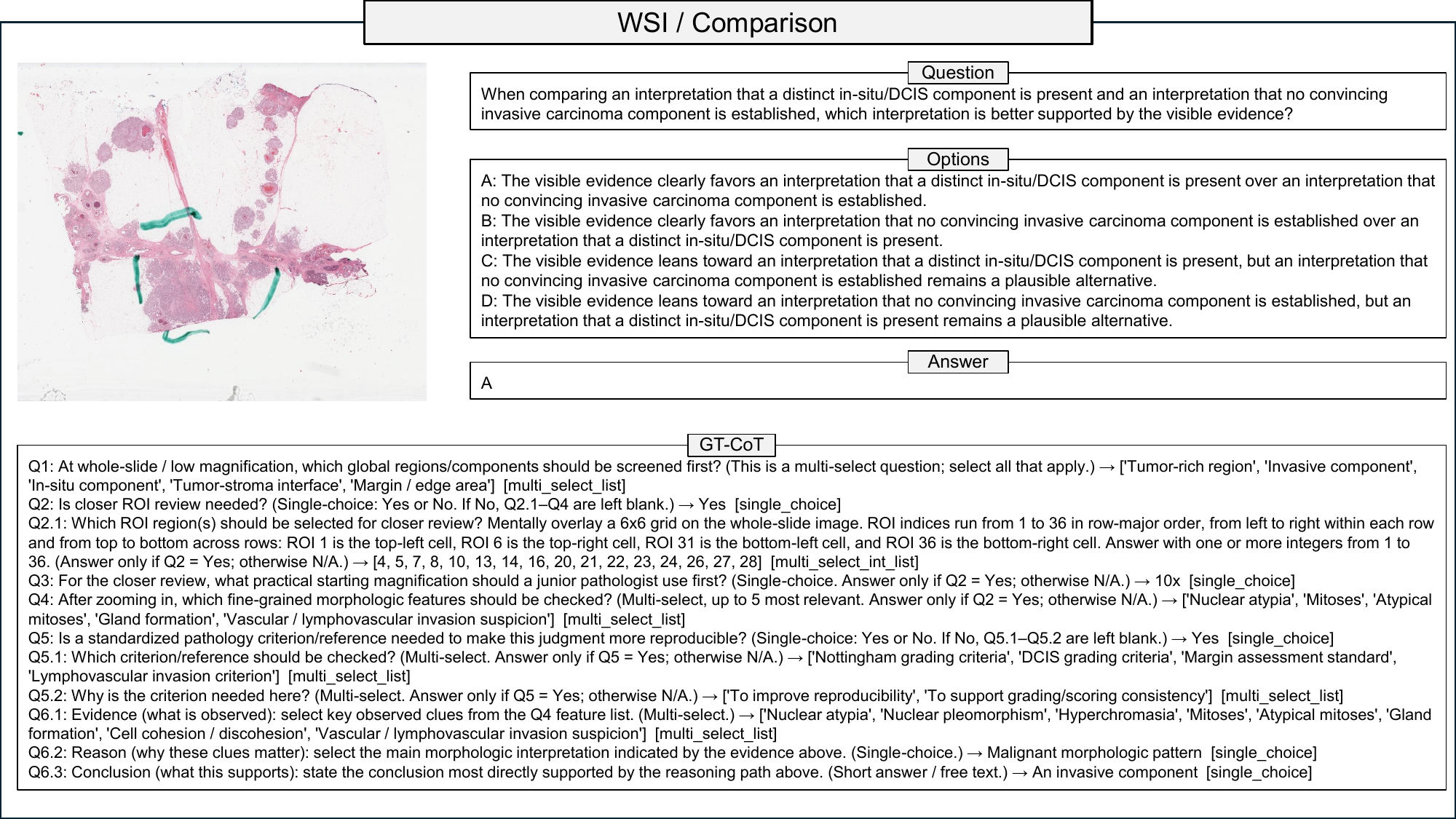}
\caption{Whole slide imaging example: Comparison.}
\label{fig:ex_wsi_comparison}
\end{figure}

\vfill

\begin{figure}[!htb]
\centering
\includegraphics[width=\textwidth]{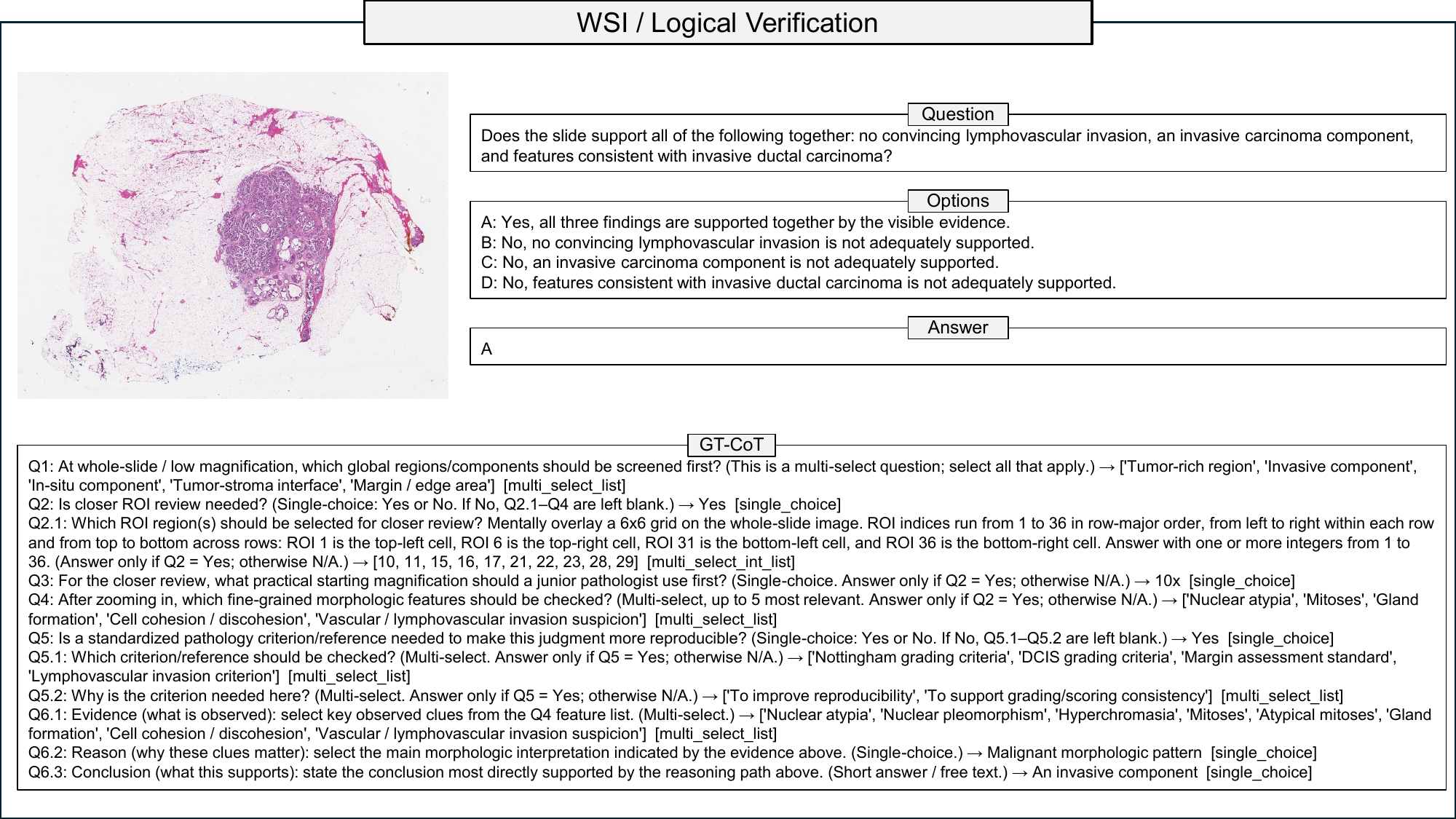}
\caption{Whole slide imaging example: Logical Verification.}
\label{fig:ex_wsi_logical_verification}
\end{figure}

\vfill

\clearpage
\newpage

\null
\vfill

\begin{figure}[!htb]
\centering
\includegraphics[width=\textwidth]{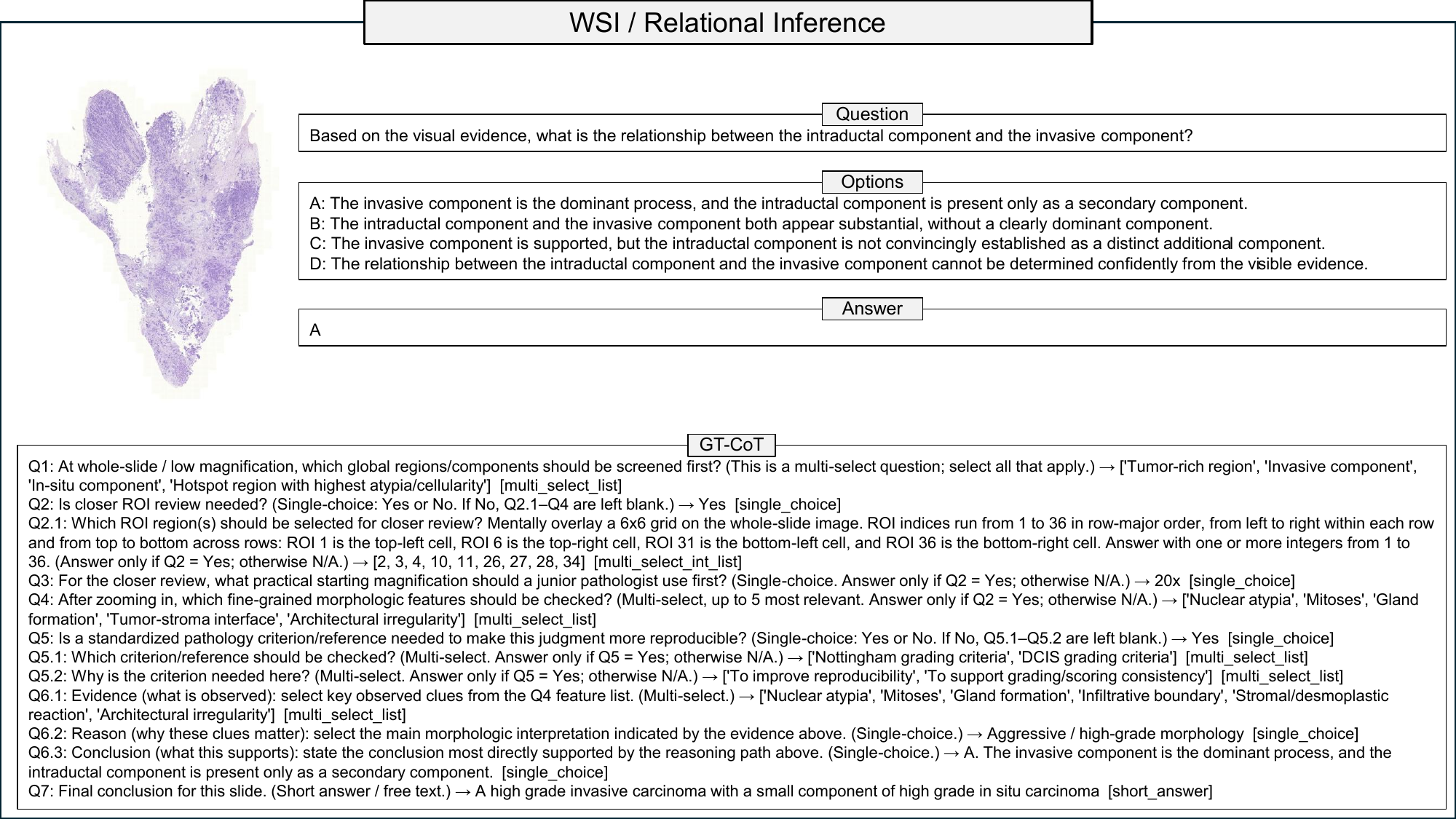}
\caption{Whole slide imaging example: Relational Inference.}
\label{fig:ex_wsi_relational_inference}
\end{figure}

\vfill

\clearpage
\newpage

\section{Prompt Templates}
\label{app:prompt_templates}

This section provides the full LLM prompt templates used for GT-CoT annotation construction in the CCTV, RS, and WSI domains.
Each template receives four runtime inputs: (1) \texttt{vqa\_meta} -- a JSON object describing the task schema and fixed subcategories; (2) \texttt{current\_reasoning\_category} -- one of \texttt{comparison}, \texttt{logical\_verification}, or \texttt{relational\_inference}; (3) \texttt{category\_meta} -- category-specific construction rules and examples; and (4) \texttt{img\_meta} -- structured annotation-derived metadata for the target image.

\subsection{CCTV Prompt Template}
\label{app:cctv_template}

Each CCTV question is built from the main prompt below plus one category-specific meta file injected as \texttt{category\_meta}.

\subsubsection*{Main Prompt}

\begin{lstlisting}[style=promptlst]
You are a CCTV VQA dataset constructor.

VQA type:
{vqa_meta}

Current reasoning category:
{current_reasoning_category}

Category-specific meta information:
{category_meta}

Structured annotation-derived metadata:
{img_meta}

Construction objective:
- Generate exactly one VQA question for the current reasoning category only.
- Use the structured annotation-derived metadata as the primary source of truth.
- Use the image only to confirm local visual details around the already selected target boxes or bundles.

How to use the structured annotation-derived metadata:
- The "current_category_candidate" section already contains the selected compact candidate for the current category.
- Do not search over the whole image for a different bundle, pair, or target unless the provided candidate is clearly unusable.
- For comparison, use the provided "group_a", "group_b", and their precomputed count-relation fields directly.
- For logical_verification, use the provided selected bundle, Claim A count, Claim B region/extremum, actual extremum evidence, evidence-pattern label, and its quadrant-wise boxes and counts directly.
- For relational_inference, use the provided selected targets, their finalized target descriptions, and their precomputed relation labels directly.
- Do not reconstruct large candidate tables from raw annotation in your answer.

Global rules:
- Keep the natural-language question text in English only.
- Never expose raw schema labels or internal field names in the final question text, options, or reasoning steps.
- Prefer compact natural noun phrases such as "standing adults" or "sitting children" over longer forms such as "adults who are standing" when both mean the same thing.
- The final question and the last reasoning step must not be semantically identical. The last step should summarize or map intermediate results to the final answer space.
- The top-level "category" field must exactly equal the current reasoning category.
- The top-level "subcategory" field must exactly match the fixed subcategory for the current reasoning category.
- The top-level "answer" field must be the correct option letter: "A", "B", "C", or "D".
- The last step inside "reasoning_supervision" must contain the semantic answer itself, while the top-level "answer" field must point to the matching option letter.
- Do not output free-form chain-of-thought.
- Every reasoning step must be closed-form, directly verifiable, and consistent with the provided metadata.
- Exactly one answer option must be correct.

How to use coordinates:
- All coordinates are normalized to the closed interval [0, 1].
- A single box must be written as [x1, y1, x2, y2].
- Multiple boxes must be written as a list of boxes.
- If a box list is required and there is no matching object, output [].
- If a count is required and there is no matching object, output 0.
- Use the bbox coordinates exactly as provided in the structured metadata.

Category-specific reminders:
- For comparison, provide one option each for difference, total, ratio, and ordering; the last reasoning step should be a claim-family label that maps to the single true relationship option.
- For logical_verification, use the provided quadrant-wise bbox lists and counts; the last reasoning step should be an evidence-pattern label that maps cleanly to the final two-claim option set.
- For relational_inference, introduce the provided target descriptions once when defining Person A and Person B in the final question, then use only Person A and Person B in the reasoning steps.

Output JSON only in this structure (no commentary, English only):
{
  "category": "{current_reasoning_category}",
  "subcategory": "...",
  "question": "...",
  "options": {
    "A": "...",
    "B": "...",
    "C": "...",
    "D": "..."
  },
  "answer": "A|B|C|D",
  "reasoning_supervision": {
    "step_type": "multi_step_qa",
    "steps": [
      {
        "step_id": "S1",
        "question": "...",
        "ground_truth": true,
        "answer_format": "boolean"
      }
    ]
  }
}

Hard validation checklist before you answer:
- The output contains exactly one question object, not a list.
- The "category" value exactly equals "{current_reasoning_category}".
- The "subcategory" value exactly matches the fixed subcategory for "{current_reasoning_category}".
- The output contains "category", "subcategory", "question", "options", "answer", and "reasoning_supervision".
- The reasoning supervision block uses "step_type": "multi_step_qa".
- No extra top-level keys appear.
- Every coordinate answer uses normalized coordinates.
- Every listed box is consistent with the current image's metadata.
- Every count is consistent with the corresponding bbox list.
- The final question is not a verbatim restatement of the last reasoning step.
- The top-level "answer" letter matches the correct option.
- No free-form CoT appears anywhere in the output.
\end{lstlisting}

\subsubsection*{Category Category: Comparison}

\begin{lstlisting}[style=promptlst]
# Comparison

Use exactly:
- category: `"comparison"`
- subcategory: `"attribute_bundle_count_difference"`

Fixed question template:
- `"Comparing [Group A] and [Group B], which option states a correct relationship between their counts?"`

Question example (format only, not a default content choice):
- `"Comparing adults riding bicycles and adults riding motorcycles, which option states a correct relationship between their counts?"`

## How to use the precomputed metadata

- Start from `"current_category_candidate"`.
- The bundle pair has already been selected in code.
- Use the provided `"group_a"` and `"group_b"` directly.
- Use the provided natural-language `"bundle_label"` values directly, or rewrite them into an equivalent but more compact natural form if needed.
- Use the provided `"count_difference"`, `"count_total"`, `"count_ratio_a_to_b"`, `"ordering_relation_label"`, and `"supported_claim_type"` directly.
- Use the provided false values directly when writing the three unsupported option families.
- Use the provided bbox lists directly. Do not search for a different bundle pair first.

## Current candidate schema

```json
{
  "group_a": {
    "bundle_label": "adults riding bicycles",
    "count": 2,
    "boxes": [
      [0.101, 0.202, 0.118, 0.220],
      [0.140, 0.230, 0.158, 0.260]
    ]
  },
  "group_b": {
    "bundle_label": "adults riding motorcycles",
    "count": 1,
    "boxes": [
      [0.601, 0.702, 0.618, 0.720]
    ]
  },
  "count_difference": 1,
  "count_total": 3,
  "count_ratio_a_to_b": "2:1",
  "ordering_relation_label": "group_a_more",
  "supported_claim_type": "difference_claim",
  "false_count_difference": 2,
  "false_count_total": 4,
  "false_count_ratio_a_to_b": "1:2",
  "false_ordering_relation_label": "group_b_more",
  "ratio_available_for_true_option": true
}
```

## Coordinate convention

- All bounding boxes use normalized coordinates in `[0, 1]`.
- The bbox format is `[x1, y1, x2, y2]`.

## Fixed reasoning supervision schema

- The steps array must contain exactly **11 steps**: `S1` to `S11`.
- `S1` answer_format must be `"boolean"`.
- `S2` answer_format must be `"bbox_coordinates_list"`.
- `S3` answer_format must be `"integer"`.
- `S4` answer_format must be `"boolean"`.
- `S5` answer_format must be `"bbox_coordinates_list"`.
- `S6` answer_format must be `"integer"`.
- `S7` answer_format must be `"integer"`.
- `S8` answer_format must be `"integer"`.
- `S9` answer_format must be `"multiple_choice"`.
- `S10` answer_format must be `"multiple_choice"`.
- `S11` answer_format must be `"multiple_choice"`.

## Schema example

```json
{
  "reasoning_supervision": {
    "step_type": "multi_step_qa",
    "steps": [
      {
        "step_id": "S1",
        "question": "Are there any adults riding bicycles in the image?",
        "ground_truth": true,
        "answer_format": "boolean"
      },
      {
        "step_id": "S2",
        "question": "What are the coordinates of the adults riding bicycles?",
        "ground_truth": [
          [0.101, 0.202, 0.118, 0.220],
          [0.140, 0.230, 0.158, 0.260]
        ],
        "answer_format": "bbox_coordinates_list"
      },
      {
        "step_id": "S3",
        "question": "How many adults riding bicycles are there?",
        "ground_truth": 2,
        "answer_format": "integer"
      },
      {
        "step_id": "S4",
        "question": "Are there any adults riding motorcycles in the image?",
        "ground_truth": true,
        "answer_format": "boolean"
      },
      {
        "step_id": "S5",
        "question": "What are the coordinates of the adults riding motorcycles?",
        "ground_truth": [
          [0.601, 0.702, 0.618, 0.720]
        ],
        "answer_format": "bbox_coordinates_list"
      },
      {
        "step_id": "S6",
        "question": "How many adults riding motorcycles are there?",
        "ground_truth": 1,
        "answer_format": "integer"
      },
      {
        "step_id": "S7",
        "question": "What is the total count of the two groups together?",
        "ground_truth": 3,
        "answer_format": "integer"
      },
      {
        "step_id": "S8",
        "question": "What is the absolute difference between the counts of the two groups?",
        "ground_truth": 1,
        "answer_format": "integer"
      },
      {
        "step_id": "S9",
        "question": "What is the reduced ratio of the count of adults riding bicycles to the count of adults riding motorcycles?",
        "ground_truth": "2:1",
        "answer_format": "multiple_choice"
      },
      {
        "step_id": "S10",
        "question": "Which ordering relation holds between the two group counts?",
        "ground_truth": "group_a_more",
        "answer_format": "multiple_choice"
      },
      {
        "step_id": "S11",
        "question": "Which claim family is supported by the results from Steps 7 to 10?",
        "ground_truth": "difference_claim",
        "answer_format": "multiple_choice"
      }
    ]
  }
}
```

## Exact step wording pattern

- `S1`: `"Are there any [Group A] in the image?"`
- `S2`: `"What are the coordinates of the [Group A]?"`
- `S3`: `"How many [Group A] are there?"`
- `S4`: `"Are there any [Group B] in the image?"`
- `S5`: `"What are the coordinates of the [Group B]?"`
- `S6`: `"How many [Group B] are there?"`
- `S7`: `"What is the total count of the two groups together?"`
- `S8`: `"What is the absolute difference between the counts of the two groups?"`
- `S9`: `"What is the reduced ratio of the count of [Group A] to the count of [Group B]?"`
- `S10`: `"Which ordering relation holds between the two group counts?"`
- `S11`: `"Which claim family is supported by the results from Steps 7 to 10?"`

## Option rules

- Provide exactly 4 options.
- The final question must ask the model to choose the correct relationship claim, not a single numeric answer.
- The 4 option families must be exactly:
  - one absolute-difference claim
  - one total-count claim
  - one ratio claim
  - one ordering claim
- The only true option must be the family specified by `supported_claim_type`.
- For the unsupported option families, use the provided false values directly:
  - `false_count_difference`
  - `false_count_total`
  - `false_count_ratio_a_to_b`
  - `false_ordering_relation_label`
- The ratio option must describe the ratio of the count of `group_a` to the count of `group_b` in reduced `A:B` form.
- The ordering option must express exactly one of these meanings:
  - `group_a_more`
  - `group_b_more`
  - `same_count`

## Recommended option wording patterns

- Difference claim:
  - `"The absolute difference between the counts of [Group A] and [Group B] is [value]."`
- Total claim:
  - `"The total count of [Group A] and [Group B] is [value]."`
- Ratio claim:
  - `"The ratio of the count of [Group A] to the count of [Group B] is [value]."`
- Ordering claim:
  - `"[Group A] are more numerous than [Group B]."`
  - `"[Group B] are more numerous than [Group A]."`
  - `"[Group A] and [Group B] have the same count."`

## Allowed intermediate answer space

- For `S9`, use the provided reduced ratio string exactly, for example `"2:1"` or `"1:1"`.
- For `S10`, use exactly one of:
  - `"group_a_more"`
  - `"group_b_more"`
  - `"same_count"`
- For `S11`, use exactly one of:
  - `"difference_claim"`
  - `"total_claim"`
  - `"ratio_claim"`
  - `"ordering_claim"`

## Category-specific validation checklist

- `[Group A]` and `[Group B]` must exactly match the provided candidate metadata.
- `S2` must contain all and only the boxes from `group_a.boxes`.
- `S5` must contain all and only the boxes from `group_b.boxes`.
- `S1` must equal whether `S2` is non-empty.
- `S4` must equal whether `S5` is non-empty.
- `len(S2)` must equal `S3`.
- `len(S5)` must equal `S6`.
- `S7` must equal `S3 + S6`.
- `S8` must equal `abs(S3 - S6)`.
- `S9` must equal `count_ratio_a_to_b`.
- `S10` must equal `ordering_relation_label`.
- `S11` must equal `supported_claim_type`.
- Exactly one option may be true.
- The true option family must match `S11`.
- If `S11 = "difference_claim"`, the difference option must use `count_difference`, and the total, ratio, and ordering options must use the provided false values.
- If `S11 = "total_claim"`, the total option must use `count_total`, and the difference, ratio, and ordering options must use the provided false values.
- If `S11 = "ratio_claim"`, the ratio option must use `count_ratio_a_to_b`, and the difference, total, and ordering options must use the provided false values.
- If `S11 = "ordering_claim"`, the ordering option must use `ordering_relation_label`, and the difference, total, and ratio options must use the provided false values.
- If `ratio_available_for_true_option` is false, the ratio option must be false.
- The final question must use natural English only.

## Invalid behavior to avoid

- Re-selecting a different bundle pair from the image.
- Rewriting the bundle labels into a different semantic meaning.
- Producing counts that do not match the listed coordinates.
- Reusing the same bbox in both groups.
- Making multiple option families true at the same time.
- Making the last reasoning step semantically identical to the final question.
- Using raw schema labels or internal field names in the question text.
\end{lstlisting}

\subsubsection*{Category Category: Logical Verification}

\begin{lstlisting}[style=promptlst]
# Logical Verification

Use exactly:
- category: `"logical_verification"`
- subcategory: `"joint_count_and_region_extremum_verification"`

Fixed question template:
- `"Which option correctly evaluates the following claims? Claim A: there are exactly [N] [question_bundle_label] in the image. Claim B: the [region] region contains the [most/fewest] [bundle_label]."`

Question example (format only, not a default content choice):
- `"Which option correctly evaluates the following claims? Claim A: there are exactly 3 standing adults in the image. Claim B: the upper-right region contains the most standing adults."`

## How to use the precomputed metadata

- Start from `"current_category_candidate"`.
- The bundle has already been selected in code.
- Use the provided `"bundle_label"` for region-wise counting and all region-extremum reasoning.
- Use `"claim_a_count"` and `"question_bundle_label"` for Claim A.
- Use `"claim_b_region"` and `"claim_b_extremum"` together with `"bundle_label"` for Claim B.
- Use the provided `"actual_extremum_regions"`, `"actual_extremum_resolution"`, `"count_match"`, `"region_statement_match"`, and `"evidence_pattern_label"` directly.
- Use the provided quadrant-wise bbox lists and counts directly. Do not search for a different bundle first.

## Current candidate schema

```json
{
  "bundle_label": "standing adults",
  "question_bundle_label": "standing adults",
  "actual_count": 3,
  "claim_a_count": 3,
  "claim_a_is_true": true,
  "claim_b_region": "upper-right",
  "claim_b_extremum": "most",
  "claim_b_is_true": false,
  "actual_extremum_regions": ["upper-left", "upper-right", "lower-left"],
  "actual_extremum_resolution": "tie",
  "count_match": true,
  "region_statement_match": false,
  "evidence_pattern_label": "count_match_and_region_mismatch",
  "combined_truth_label": "only_claim_a_true",
  "quadrant_to_boxes": {
    "upper-left": [[0.112, 0.284, 0.138, 0.402]],
    "upper-right": [[0.614, 0.216, 0.640, 0.336]],
    "lower-left": [[0.228, 0.704, 0.252, 0.836]],
    "lower-right": []
  },
  "quadrant_to_count": {
    "upper-left": 1,
    "upper-right": 1,
    "lower-left": 1,
    "lower-right": 0
  }
}
```

## Coordinate convention

- All bounding boxes use normalized coordinates in `[0, 1]`.
- The bbox format is `[x1, y1, x2, y2]`.

## Fixed reasoning supervision schema

- The steps array must contain exactly **13 steps**: `S1` to `S13`.
- `S1` answer_format must be `"bbox_coordinates_list"`.
- `S2` answer_format must be `"integer"`.
- `S3` answer_format must be `"bbox_coordinates_list"`.
- `S4` answer_format must be `"integer"`.
- `S5` answer_format must be `"bbox_coordinates_list"`.
- `S6` answer_format must be `"integer"`.
- `S7` answer_format must be `"bbox_coordinates_list"`.
- `S8` answer_format must be `"integer"`.
- `S9` answer_format must be `"integer"`.
- `S10` answer_format must be `"multiple_choice"`.
- `S11` answer_format must be `"boolean"`.
- `S12` answer_format must be `"boolean"`.
- `S13` answer_format must be `"multiple_choice"`.

## Schema example

```json
{
  "reasoning_supervision": {
    "step_type": "multi_step_qa",
    "steps": [
      {
        "step_id": "S1",
        "question": "What are the coordinates of the standing adults in the upper-left region?",
        "ground_truth": [
          [0.112, 0.284, 0.138, 0.402]
        ],
        "answer_format": "bbox_coordinates_list"
      },
      {
        "step_id": "S2",
        "question": "How many standing adults are there in the upper-left region?",
        "ground_truth": 1,
        "answer_format": "integer"
      },
      {
        "step_id": "S3",
        "question": "What are the coordinates of the standing adults in the upper-right region?",
        "ground_truth": [
          [0.614, 0.216, 0.640, 0.336]
        ],
        "answer_format": "bbox_coordinates_list"
      },
      {
        "step_id": "S4",
        "question": "How many standing adults are there in the upper-right region?",
        "ground_truth": 1,
        "answer_format": "integer"
      },
      {
        "step_id": "S5",
        "question": "What are the coordinates of the standing adults in the lower-left region?",
        "ground_truth": [
          [0.228, 0.704, 0.252, 0.836]
        ],
        "answer_format": "bbox_coordinates_list"
      },
      {
        "step_id": "S6",
        "question": "How many standing adults are there in the lower-left region?",
        "ground_truth": 1,
        "answer_format": "integer"
      },
      {
        "step_id": "S7",
        "question": "What are the coordinates of the standing adults in the lower-right region?",
        "ground_truth": [],
        "answer_format": "bbox_coordinates_list"
      },
      {
        "step_id": "S8",
        "question": "How many standing adults are there in the lower-right region?",
        "ground_truth": 0,
        "answer_format": "integer"
      },
      {
        "step_id": "S9",
        "question": "What is the total number of standing adults in the whole image?",
        "ground_truth": 3,
        "answer_format": "integer"
      },
      {
        "step_id": "S10",
        "question": "Which region has the most standing adults, or is there a tie?",
        "ground_truth": "tie",
        "answer_format": "multiple_choice"
      },
      {
        "step_id": "S11",
        "question": "Does the total number from Step 9 equal 3?",
        "ground_truth": true,
        "answer_format": "boolean"
      },
      {
        "step_id": "S12",
        "question": "Does the result from Step 10 match the statement that the upper-right region has the most standing adults?",
        "ground_truth": false,
        "answer_format": "boolean"
      },
      {
        "step_id": "S13",
        "question": "Which evidence pattern is supported by the image?",
        "ground_truth": "count_match_and_region_mismatch",
        "answer_format": "multiple_choice"
      }
    ]
  }
}
```

## Exact step wording pattern

- `S1`: `"What are the coordinates of the [bundle_label] in the upper-left region?"`
- `S2`: `"How many [bundle_label] are there in the upper-left region?"`
- `S3`: `"What are the coordinates of the [bundle_label] in the upper-right region?"`
- `S4`: `"How many [bundle_label] are there in the upper-right region?"`
- `S5`: `"What are the coordinates of the [bundle_label] in the lower-left region?"`
- `S6`: `"How many [bundle_label] are there in the lower-left region?"`
- `S7`: `"What are the coordinates of the [bundle_label] in the lower-right region?"`
- `S8`: `"How many [bundle_label] are there in the lower-right region?"`
- `S9`: `"What is the total number of [bundle_label] in the whole image?"`
- `S10`: `"Which region has the [claim_b_extremum] [bundle_label], or is there a tie?"`
- `S11`: `"Does the total number from Step 9 equal [claim_a_count]?"`
- `S12`: `"Does the result from Step 10 match the statement that the [claim_b_region] region has the [claim_b_extremum] [bundle_label]?"`
- `S13`: `"Which evidence pattern is supported by the image?"`

## Option rules

- Provide exactly 4 options.
- The option texts must be exactly:
  - `"Both Claim A and Claim B are true."`
  - `"Only Claim A is true."`
  - `"Only Claim B is true."`
  - `"Neither Claim A nor Claim B is true."`
- Exactly one option must match the semantic answer implied by `S13`.

## Allowed intermediate answer space

- For `S10`, use exactly one of:
  - `"upper-left"`
  - `"upper-right"`
  - `"lower-left"`
  - `"lower-right"`
  - `"tie"`
- For `S13`, use exactly one of:
  - `"count_match_and_region_match"`
  - `"count_match_and_region_mismatch"`
  - `"count_mismatch_and_region_match"`
  - `"count_mismatch_and_region_mismatch"`

## Category-specific validation checklist

- `S1`, `S3`, `S5`, and `S7` must exactly match the provided quadrant bbox lists.
- `S2`, `S4`, `S6`, and `S8` must exactly match the provided quadrant counts.
- `S9` must equal `S2 + S4 + S6 + S8`.
- `S10` must equal `actual_extremum_resolution`.
- `S11` must equal `count_match`.
- `S12` must equal `region_statement_match`.
- `S13` must equal:
  - `"count_match_and_region_match"` if `S11` is true and `S12` is true
  - `"count_match_and_region_mismatch"` if `S11` is true and `S12` is false
  - `"count_mismatch_and_region_match"` if `S11` is false and `S12` is true
  - `"count_mismatch_and_region_mismatch"` if `S11` is false and `S12` is false
- The final question must quote both Claim A and Claim B.
- Claim A must use `claim_a_count` and `question_bundle_label`.
- Claim B must use `claim_b_region`, `claim_b_extremum`, and `bundle_label`.
- The final option mapping must be:
  - `S13 = "count_match_and_region_match"` -> `"Both Claim A and Claim B are true."`
  - `S13 = "count_match_and_region_mismatch"` -> `"Only Claim A is true."`
  - `S13 = "count_mismatch_and_region_match"` -> `"Only Claim B is true."`
  - `S13 = "count_mismatch_and_region_mismatch"` -> `"Neither Claim A nor Claim B is true."`
- The final question must use natural English only.

## Invalid behavior to avoid

- Re-selecting a different bundle from the image.
- Changing the bundle meaning while rewriting the question.
- Producing bbox lists and counts that disagree with each other.
- Ignoring the fixed four-quadrant partition.
- Asking `Is Claim A true ...` or `Is Claim B true ...` directly in the reasoning steps.
- Making the last reasoning step a restatement of the final question.
- Using raw schema labels or internal field names in the question text.
\end{lstlisting}

\subsubsection*{Category Category: Relational Inference}

\begin{lstlisting}[style=promptlst]
# Relational Inference

Use exactly:
- category: `"relational_inference"`
- subcategory: `"relative_position_between_red_green_targets"`

Fixed question template pool:

- `"Person A is [Target A description], and Person B is [Target B description]. Which option best describes the position of Person A relative to Person B?"`
- `"Consider Person A: [Target A description]. Consider Person B: [Target B description]. Relative to Person B, which option best describes where Person A is?"`

## How to use the precomputed metadata

- Start from `"current_category_candidate"`.
- The two targets have already been selected in code.
- Use the provided bbox coordinates, facing directions, finalized target descriptions, and precomputed relation labels directly.
- Use the provided `"target_description"` fields directly when introducing Person A and Person B in the final question.
- After the final question introduces Person A and Person B, use only `Person A` and `Person B` in reasoning steps `S1` to `S7`.
- Do not reconstruct a new description from raw visual attributes or any other internal fields.
- Do not rewrite the target description unless very minor smoothing is absolutely necessary.
- Do not mention `red`, `green`, `red bounding box`, or `green bounding box` in the final question or reasoning steps.
- Do not search for a different pair first.

## Current candidate schema

```json
{
  "front_back_relation": "a_is_in_front_of_b",
  "left_right_relation": "a_is_left_of_b",
  "final_relation": "front-left",
  "target_a": {
    "label": "Person A",
    "bbox_normalized": [0.120, 0.340, 0.168, 0.438],
    "direction": "left",
    "target_description": "the adult standing near the center of the upper-left quadrant, wearing pink upper clothing"
  },
  "target_b": {
    "label": "Person B",
    "bbox_normalized": [0.802, 0.118, 0.836, 0.186],
    "direction": "right",
    "target_description": "the adult standing near the center of the upper-right quadrant, wearing light upper clothing"
  }
}
```

## Coordinate convention

- All bounding box coordinates must be normalized to `[0, 1]`.
- The bbox format is `[x1, y1, x2, y2]`.

## Fixed answer space

The final answer must be exactly one of:
- `"front-left"`
- `"front-right"`
- `"back-left"`
- `"back-right"`

## Fixed reasoning supervision schema

- The steps array must contain exactly **7 steps**: `S1` to `S7`.
- `S1` answer_format must be `"bbox_coordinates_normalized"`.
- `S2` answer_format must be `"bbox_coordinates_normalized"`.
- `S3` answer_format must be `"multiple_choice"`.
- `S4` answer_format must be `"multiple_choice"`.
- `S5` answer_format must be `"multiple_choice"`.
- `S6` answer_format must be `"multiple_choice"`.
- `S7` answer_format must be `"multiple_choice"`.

## Schema example

```json
{
  "reasoning_supervision": {
    "step_type": "multi_step_qa",
    "steps": [
      {
        "step_id": "S1",
        "question": "What are the normalized coordinates of Person A?",
        "ground_truth": [0.120, 0.340, 0.168, 0.438],
        "answer_format": "bbox_coordinates_normalized"
      },
      {
        "step_id": "S2",
        "question": "What are the normalized coordinates of Person B?",
        "ground_truth": [0.802, 0.118, 0.836, 0.186],
        "answer_format": "bbox_coordinates_normalized"
      },
      {
        "step_id": "S3",
        "question": "What is the facing direction of Person A?",
        "ground_truth": "left",
        "answer_format": "multiple_choice"
      },
      {
        "step_id": "S4",
        "question": "What is the facing direction of Person B?",
        "ground_truth": "right",
        "answer_format": "multiple_choice"
      },
      {
        "step_id": "S5",
        "question": "Which person is in front, Person A or Person B?",
        "ground_truth": "a_is_in_front_of_b",
        "answer_format": "multiple_choice"
      },
      {
        "step_id": "S6",
        "question": "Which side is Person A on relative to Person B?",
        "ground_truth": "a_is_left_of_b",
        "answer_format": "multiple_choice"
      },
      {
        "step_id": "S7",
        "question": "Combining the front/behind result from Step 5 and the left/right result from Step 6, what is the final relative position label for Person A relative to Person B?",
        "ground_truth": "front-left",
        "answer_format": "multiple_choice"
      }
    ]
  }
}
```

## Exact step wording pattern

- `S1`: `"What are the normalized coordinates of Person A?"`
- `S2`: `"What are the normalized coordinates of Person B?"`
- `S3`: `"What is the facing direction of Person A?"`
- `S4`: `"What is the facing direction of Person B?"`
- `S5`: `"Which person is in front, Person A or Person B?"`
- `S6`: `"Which side is Person A on relative to Person B?"`
- `S7`: `"Combining the front/behind result from Step 5 and the left/right result from Step 6, what is the final relative position label for Person A relative to Person B?"`

## Option rules

- Provide exactly 4 options.
- The 4 option texts must be exactly:
  - `"front-left"`
  - `"front-right"`
  - `"back-left"`
  - `"back-right"`
- Exactly one option must match `S7`.

## Allowed intermediate answer space

- For `S3` and `S4`, use exactly one of:
  - `"forward"`
  - `"backward"`
  - `"left"`
  - `"right"`
- For `S5`, use exactly one of:
  - `"a_is_in_front_of_b"`
  - `"a_is_behind_b"`
- For `S6`, use exactly one of:
  - `"a_is_left_of_b"`
  - `"a_is_right_of_b"`

## Category-specific validation checklist

- The two target descriptions must refer to the provided selected pair only.
- The final question must define Person A and Person B using the provided target descriptions.
- The reasoning steps must use only Person A and Person B, not the full target descriptions and not any red/green wording.
- The final target descriptions must exactly use the provided `"target_description"` values, or differ only by very minor smoothing.
- The final target descriptions must not include facing direction.
- `S1` and `S2` must exactly match the provided bbox coordinates.
- `S3` and `S4` must exactly match the provided facing directions.
- `S5` must exactly match `"front_back_relation"`.
- `S6` must exactly match `"left_right_relation"`.
- `S7` must exactly match `"final_relation"`.
- The final question must ask for option selection, while `S7` should perform the summary mapping from intermediate relations to the final label.

## Invalid behavior to avoid

- Re-selecting a different target pair from the image.
- Reconstructing a new target description from hidden cues or raw visual details.
- Re-running direction estimation inside this prompt.
- Mentioning `red` or `green` bounding boxes in the question or reasoning steps.
- Using raw pixel coordinates instead of normalized coordinates.
- Producing a final relation inconsistent with the provided relation labels.
- Making the final question text semantically identical to `S7`.
\end{lstlisting}

\subsection{RS Prompt Template}
\label{app:rs_template}

Each RS question is built from the main prompt below plus one category-specific meta file injected as \texttt{category\_meta}.

\subsubsection*{Main Prompt}

\begin{lstlisting}[style=promptlst]
You are a remote sensing VQA dataset constructor.

VQA type:
{vqa_meta}

Current reasoning category:
{current_reasoning_category}

Category-specific meta information:
{category_meta}

Structured annotation-derived metadata:
{img_meta}

Construction objective:
- Generate exactly one VQA question for the current reasoning category only.
- Use the structured annotation-derived metadata as the primary source of truth.
- Use the image only to inspect local visual structures around shortlisted annotation boxes, validate semantic conditions, and reject ambiguous choices.

How to use the structured annotation-derived metadata:
- The "regions" section already groups annotation boxes by quadrant using bbox-center assignment.
- The "regions" section already provides per-category object counts, raw boxes, orientation summaries, and unique-extreme summaries inside each quadrant.
- The "current_category_candidates" section already provides the code-precomputed candidate list for the current reasoning category.
- Do not rebuild this candidate table from scratch from the raw boxes unless the current category candidate list is empty.
- For comparison, choose the region pair and object type from "current_category_candidates", then use the image to decide an image-specific semantic condition and filter the raw candidate boxes from the corresponding region/category entries.
- For logical_verification, choose one candidate from "current_category_candidates" and use it directly. Do not invent a different region pair or orientation pattern unless the candidate list is empty.
- For relational_inference, choose one candidate from "current_category_candidates" and use it directly. Do not invent a different selector pair or relation pattern unless the candidate list is empty.

Global rules:
- Treat the structured annotation-derived metadata as the source of truth for object existence, region membership, counting, orientation, extreme-object selection, and spatial relations.
- Use only evidence visible in the image and the provided structured metadata.
- Use the image to validate local structures and semantic conditions, not to invent extra objects or hidden counts.
- Keep the natural-language question text in English only.
- Never expose raw metadata category ids with underscores in the final question text, options, or reasoning steps.
- When the structured metadata provides "object_type_label", use it in natural-language text. Otherwise replace "_" with spaces before writing the object name.
- Use quadrant names exactly from this set: "upper-left", "upper-right", "lower-left", "lower-right".
- The example and schema example in the category-specific meta information are format references only. Copying their concrete region pair, object type, condition, selector, or relation by default is an error.
- If a candidate question is ambiguous, trivial, unsupported, or too similar to the example without image-specific justification, discard it and choose another valid candidate within the same reasoning category.
- Prefer candidates with manageable support size and clean boundaries. Avoid very dense object sets or messy clusters when a clearer candidate exists.
- Do not output free-form chain-of-thought.
- Every reasoning step must be a clearly defined, closed-form, verifiable question.
- Every final answer must be derivable from the intermediate answers.
- Exactly one answer option must be correct.
- For comparison, the four option texts must be distinct non-negative integer values, and the semantic answer must be the absolute difference between the two regional counts.
- The top-level "category" field must exactly equal the current reasoning category.
- The top-level "subcategory" field must exactly match the fixed subcategory for the current reasoning category.
- The top-level "answer" field must be the correct option letter: "A", "B", "C", or "D".
- The last step inside "reasoning_supervision" must contain the semantic answer itself, while the top-level "answer" field must point to the matching option letter.

How to use coordinates and annotation rules:
- Coordinates are normalized to 0-1.
- Coordinate values should be decimal numbers in the closed interval [0, 1].
- When a reasoning step requires object localization, use the bbox coordinates exactly as provided in the structured metadata.
- A single box must be written as [x1, y1, x2, y2].
- Multiple boxes must be written as a list of boxes:
  [
    [x1, y1, x2, y2],
    [x1, y1, x2, y2]
  ]
- If a box list is required and there is no matching object, output [].
- If a count is required and there is no matching object, output 0.
- Every listed box must come from the current image's structured metadata.
- No box may appear in two different regions within the same question.
- For every count step, the integer ground truth must exactly equal the number of boxes listed in the corresponding coordinate step.
- For comparison, every filtered box must be a subset of the chosen region/category box list in the "regions" section.
- For logical_verification, every listed box must match the chosen candidate entry's object type, region, and orientation.
- For relational_inference, every selected box, selector, and relation must match the chosen candidate entry.

Output JSON only in this structure (no commentary, English only):
{
  "category": "{current_reasoning_category}",
  "subcategory": "...",
  "question": "...",
  "options": {
    "A": "...",
    "B": "...",
    "C": "...",
    "D": "..."
  },
  "answer": "A|B|C|D",
  "reasoning_supervision": {
    "step_type": "multi_step_qa",
    "steps": [
      {
        "step_id": "S1",
        "question": "...",
        "ground_truth": true,
        "answer_format": "boolean"
      }
    ]
  }
}

Hard validation checklist before you answer:
- The output contains exactly one question object, not a list of questions.
- The "category" value exactly equals "{current_reasoning_category}".
- The "subcategory" value exactly matches the fixed subcategory for "{current_reasoning_category}".
- The question follows the current category's fixed template and style, but its slots are chosen from the current image's structured metadata rather than copied from the example.
- The chosen region pair and object type come from the current category candidate list unless that list is empty.
- The output contains "category", "subcategory", "question", "options", "answer", and "reasoning_supervision".
- The reasoning supervision block uses "step_type": "multi_step_qa".
- Every step is closed-form and verifiable.
- Every coordinate ground truth uses bbox coordinates only.
- Every coordinate list is region-consistent under the structured metadata.
- No duplicate box is assigned to two different regions.
- Every count is consistent with the corresponding coordinate list.
- For comparison, the last reasoning step must be an integer equal to the absolute difference between the two count steps, and exactly one option value must match it.
- Every orientation, selector, and spatial relation is consistent with the structured metadata rules.
- The top-level "answer" letter matches the correct option.
- No free-form CoT appears anywhere in the output.
- No extra top-level keys appear.
\end{lstlisting}

\subsubsection*{Category Category: Comparison}

\begin{lstlisting}[style=promptlst]
- category: "comparison"
- subcategory: "conditioned_count_comparison"

Fixed question template:
- "What is the absolute difference between the counts of [object type] in [region A] satisfying [condition A] and [object type] in [region B] satisfying [condition B]?"

Question example (format only, not a default content choice):
- "What is the absolute difference between the counts of small vehicles in the upper-left quadrant near the red-roofed building and small vehicles in the lower-left quadrant adjacent to the L-shaped warehouse?"

Answer options format:
- The 4 options must be 4 distinct non-negative integer values.
- Exactly one option must equal the final absolute difference (S9).
- Use plausible nearby distractors when possible.
- Do not use region labels, ranges, or text options such as "same".
- Example: A: "0", B: "1", C: "2", D: "3"

How to use the annotation context:
- Start from "current_category_candidates".
- Choose [region A], [region B], and [object type] from one precomputed comparison candidate entry.
- Use the candidate entry's "object_type_label" in the question and step text. Do not expose raw underscore category ids such as "small_vehicle".
- Use the corresponding raw boxes in the "regions" section together with the image to decide [condition A] for [region A] and [condition B] for [region B].
- Prefer different conditions that are unique to each region, but use the same condition for both if no distinct unique conditions can be found.
- The final filtered boxes in reasoning supervision must be subsets of the chosen region/category box lists in the "regions" section.
- Use the "regions" section only to inspect additional region-level context if needed. Do not invent a region pair or object type outside the candidate list unless the comparison candidate list is empty.

Candidate-selection rules:
- Prefer candidate entries where:
  - "recommended_for_condition_search" is true
  - both regions are non-empty
  - support_size is "compact" or "manageable"
- Prefer candidates whose raw regional counts are close enough that a small numeric difference remains plausible after condition filtering.
- Do not default to the example's region pair, object type, or condition.
- Each condition ([condition A] and [condition B]) must be derived from local visual structure around the chosen candidate boxes in its respective region, not only from the object category itself.
- Each condition should be image-specific and supported around every listed box in its region.
- Each condition must refer to a **specific, visually unique, and spatially localizable landmark or reference structure** visible in the image. The landmark must be identifiable by a distinctive visual attribute (e.g., unique color, unique shape, unique size, unique pattern) so that it can be unambiguously located in the region.
  - Good examples: "near the red-roofed building", "adjacent to the L-shaped warehouse", "within the fenced rectangular lot", "beside the circular water tank", "along the tree-lined boulevard on the east side".
  - Bad examples (too broad or generic): "near a road", "visible on pavement", "in an open area", "near buildings", "on grass". These are unacceptable because they cannot be pinpointed to a single localizable structure in the image.
- The landmark defining each condition must itself be localizable with a bounding box. If you cannot draw a clear bounding box around the reference structure, the condition is too vague - reject it and pick a more specific one.
- Each condition must narrow the candidate set in its region. If it still leaves a very dense, messy, hard-to-verify, or semantically inconsistent set when another candidate is cleaner, reject it and choose another candidate.
- After condition filtering, both regional counts must be unambiguous.
- Prefer questions whose final absolute difference is small but non-trivial, ideally 1 to 3, when the image supports that choice cleanly.

How to determine the correct answer:
- Locate the landmark defining [condition A] in [region A] -> report its bbox in S1.
- Filter objects in region A by [condition A] -> list their boxes in S3 -> count = S4.
- Locate the landmark defining [condition B] in [region B] -> report its bbox in S5.
- Filter objects in region B by [condition B] -> list their boxes in S7 -> count = S8.
- Compute the absolute difference: S9 = abs(S4 - S8).
- The correct option is the one whose integer value equals S9.

Fixed reasoning supervision schema:
- The steps array must contain exactly 9 steps: S1 to S9.
- S1 answer_format must be "bbox_coordinates".
- S2 answer_format must be "boolean".
- S3 answer_format must be "bbox_coordinates_list".
- S4 answer_format must be "integer".
- S5 answer_format must be "bbox_coordinates".
- S6 answer_format must be "boolean".
- S7 answer_format must be "bbox_coordinates_list".
- S8 answer_format must be "integer".
- S9 answer_format must be "integer".

Step ground-truth rules:
- S1 is the bbox coordinates of the specific landmark or reference structure in region A that defines [condition A].
- S2 is true when region A contains at least one [object type] satisfying [condition A], false otherwise.
- S3 is the list of bbox coordinates of all [object type] in region A satisfying [condition A]. Empty list [] when S2 is false.
- S4 equals len(S3).
- S5 is the bbox coordinates of the specific landmark or reference structure in region B that defines [condition B].
- S6 is true when region B contains at least one [object type] satisfying [condition B], false otherwise.
- S7 is the list of bbox coordinates of all [object type] in region B satisfying [condition B]. Empty list [] when S6 is false.
- S8 equals len(S7).
- S9 equals abs(S4 - S8).

Answer mapping:
- The correct option is the one whose integer value equals S9.

Schema example (slot-based skeleton; replace every slot with current-image content):
```json
{
  "reasoning_supervision": {
    "step_type": "multi_step_qa",
    "steps": [
      {
        "step_id": "S1",
        "question": "Locate the landmark that defines [condition A] in [region A]. Report its bbox coordinates.",
        "ground_truth": [0.050, 0.150, 0.130, 0.250],
        "answer_format": "bbox_coordinates"
      },
      {
        "step_id": "S2",
        "question": "Does [region A] contain any [object type] satisfying [condition A]?",
        "ground_truth": true,
        "answer_format": "boolean"
      },
      {
        "step_id": "S3",
        "question": "Locate every [object type] in [region A] that satisfy [condition A]. Report bbox coordinates.",
        "ground_truth": [
          [0.101, 0.202, 0.118, 0.220]
        ],
        "answer_format": "bbox_coordinates_list"
      },
      {
        "step_id": "S4",
        "question": "How many such [object type] are in [region A] that satisfy [condition A]?",
        "ground_truth": 1,
        "answer_format": "integer"
      },
      {
        "step_id": "S5",
        "question": "Locate the landmark that defines [condition B] in [region B]. Report its bbox coordinates.",
        "ground_truth": [0.550, 0.650, 0.680, 0.780],
        "answer_format": "bbox_coordinates"
      },
      {
        "step_id": "S6",
        "question": "Does [region B] contain any [object type] satisfying [condition B]?",
        "ground_truth": true,
        "answer_format": "boolean"
      },
      {
        "step_id": "S7",
        "question": "Locate every [object type] in [region B] that satisfy [condition B]. Report bbox coordinates.",
        "ground_truth": [
          [0.601, 0.702, 0.618, 0.720],
          [0.640, 0.744, 0.658, 0.760]
        ],
        "answer_format": "bbox_coordinates_list"
      },
      {
        "step_id": "S8",
        "question": "How many such [object type] are in [region B] that satisfy [condition B]?",
        "ground_truth": 2,
        "answer_format": "integer"
      },
      {
        "step_id": "S9",
        "question": "Compute the absolute difference |Step 4 count - Step 8 count|.",
        "ground_truth": 1,
        "answer_format": "integer"
      }
    ]
  }
}
```

Exact step wording pattern:
- S1: "Locate the landmark that defines [condition A] in [region A]. Report its bbox coordinates."
- S2: "Does [region A] contain any [object type] satisfying [condition A]?"
- S3: "Locate every [object type] in [region A] that satisfy [condition A]. Report bbox coordinates."
- S4: "How many such [object type] are in [region A] that satisfy [condition A]?"
- S5: "Locate the landmark that defines [condition B] in [region B]. Report its bbox coordinates."
- S6: "Does [region B] contain any [object type] satisfying [condition B]?"
- S7: "Locate every [object type] in [region B] that satisfy [condition B]. Report bbox coordinates."
- S8: "How many such [object type] are in [region B] that satisfy [condition B]?"
- S9: "Compute the absolute difference |Step 4 count - Step 8 count|."

Category-specific validation checklist:
- The chosen region pair and object type must come from one comparison candidate entry unless the candidate list is empty.
- Each condition must reference a specific, visually unique, spatially localizable landmark - not a broad or generic feature.
- S1 must be a valid bbox that tightly encloses the landmark defining [condition A] in region A.
- S5 must be a valid bbox that tightly encloses the landmark defining [condition B] in region B.
- S1 bbox must lie within region A boundaries.
- S5 bbox must lie within region B boundaries.
- Every box in S3 must belong to region A and must be a subset of the chosen region/category box list in the "regions" section.
- Every box in S7 must belong to region B and must be a subset of the chosen region/category box list in the "regions" section.
- S3 and S7 must be disjoint sets of boxes.
- S3 must contain all and only the counted filtered boxes for region A.
- S7 must contain all and only the counted filtered boxes for region B.
- S2 must equal whether S3 is non-empty.
- S6 must equal whether S7 is non-empty.
- len(S3) must equal S4.
- len(S7) must equal S8.
- S9 must equal abs(S4 - S8).
- The 4 option values must be distinct non-negative integers.
- Exactly one option value must equal S9.
- The top-level "answer" letter must match the correct option.
- The chosen region pair, object type, and condition must be justified by the current image rather than by example reuse.

Invalid behavior to avoid:
- Ignoring the precomputed comparison candidates and inventing a different region pair first.
- Picking a region pair because it matches the example, then forcing the counts afterward.
- Writing conditions first and only loosely matching boxes to them.
- Reusing the same box in two different regions.
- Producing count values or a final difference that do not match the listed bbox coordinates.
- Using region names as answer options or making the final answer depend on "which region" rather than on the numeric difference.
- Using a signed difference or an unclear subtraction order instead of the absolute difference.
- Using broad, generic, or non-localizable conditions such as "near a road", "on pavement", "in an open area", or "near buildings". Every condition must pinpoint a specific identifiable structure with a clear bounding box.
- Providing a landmark bbox in S1 or S5 that does not correspond to a real, visible structure in the image.
\end{lstlisting}

\subsubsection*{Category Category: Logical Verification}

\begin{lstlisting}[style=promptlst]
- category: "logical_verification"
- subcategory: "orientation_verification"

Fixed question template:
- "Which statement correctly describes the presence of [orientation-type] [object type] in [region A] and [region B]?"

Question example (format only, not a default content choice):
- "Which statement correctly describes the presence of horizontally oriented vehicles in the upper-left and upper-right quadrants?"

Answer options format (use exactly these four patterns, replacing slots with current-image content):
- A: "[orientation-type] [object type] are present in both [region A] and [region B]"
- B: "[orientation-type] [object type] are present in [region A] but not in [region B]"
- C: "[orientation-type] [object type] are present in [region B] but not in [region A]"
- D: "[orientation-type] [object type] are absent from both [region A] and [region B]"

How to use the annotation context:
- Start from "current_category_candidates".
- Choose one precomputed logical candidate entry and use its [orientation-type], [object type], [region A], [region B], and region_a_boxes directly.
- Use the candidate entry's "object_type_label" in the question, step text, and option text. Do not expose raw underscore category ids such as "small_vehicle".
- Do not invent a different region pair, object type, or orientation pattern unless the logical_verification candidate list is empty.
- Use the "regions" section only for additional region-level inspection if needed.

Candidate-selection rules:
- Prefer candidate entries where:
  - "recommended" is true
  - support_size is "compact" or "manageable"
- Prioritize candidates whose object type is one of: swimming pool, small vehicle, ship, helicopter, tennis court. These object types tend to have visually distinctive orientations and produce cleaner questions. Only fall back to other object types when none of the prioritized types appear in the candidate list.
- The orientation type must be exactly "horizontal" or "vertical".
- Determine the ground-truth orientation from the precomputed metadata only.
- Do not default to "horizontal small vehicle" unless it is actually the strongest current-image candidate.
- Exactly one of the four options must be correct based on the actual presence in both regions.
- Do not default to the example's region pair, object type, or orientation unless it is actually the strongest current-image candidate.

How to determine the correct answer:
- Count [orientation-type] [object type] in region A (S2) and region B (S4).
- If both S2 > 0 and S4 > 0 -> answer A (both present).
- If S2 > 0 and S4 = 0 -> answer B (region A only).
- If S2 = 0 and S4 > 0 -> answer C (region B only).
- If S2 = 0 and S4 = 0 -> answer D (neither present).

Fixed reasoning supervision schema:
- The steps array must contain exactly 5 steps: S1 to S5.
- S1 answer_format must be "bbox_coordinates_list".
- S2 answer_format must be "integer".
- S3 answer_format must be "bbox_coordinates_list".
- S4 answer_format must be "integer".
- S5 answer_format must be "boolean".

Step ground-truth rules:
- S1 is the list of bbox coordinates of all [orientation-type] [object type] in region A. Empty list [] when none exist.
- S2 equals len(S1).
- S3 is the list of bbox coordinates of all [orientation-type] [object type] in region B. Empty list [] when none exist.
- S4 equals len(S3).
- S5 is true when S2 > 0 and S4 > 0, false otherwise.

Answer mapping:
- S2 > 0 and S4 > 0 -> answer A.
- S2 > 0 and S4 = 0 -> answer B.
- S2 = 0 and S4 > 0 -> answer C.
- S2 = 0 and S4 = 0 -> answer D.

Schema example (slot-based skeleton; replace every slot with current-image content):
```json
{
  "reasoning_supervision": {
    "step_type": "multi_step_qa",
    "steps": [
      {
        "step_id": "S1",
        "question": "Locate every [orientation-type] [object type](s) in [region A]. Report bbox coordinates.",
        "ground_truth": [
          [0.141, 0.244, 0.160, 0.255]
        ],
        "answer_format": "bbox_coordinates_list"
      },
      {
        "step_id": "S2",
        "question": "How many [orientation-type] [object type] are in [region A]?",
        "ground_truth": 1,
        "answer_format": "integer"
      },
      {
        "step_id": "S3",
        "question": "Locate every [orientation-type] [object type](s) in [region B]. Report bbox coordinates.",
        "ground_truth": [],
        "answer_format": "bbox_coordinates_list"
      },
      {
        "step_id": "S4",
        "question": "How many [orientation-type] [object type] are in [region B]?",
        "ground_truth": 0,
        "answer_format": "integer"
      },
      {
        "step_id": "S5",
        "question": "Based on the counts from Step 2 and Step 4, do both [region A] and [region B] contain at least one [orientation-type] [object type]?",
        "ground_truth": false,
        "answer_format": "boolean"
      }
    ]
  }
}
```

Exact step wording pattern:
- S1: "Locate every [orientation-type] [object type](s) in [region A]. Report bbox coordinates."
- S2: "How many [orientation-type] [object type] are in [region A]?"
- S3: "Locate every [orientation-type] [object type](s) in [region B]. Report bbox coordinates."
- S4: "How many [orientation-type] [object type] are in [region B]?"
- S5: "Based on the counts from Step 2 and Step 4, do both [region A] and [region B] contain at least one [orientation-type] [object type]?"

Category-specific validation checklist:
- The chosen orientation, object type, and region pair must come from one logical_verification candidate entry unless the candidate list is empty.
- Every box in S1 must belong to region A and must be one of the chosen candidate entry's region_a_boxes.
- len(S1) must equal S2.
- Every box in S3 must belong to region B. S3 must be empty when the candidate entry's region_b_count is 0.
- len(S3) must equal S4.
- S5 must equal (S2 > 0 and S4 > 0).
- The 4 options must follow the fixed presence/absence pattern exactly.
- The top-level "answer" letter must follow the answer mapping.
- The chosen orientation, object type, and region pair must be justified by the current image rather than by example reuse.

Invalid behavior to avoid:
- Ignoring the precomputed logical candidates and inventing a different region pair first.
- Using a different orientation from the chosen candidate entry.
- Listing only a convenient subset of positive boxes in S1 when the chosen candidate entry contains more.
- Saying region B has none when the chosen candidate entry still indicates a valid positive count there.
- Using options that do not follow the fixed four-way presence/absence pattern.
\end{lstlisting}

\subsubsection*{Category Category: Relational Inference}

\begin{lstlisting}[style=promptlst]
- category: "relational_inference"
- subcategory: "adjacent_region_extreme_object_relation"

Fixed question template:
- "What is the positional relationship between the [selector A] [object type] in [region A] and the [selector B] [object type] in [region B]?"

Question example (format only, not a default content choice):
- "What is the positional relationship between the leftmost aircraft in the upper-left quadrant and the rightmost aircraft in the upper-right quadrant?"

Answer options format:
- For left-right adjacent regions (cross-axis: north/south):
  - A: "The [selector A] [object type] in [region A] is north of the [selector B] [object type] in [region B]"
  - B: "The [selector A] [object type] in [region A] is south of the [selector B] [object type] in [region B]"
  - C: "The [region A] does not contain any [object type]"
  - D: "The [region B] does not contain any [object type]"
- For top-bottom adjacent regions (cross-axis: west/east):
  - A: "The [selector A] [object type] in [region A] is west of the [selector B] [object type] in [region B]"
  - B: "The [selector A] [object type] in [region A] is east of the [selector B] [object type] in [region B]"
  - C: "The [region A] does not contain any [object type]"
  - D: "The [region B] does not contain any [object type]"
- Option A always states the first spatial relation (north / west). Option B always states the opposite (south / east). Option C always states region A lacks the object. Option D always states region B lacks the object.

How to use the annotation context:
- Start from "current_category_candidates".
- For answer A or B (both regions contain the object): choose one precomputed relational candidate entry and use its [region A], [region B], [object type], [selector A], [selector B], selected boxes, and relation_truths directly.
- For answer C or D (one region lacks the object): pick an adjacent region pair and an object type from the "regions" section where one region contains instances of that object type but the adjacent region does not. Use the extreme selector from the region that has the object.
- Use the candidate entry's or region section's "object_type_label" in the question and step text. Do not expose raw underscore category ids such as "small_vehicle".
- Use the "regions" section to verify object existence per region and for additional region-level inspection.

Candidate-selection rules:
- Prefer candidate entries where:
  - "recommended" is true
  - support_size is "compact" or "manageable"
- Region A and Region B must be adjacent regions only.
- Never use diagonal region pairs.
- Ask only about the cross-axis for the chosen adjacent pair:
  - left-right adjacent regions -> "north of" / "south of"
  - top-bottom adjacent regions -> "west of" / "east of"
- Exactly one of the four options must be correct.
- The chosen object type must exist in at least one of the two regions. Both regions lacking the object is forbidden.
- Do not default to the example's region pair, object type, selectors, or relation unless it is actually the strongest current-image candidate.
- The chosen region pair, object type, selectors, and relation must be justified by the current image rather than by example reuse.

How to determine the correct answer:
- If region A does not contain any instance of the chosen object type -> answer is C.
- If region B does not contain any instance of the chosen object type -> answer is D.
- If both regions contain the object type, compare the center coordinates of the two selected bounding boxes on the cross-axis:
  - For left-right adjacent regions: compare y-center. Smaller y-center means north. If the object from region A is north of the object from region B -> A. Otherwise -> B.
  - For top-bottom adjacent regions: compare x-center. Smaller x-center means west. If the object from region A is west of the object from region B -> A. Otherwise -> B.

Fixed reasoning supervision schema:
- The steps array must contain exactly 5 steps: S1 to S5.
- S1 answer_format must be "boolean".
- S2 answer_format must be "bbox_coordinates".
- S3 answer_format must be "boolean".
- S4 answer_format must be "bbox_coordinates".
- S5 answer_format must be "boolean".

Step ground-truth rules:
- S1 is true when region A contains the chosen object type, false otherwise.
- S2 is the bbox coordinates of the selected object in region A when S1 is true, null when S1 is false.
- S3 is true when region B contains the chosen object type, false otherwise.
- S4 is the bbox coordinates of the selected object in region B when S3 is true, null when S3 is false.
- S5 determines whether the first spatial relation (option A) holds by comparing the center coordinates. S5 is true when option A is correct, false when option B is correct, null when S1 or S3 is false.

Answer mapping:
- S1 is false -> answer C.
- S1 is true and S3 is false -> answer D.
- S1 is true and S3 is true and S5 is true -> answer A.
- S1 is true and S3 is true and S5 is false -> answer B.

Schema example (both regions contain the object; replace every slot with current-image content):
```json
{
  "reasoning_supervision": {
    "step_type": "multi_step_qa",
    "steps": [
      {
        "step_id": "S1",
        "question": "Does [region A] contain any [object type]?",
        "ground_truth": true,
        "answer_format": "boolean"
      },
      {
        "step_id": "S2",
        "question": "Locate the [selector A] [object type] in [region A]. Report bbox coordinates.",
        "ground_truth": [0.132, 0.188, 0.176, 0.224],
        "answer_format": "bbox_coordinates"
      },
      {
        "step_id": "S3",
        "question": "Does [region B] contain any [object type]?",
        "ground_truth": true,
        "answer_format": "boolean"
      },
      {
        "step_id": "S4",
        "question": "Locate the [selector B] [object type] in [region B]. Report bbox coordinates.",
        "ground_truth": [0.702, 0.144, 0.760, 0.194],
        "answer_format": "bbox_coordinates"
      },
      {
        "step_id": "S5",
        "question": "Comparing the center coordinates of the bounding box from Step 2 and the bounding box from Step 4, is the object from Step 2 [relation 1] the object from Step 4?",
        "ground_truth": true,
        "answer_format": "boolean"
      }
    ]
  }
}
```

Schema example (region A does not contain the object):
```json
{
  "reasoning_supervision": {
    "step_type": "multi_step_qa",
    "steps": [
      {
        "step_id": "S1",
        "question": "Does [region A] contain any [object type]?",
        "ground_truth": false,
        "answer_format": "boolean"
      },
      {
        "step_id": "S2",
        "question": "Locate the [selector A] [object type] in [region A]. Report bbox coordinates.",
        "ground_truth": null,
        "answer_format": "bbox_coordinates"
      },
      {
        "step_id": "S3",
        "question": "Does [region B] contain any [object type]?",
        "ground_truth": true,
        "answer_format": "boolean"
      },
      {
        "step_id": "S4",
        "question": "Locate the [selector B] [object type] in [region B]. Report bbox coordinates.",
        "ground_truth": [0.702, 0.144, 0.760, 0.194],
        "answer_format": "bbox_coordinates"
      },
      {
        "step_id": "S5",
        "question": "Comparing the center coordinates of the bounding box from Step 2 and the bounding box from Step 4, is the object from Step 2 [relation 1] the object from Step 4?",
        "ground_truth": null,
        "answer_format": "boolean"
      }
    ]
  }
}
```

Exact step wording pattern:
- S1: "Does [region A] contain any [object type]?"
- S2: "Locate the [selector A] [object type] in [region A]. Report bbox coordinates."
- S3: "Does [region B] contain any [object type]?"
- S4: "Locate the [selector B] [object type] in [region B]. Report bbox coordinates."
- S5: "Comparing the center coordinates of the bounding box from Step 2 and the bounding box from Step 4, is the object from Step 2 [relation 1] the object from Step 4?"

Category-specific validation checklist:
- Region A and Region B must be one of the allowed adjacent pairs.
- Never use diagonal region pairs.
- The chosen object type must exist in at least one of the two regions. Both S1 and S3 must not be false simultaneously.
- For answers A or B: the chosen region pair, object type, selectors, and selected boxes must come from one relational_inference candidate entry unless the candidate list is empty.
- S2 must exactly match the chosen candidate entry's region_a_box when S1 is true.
- S4 must exactly match the chosen candidate entry's region_b_box when S3 is true.
- S2 ground_truth must be null when S1 is false.
- S4 ground_truth must be null when S3 is false.
- S5 ground_truth must be null when S1 or S3 is false.
- The object types in S2 and S4 must match.
- The spatial relation axis must be the cross-axis for the chosen adjacent pair.
- S5 ground_truth must agree with the relation_truths when both regions contain the object.
- The top-level answer must follow the answer mapping.
- The chosen region pair, object type, selectors, and relation must be justified by the current image rather than by example reuse.

Invalid behavior to avoid:
- Ignoring the precomputed relational candidates and inventing a different adjacent pair first.
- Using a diagonal region pair.
- Using a same-axis relation (e.g., east/west for left-right adjacent regions).
- Changing the selector pair after picking the candidate entry.
- Choosing an object type that does not exist in either region.
- Having both S1 and S3 as false.
- Having S5 as true or false when S1 or S3 is false.
- Having S2 or S4 as non-null when the corresponding S1 or S3 is false.
\end{lstlisting}

\subsection{WSI Construction Templates}
\label{app:wsi_template}

WSI GT-CoT annotations are constructed using four question templates (T1--T4), each targeting a specific reasoning pattern.
The output JSON format follows the same schema as CCTV and RS.
When the correct answer cannot be reliably derived from structured legacy labels (e.g., for many T4 instances), the \texttt{answer} field is left empty and filled by a domain expert.

\begin{lstlisting}[style=promptlst]
# WSI QA Construction Templates

## Overall design

The current QA construction framework is organized into **three reasoning categories**:

1. **Comparison**
2. **Logical Verification**
3. **Relational Inference**

Each template should satisfy the following general principles:

### General Principle 1: Prefer same-level variables
Variables used in one question should belong to the same semantic level whenever possible. Avoid mixing diagnosis type, growth pattern, vascular spread, and unrelated morphology in a single option set.

### General Principle 2: Prefer 2-3 core elements
For templates that use integrated summaries or multiple attributes, prefer **2-3 core attributes** instead of using too many variables at once. Too many elements often create:
- clinically implausible combinations
- obvious distractors
- text-only shortcuts

### General Principle 3: Distractors should be near-miss alternatives
Wrong options should not be obviously false or conceptually contradictory. Whenever possible, each distractor should differ from the correct answer by **only one key attribute**.

### General Principle 4: Legacy labels are mainly used for candidate generation
Existing WSIVQA attributes can often support:
- question selection
- variable instantiation
- option construction

However, some templates, especially relational ones, may still require **expert annotation for the final answer**.

### General Principle 5: Answer field rule in output JSON
When the correct answer **can be derived from legacy labels**, put the correct option letter in the `answer` field.

When the correct answer **cannot be reliably derived from legacy labels** (for example, many `T4` cases), leave the `answer` field empty:

```json
"answer": ""
```

---

## Output JSON format

Each QA item should follow this format:

```json
{
  "slide": "TCGA-AQ-A54N",
  "question": "Compared with a hypothetical multifocal pattern, which growth pattern is better supported by the visible evidence in this slide?",
  "options": {
    "A": "Multifocal growth pattern with spatially separated tumor foci",
    "B": "Unifocal growth pattern centered on a single dominant lesion",
    "C": "Purely in-situ growth pattern without invasion",
    "D": "Diffuse lymphovascular spread pattern"
  },
  "answer": "B",
  "category": "comparison"
}
```

If no reliable answer can be derived from legacy labels:

```json
{
  "slide": "TCGA-XX-XXXX",
  "question": "Based on the visual evidence, what is the relationship between the in-situ component and the invasive component?",
  "options": {
    "A": "The invasive component is the dominant process, and the in-situ component is present only as a secondary component.",
    "B": "The in-situ and invasive components both appear substantial, without a clearly dominant component.",
    "C": "The invasive component is supported, but a distinct in-situ component is not convincingly established.",
    "D": "The relationship between the in-situ and invasive components cannot be determined confidently from the visible evidence."
  },
  "answer": "",
  "category": "relational_inference"
}
```

---

# Category 1: Comparison

## Template T1: Interpretation Comparison

### Goal
Compare two competing interpretations and determine which one is better supported by the visible evidence.

### Question Template
**When comparing [A] and [B], which interpretation is better supported by the visible evidence?**

### Options Template
- **A.** The visible evidence clearly favors **[A]** over **[B]**.
- **B.** The visible evidence clearly favors **[B]** over **[A]**.
- **C.** The visible evidence leans toward **[A]**, but **[B]** remains a plausible alternative.
- **D.** The visible evidence leans toward **[B]**, but **[A]** remains a plausible alternative.

### Variable Space

#### [A], [B]
Two competing interpretations that are:
- visually distinguishable in principle
- semantically comparable
- on the same comparison axis

#### Preferred variable types
- **unifocal** vs **multifocal**
- **purely invasive interpretation** vs **mixed invasive-and-in-situ interpretation**
- **distinct in-situ component present** vs **no convincing in-situ component**

#### Avoid
- unifocal vs lymphovascular invasion
- invasive vs multifocal
- in-situ vs vascular spread
- any pair crossing different semantic levels

### Variable Selection Rules

#### Rule 1
Choose **only one comparison axis** per question.

#### Rule 2
Prefer pairs that are meaningful at the **slide level** and can plausibly be inferred from visible evidence.

#### Rule 3
Prefer interpretation pairs that are not trivially distinguishable by text alone.

#### Rule 4
If using old labels, prioritize attributes already available in WSIVQA, such as:
- focality
- in-situ presence
- invasive status
- LVI presence/absence

### Construction Notes
- T1 is well suited for old-label-driven automatic generation.
- The question should remain neutral and should not pre-embed the answer direction.
- Avoid replacing this template with direct morphology definitions.
- If the correct interpretation can be derived from normalized legacy labels, fill `answer` with the correct option letter.

---

# Category 2: Logical Verification

## Template T2: Multi-Attribute Verification

### Goal
Verify whether several slide-level findings are supported together by the visible evidence.

### Question Template
**Does the slide support all of the following together: [x], [y], and [z]?**

### Options Template
- **A.** Yes, all three findings are supported together by the visible evidence.
- **B.** No, **[x]** is not adequately supported.
- **C.** No, **[y]** is not adequately supported.
- **D.** No, **[z]** is not adequately supported.

### Variable Space

#### [x], [y], [z]
Three slide-level attributes that can plausibly be supported or contradicted by image evidence.

#### Preferred variable types
- **high-grade lesion**
- **unifocal growth**
- **no lymphovascular invasion**
- **distinct in-situ component present**
- **invasive carcinoma present**

#### Recommended combinations
- grade + focality + LVI
- focality + in-situ + LVI
- invasive + in-situ + focality

#### Avoid
- more than 3 attributes
- mixing attributes with very different certainty levels
- combinations that are likely to produce obvious contradictions

### Variable Selection Rules

#### Rule 1
Use **exactly 3 attributes**.

#### Rule 2
All three attributes should be interpretable at the **same slide-summary level**.

#### Rule 3
Prefer combinations with good clinical coherence.

#### Rule 4
If some attributes are much less visually reliable than others, avoid using them together in the same question.

#### Rule 5
When generating from legacy labels, only include attributes that can be reasonably grounded in available slide annotations.

### Construction Notes
- T2 is one of the most suitable templates for legacy-label-based generation.
- The options should remain strictly parallel.
- Avoid "Yes, but..." style options, which create text-only logical shortcuts.
- If all three attributes are available and the correct answer can be determined from labels, fill `answer` with the correct option letter.

---

## Template T3: Summary Selection

### Goal
Select the integrated slide-level summary that is most consistent with the visible evidence.

### Question Template
Preferred version:
**Which integrated summary is most consistent with the overall findings of this slide?**

Alternative shorter version:
**Which summary is more consistent with the visible evidence?**

### Options Template
Each option should be a short integrated summary composed of **2-3 core attributes**.

Recommended generation strategy:
- one correct summary using the true attribute combination
- three distractors, each created by flipping only one key attribute

#### Example option pattern
- **A.** A high-grade invasive carcinoma with unifocal growth and no lymphovascular invasion.
- **B.** A high-grade invasive carcinoma with multifocal growth and no lymphovascular invasion.
- **C.** A high-grade invasive carcinoma with unifocal growth and lymphovascular invasion present.
- **D.** A lower-grade invasive carcinoma with unifocal growth and no lymphovascular invasion.

### Variable Space

#### Core attributes for summary construction
Prefer using only **2-3** of the following:
- **grade**
- **focality**
- **LVI**
- **in-situ component**
- **invasive status**

#### Preferred combinations
- grade + focality + LVI
- focality + LVI + in-situ
- grade + focality
- grade + in-situ

#### Avoid
- summaries with too many attributes
- obvious clinical contradictions
- distractors that differ from the correct answer in multiple ways

### Variable Selection Rules

#### Rule 1
Use only **2-3 core attributes** per summary.

#### Rule 2
Keep all options at the same semantic level:
- all should look like plausible integrated case summaries

#### Rule 3
Each wrong option should ideally flip **only one attribute** relative to the correct option.

#### Rule 4
Avoid adding extra detail merely to make options longer.

#### Rule 5
If a variable produces too many clinically implausible combinations, remove it from this template.

### Construction Notes
- T3 can be generated from legacy labels, but it is more vulnerable to text-only shortcuts than T1 and T2.
- It works best when all options are short, parallel, and attribute-balanced.
- This template should be treated as a **slide-level summary reasoning task**, not a region-grounded evidence task.
- If the correct summary can be derived from normalized legacy labels, fill `answer` with the correct option letter.

---

# Category 3: Relational Inference

## Template T4: Component Relationship Inference

### Goal
Infer the relationship between two components based on visual evidence, especially in terms of dominance, coexistence, or uncertainty.

### Question Template
**Based on the visual evidence, what is the relationship between [component_A] and [component_B]?**

### Options Template
- **A.** **[component_B]** is the dominant process, and **[component_A]** is present only as a secondary component.
- **B.** **[component_A]** and **[component_B]** both appear substantial, without a clearly dominant component.
- **C.** **[component_B]** is supported, but **[component_A]** is not convincingly established as a distinct additional component.
- **D.** The relationship between **[component_A]** and **[component_B]** cannot be determined confidently from the visible evidence.

### Variable Space

#### [component_A]
A component that may be:
- secondary
- accompanying
- subordinate
- visually harder to assess in dominance relation

#### Preferred [component_A]
- **in-situ component**
- **intraductal component**
- **secondary lower-grade-like component**
- **secondary architectural pattern**

#### [component_B]
A component that can plausibly serve as the dominant process.

#### Preferred [component_B]
- **invasive component**
- **dominant invasive process**
- **high-grade invasive component**

### Variable Selection Rules

#### Rule 1
Choose component pairs that can plausibly coexist in the same slide.

#### Rule 2
Prefer pairs whose relationship is **not trivially answerable from a thumbnail**.

#### Rule 3
Prefer clinically meaningful hierarchy relations, such as:
- in-situ vs invasive
- intraductal vs invasive
- secondary lower-grade-like component vs dominant high-grade invasive component

#### Rule 4
Do not generate this template if the old labels do not even support the plausible presence of both components.

#### Rule 5
For this template, legacy labels should mainly be used to:
- identify candidate component pairs
- generate the question and options

The final answer may still need expert annotation.

### Construction Notes
- T4 is not a simple yes/no template.
- The four options correspond to four distinct relationship states:
  - dominant-secondary
  - co-substantial coexistence
  - only component_B convincingly established
  - indeterminate relationship
- This template is best used when the goal is to capture harder relational reasoning that cannot be fully derived from legacy labels alone.
- In many cases, `answer` should be left empty unless a trusted source explicitly supports the relationship.

---

# Recommended Attribute Priorities

## High-priority legacy-label attributes
These are the most useful attributes for automatic question generation:
- **grade**
- **focality**
- **lymphovascular invasion**
- **in-situ presence**
- **invasive status**

## Best template usage by attribute type

### Focality
Best suited for:
- T1
- T2
- T3

### Grade
Best suited for:
- T2
- T3
- sometimes T1 if used carefully

### LVI
Best suited for:
- T2
- T3

### In-situ / invasive coexistence
Best suited for:
- T1
- T2
- T4

---

# Suggested Construction Pipeline for Code

The next-step code generation can follow this logic:

## Step 1: Extract slide-level attributes from WSIVQA
For each slide, gather all available structured attributes such as:
- grade
- focality
- LVI
- in-situ presence
- invasive status
- histologic type
- any other relevant labels

## Step 2: Normalize attribute values
Convert free-text labels into normalized values suitable for templating, such as:
- `grade_1` / `grade_2` / `grade_3`
- `unifocal` / `multifocal`
- `lvi_present` / `lvi_absent`
- `in_situ_present` / `in_situ_absent`

## Step 3: Determine eligible templates for each slide
For each slide, check which templates are allowed based on available attributes:
- T1 requires a valid comparison pair
- T2 requires 3 usable slide-level attributes
- T3 requires 2-3 summary attributes
- T4 requires a plausible component pair

## Step 4: Instantiate question and options
Fill each template with the selected variables according to the rules above.

## Step 5: Assign answer
- For **T1-T3**, the answer can often be derived from normalized labels if the QA is label-driven.
- For **T4**, the answer may need expert annotation unless a trusted label source supports the relationship directly.
- If the answer cannot be reliably derived, leave it empty:

```json
"answer": ""
```

## Step 6: Export JSON
Export one QA item per generated question using the required schema.

---

# Final Minimal Template Summary

## Comparison

### T1: Interpretation Comparison
**Question**
When comparing **[A]** and **[B]**, which interpretation is better supported by the visible evidence?

**Options**
- clearly favors A
- clearly favors B
- leans toward A but B remains plausible
- leans toward B but A remains plausible

---

## Logical Verification

### T2: Multi-Attribute Verification
**Question**
Does the slide support all of the following together: **[x]**, **[y]**, and **[z]**?

**Options**
- yes, all supported
- no, x not supported
- no, y not supported
- no, z not supported

### T3: Summary Selection
**Question**
Which integrated summary is most consistent with the overall findings of this slide?

**Options**
- 4 short summary statements
- use only 2-3 core attributes
- one correct summary, three single-attribute flips

---

## Relational Inference

### T4: Component Relationship Inference
**Question**
Based on the visual evidence, what is the relationship between **[component_A]** and **[component_B]**?

**Options**
- B dominant, A secondary
- A and B both substantial
- B supported, A not convincingly established
- relationship indeterminate

---

# Final Implementation Reminder

When implementing the generator:
- extract all usable slide-level attributes first
- normalize labels before template selection
- only instantiate templates whose required variables are available
- fill `answer` only when it can be derived reliably from legacy labels
- otherwise leave `answer` as an empty string
- export all generated QA items as JSON objects with the required schema

---

# Extracted CoT Questions And Options From WSI JSON

Source example:
- `/tealab-data/visual_reasoning/benchmark/qa/wsi_120/TCGA-A1-A0SJ.json`
- field path: `categories.comparison.reasoning_supervision.steps`

Note:
- The current WSI JSON does not provide a separate free-form `cot` paragraph.
- The reasoning supervision chain for this example runs from `Q1` to `Q6.3`.
- Below I am attaching the extracted `question + options` directly from the JSON.

## Q1

**Question**
At whole-slide / low magnification, which global regions/components should be screened first? (This is a multi-select question; select all that apply.)

**Options**
- Tumor-rich region
- Invasive component
- In-situ component
- Tumor-stroma interface
- Margin / edge area
- Necrotic area
- Hemorrhagic area
- Stromal/desmoplastic region
- Hotspot region with highest atypia/cellularity
- Background benign/normal tissue for comparison

## Q2

**Question**
Is closer ROI review needed? (Single-choice: Yes or No. If No, Q2.1-Q4 are left blank.)

**Options**
- Yes
- No

## Q2.1

**Question**
Which ROI region(s) should be selected for closer review? Mentally overlay a 6x6 grid on the whole-slide image. ROI indices run from 1 to 36 in row-major order, from left to right within each row and from top to bottom across rows: ROI 1 is the top-left cell, ROI 6 is the top-right cell, ROI 31 is the bottom-left cell, and ROI 36 is the bottom-right cell. Answer with one or more integers from 1 to 36. (Answer only if Q2 = Yes; otherwise N/A.)

**Options**
- No explicit options list in JSON; answer format is ROI integer selection from 1 to 36.

## Q3

**Question**
For the closer review, what practical starting magnification should a junior pathologist use first? (Single-choice. Answer only if Q2 = Yes; otherwise N/A.)

**Options**
- 10x
- 20x
- 40x

## Q4

**Question**
After zooming in, which fine-grained morphologic features should be checked? (Multi-select, up to 5 most relevant. Answer only if Q2 = Yes; otherwise N/A.)

**Options**
- Nuclear atypia
- Nuclear pleomorphism
- Hyperchromasia
- Mitoses
- Atypical mitoses
- Cell density
- Gland formation
- Lumen formation
- Keratinization
- Necrosis
- Apoptosis
- Infiltrative boundary
- Tumor-stroma interface
- Stromal/desmoplastic reaction
- Mucin production
- Architectural irregularity
- Loss of polarity
- Cell cohesion / discohesion
- Vascular / lymphovascular invasion suspicion

## Q5

**Question**
Is a standardized pathology criterion/reference needed to make this judgment more reproducible? (Single-choice: Yes or No. If No, Q5.1-Q5.2 are left blank.)

**Options**
- Yes
- No

## Q5.1

**Question**
Which criterion/reference should be checked? (Multi-select. Answer only if Q5 = Yes; otherwise N/A.)

**Options**
- Nottingham grading criteria
- DCIS grading criteria
- Margin assessment standard
- Lymphovascular invasion criterion
- Tumor budding criterion

## Q5.2

**Question**
Why is the criterion needed here? (Multi-select. Answer only if Q5 = Yes; otherwise N/A.)

**Options**
- To improve reproducibility
- To resolve borderline morphology
- To distinguish competing interpretations
- To support grading/scoring consistency

## Q6.1

**Question**
Evidence (what is observed): select key observed clues from the Q4 feature list. (Multi-select.)

**Options**
- Nuclear atypia
- Nuclear pleomorphism
- Hyperchromasia
- Mitoses
- Atypical mitoses
- Cell density
- Gland formation
- Lumen formation
- Keratinization
- Necrosis
- Apoptosis
- Infiltrative boundary
- Tumor-stroma interface
- Stromal/desmoplastic reaction
- Mucin production
- Architectural irregularity
- Loss of polarity
- Cell cohesion / discohesion
- Vascular / lymphovascular invasion suspicion

## Q6.2

**Question**
Reason (why these clues matter): select the main morphologic interpretation indicated by the evidence above. (Single-choice.)

**Options**
- Cytologic malignancy
- Architectural abnormality
- Infiltrative growth
- Loss of normal organization
- Glandular differentiation
- Aggressive / high-grade morphology
- Stromal invasion pattern
- Malignant morphologic pattern

## Q6.3

**Question**
Conclusion (what this supports): state the conclusion most directly supported by the reasoning path above. (Short answer / free text.)

**Options**
- A high-grade pattern
- An invasive component
- An in-situ component
- A malignant gland-forming component
- Others (Please justify)
\end{lstlisting}

\end{document}